\documentclass{article}

\usepackage[english]{babel}

\usepackage[a4paper,
            left=2.5cm,
            right=2.5cm,
            top=2.8cm,
            bottom=4cm]{geometry}

\usepackage{imakeidx}
\usepackage{graphicx}
\usepackage{multicol}  
\usepackage[bottom]{footmisc}
\usepackage{newtxtext}
\usepackage{newtxmath}
\usepackage{natbib}
\usepackage{subcaption}
\usepackage{color}
\usepackage{dsfont}
\usepackage{booktabs}
\usepackage{siunitx}
\usepackage{verbatim}
\usepackage{tablefootnote}
\usepackage{wrapfig}
\usepackage{bm}
\usepackage[most]{tcolorbox}
\setkeys{Gin}{width=\linewidth, keepaspectratio, clip}

\usepackage{hyperref}
\hypersetup{
    colorlinks,
    citecolor=black,
    filecolor=black,
    linkcolor=black,
    urlcolor=black
}

\usepackage{authblk}
\usepackage{microtype}

\usepackage{caption}
\title{Machine Learning in Epidemiology}

\author[1,2,3]{Marvin N. Wright}
\author[1,2]{Lukas Burk}
\author[1,2]{Pegah Golchian}
\author[1,2]{Jan Kapar}
\author[1,2]{Niklas Koenen}
\author[1,2]{Sophie Hanna Langbein}

\affil[1]{Leibniz Institute for Prevention Research and Epidemiology - BIPS, Bremen, Germany}
\affil[2]{University of Bremen, Germany}
\affil[3]{University of Copenhagen, Denmark}

\date{}

\makeindex   

\begin{document}

\maketitle

\begin{tcolorbox}[colback=blue!5!white,colframe=blue!75!black]
This is the preprint of the following book chapter:\\
Wright, M.N., Burk, L., Golchian, P., Kapar, J., Koenen, N., Langbein, S.H. (2024). Machine Learning in Epidemiology. In: Ahrens, W., Pigeot, I. (eds) \textit{Handbook of Epidemiology}. Springer, New York, NY, \url{https://doi.org/10.1007/978-1-4614-6625-3_81-1}.
\end{tcolorbox}

\abstract{In the age of digital epidemiology, epidemiologists are faced by an increasing amount of data of growing complexity and dimensionality. Machine learning is a set of powerful tools that can help to analyze such enormous amounts of data. This chapter lays the methodological foundations for successfully applying machine learning in epidemiology. It covers the principles of supervised and unsupervised learning and discusses the most important machine learning methods. Strategies for model evaluation and hyperparameter optimization are developed and interpretable machine learning is introduced. All these theoretical parts are accompanied by code examples in \texttt{R}, where an example dataset on heart disease is used throughout the chapter.}


\section{Introduction}
\label{sec:intro}

Machine learning has become an integral part of almost all businesses and scientific fields alike, including epidemiology. With the rise of deep learning, machine learning revolutionized various applications, from image and speech recognition to natural language processing. Alongside this hype, epidemiologists are faced with an ever-increasing amount of data of growing complexity and dimensionality, including data from electronic health records, wearable devices, social media and genetics. Machine learning methods are able to efficiently analyze such enormous amounts of data. Thus, in the age of digital epidemiology, machine learning is an essential tool that every modern epidemiologist should know about.

One of the major advantages of machine learning is that it does not require exact model specifications. Instead, one simply indicates which variables or features to include, and relies on the machine learning method to find all the interactions and other important factors. However, this increased model flexibility comes at a high computational cost, a loss of interpretability and the risk of overfitting, i.e., fitting training data too closely, leading to poor generalization performance and the need for proper model evaluation. Further, most machine learning methods have to be configured by setting so-called hyperparameters, which heavily influence performance and thereby have to be chosen carefully or tested systematically. 

With this book chapter, we aim to give epidemiologists the foundation for successfully applying machine learning to their research. Notably, this does not require knowing all the details of all the different machine learning methods. Instead, we focus on general principles such as supervised learning (Sec.~\ref{sec:supervised}), model evaluation (Sec.~\ref{sec:eval}) and hyperparameter optimization (Sec.~\ref{sec:tuning}). Nevertheless, we cover two of the most important machine learning methods in Sec.~\ref{sec:supervised}. These methods are focused on making predictions, which is useful in many epidemiological tasks, however, does not help in understanding diseases, identifying risk factors or generating synthetic data. In this regard, Sec.~\ref{sec:iml} introduces the basics of interpretable machine learning and Sec.~\ref{sec:unsupervisedgen} covers unsupervised learning and generative modeling. Throughout the chapter, we use an example dataset on heart disease and show how to apply the covered methods in \texttt{R} using the \texttt{mlr3} framework \citep{mlr3}.\\

\noindent\textbf{Heart Disease Data Example}\newline

\noindent The \emph{heart disease} data \citep{heartdiseaseuci} are available from OpenML \citep{openML2013}. The labeled dataset contains $n=270$ instances of patients with $p=13$ features. The features include a patient's age (\texttt{age}), results of a thallium stress test (\texttt{thal}) and four types of chest pain (\texttt{chest\_pain}), among others. We aim to predict whether the target \texttt{heart\_disease} $y \in \{1,2\}$ is absent (1) or present (2). Since $y$ is categorical with two classes, it is a binary classification task. For details on the dataset, preprocessing, software and code examples, we refer to the appendix, our GitHub page \url{https://github.com/bips-hb/epi-handbook-ml}, and the \texttt{mlr3} book \citep{Bischl2024}.

\section{Supervised Learning}
\label{sec:supervised}

Supervised learning\index{supervised learning} refers to learning a functional relationship, or model, $\hat{f}:X \to Y$ between a set of $p$ features $\bm{x} \in X \subseteq \mathbb{R}^p$ and a target  $y \in Y \subseteq \mathbb{R}$ from data $\mathcal{D} = \{(\bm{x}^i,y^i)\}^n_{i=1}$ with $n \in \mathbb{N}$ instances. It is uniquely characterized by the use of labeled training data for learning the underlying relationship $f: X \to Y$. We refer to labeled data if both features and targets are observed. The model $\hat{f}$ is then used to make predictions for the target of new data $\hat{y} = \hat{f}(\bm{x})$, where the features but not the targets are available. An example prediction task for the field of epidemiology may be predicting the risk of a specific disease, based on genetic and lifestyle features. 

It is important to note that the term \textit{model}\index{model} is used for many different concepts in science, which may lead to confusion. In this chapter, the term \textit{model} is used to refer only to the functional relationship $\hat{f}$ between $\bm{x}$ and $y$. The algorithm that is used to find the model is termed \textit{inducer}\index{inducer} or \textit{learner}\index{learner}.

Generally, supervised learning with a continuous target $y \in \mathbb{R}$ is referred to as a \textit{regression}\index{regression} task. A \textit{classification}\index{classification} task is presented, when $y$ is categorical, i.e., $y \in \{1,\dots,C\}$, where $C \in \mathbb{N}$ is the number of classes. In this case, the prediction can either be categorical, i.e., $\hat{y} \in \{1,\dots,C\}$, or probabilistic in nature with $\hat{\pi} = P(y = c|\bm{x})$ for each class $c \in \{1,\dots,C\}$. For only two classes ($C=2$), typically a $\{0,1\}$ target is used, hence it is referred to as \emph{binary classification}, while $C>2$ is called \emph{multiclass classification}. 

For evaluation purposes, the difference between the predicted values $\hat{y}$ and the actual values $y$ is usually quantified in the form of a loss function\index{loss function} $L(\hat{y}, y)$. It measures the performance of a model and the goal is often to minimize this function during training to improve the model's accuracy, for further details see Sec.~\ref{sec:eval}.

\subsection{Tree-based Machine Learning Methods}
\label{sec:treebased}
As the name suggests, tree-based machine learning methods are based on so-called \emph{decision trees}\index{decision tree}. Decision trees are used in a variety of contexts apart from machine learning for the purpose of simple, rule-based decision-making.
In general, trees are said to reflect the human decision-making process well \citep{Gareth_intro_stat_learning_2021_short}. Although decision trees may seem straightforward or even trivial, they are indeed building blocks of powerful learners.
The following sections first explain classification and regression trees (CART) -- one of the major decision tree algorithms -- followed by so-called ensemble methods, which combine several decision trees. Finally, we give an example of a decision tree on the heart disease data introduced in Sec.~\ref{sec:intro}. 

\subsubsection{Classification and Regression Trees}

\begin{figure}
\centering
    \begin{subfigure}[b]{0.45\textwidth}     
        \centering
        \includegraphics[width=\textwidth]{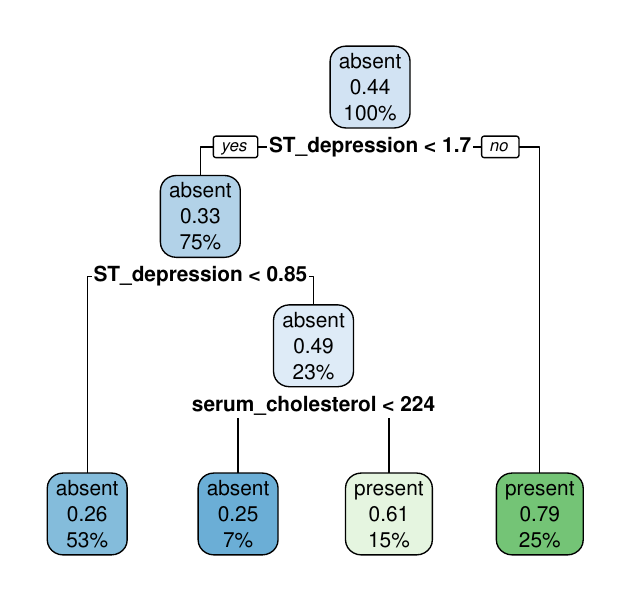}
        \caption{Decision tree with 2 features.}
    \end{subfigure}%
    \begin{subfigure}[b]{0.45\textwidth}
        \centering
        \includegraphics[width=\textwidth]{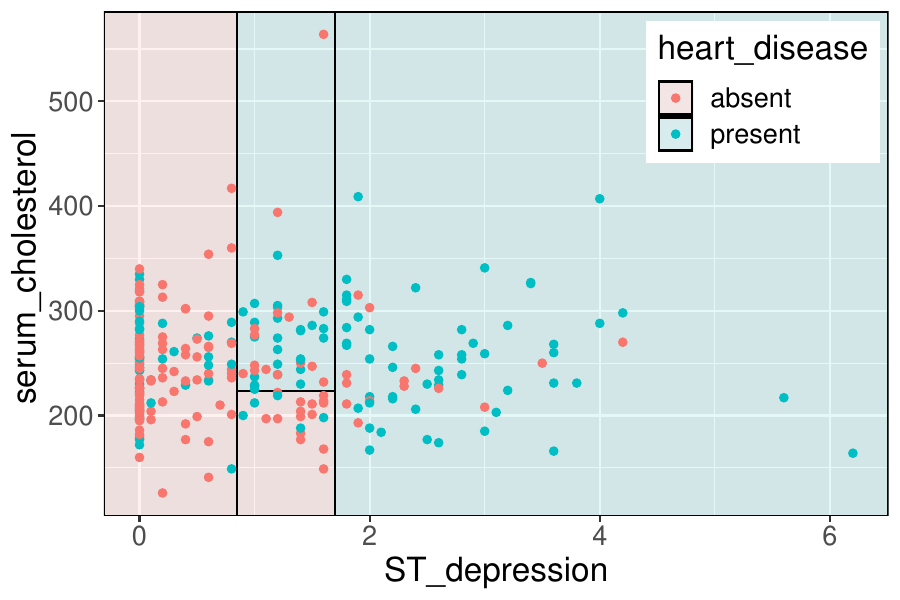}
        \caption{Partition plot.}
    \end{subfigure}
    \caption{(a) Decision tree of the heart disease dataset using only the features \texttt{ST\_depression} and \texttt{serum\_cholesterol}. (b) The corresponding partition plot. The points are the instances/patients, and the shaded areas denote the model predictions.}\label{fig:tree_2var}
\end{figure}

Classification and regression trees (CART) are constructed by recursively partitioning the instances of the training dataset into subgroups.
An example is shown in Fig.~\ref{fig:tree_2var}\,(a), showing a tree of seven nodes\index{node}, including a root node at the top of the tree, four leaf nodes at the bottom and two internal nodes in between.
The algorithm starts at the root node with the full training data and finds the best split with regard to a loss function. 
This results in two child nodes, which are then recursively split according to the same scheme, until a stopping criterion is reached.
In the resulting terminal nodes, or leaves\index{leaf}, a prediction is made based on a probability estimate or majority vote. Fig.~\ref{fig:tree_2var}\,(b) illustrates the corresponding partitioning of two exemplary features, where the shaded regions represent the binary predictions. 

More formally, the space $X$ is divided into $M$ disjoint regions $R_m$ with $m = 1, \dots, M$, corresponding to the leaves of the tree. In each region, a simple prediction model is fitted. In CART, a constant prediction $\hat{y}_m$ is used, i.e., for every instance in a particular region $\bm{x} \in R_m$, the same prediction $\hat{y}_m$ is made.
With this and $\Theta = \{ (R_m, \hat{y}_m)\}_{m=1}^M$, a tree model $\hat{f}_\text{tree}$ can be defined as 
\begin{align*}
    \hat{f}_\text{tree}(\bm{x};\Theta) = \sum_{m=1}^M \hat{y}_m \mathds{1}\left(\bm{x}\in R_m\right),
\end{align*}
where $\mathds{1}(\cdot)$ is the indicator function and $\Theta$ is fitted by minimizing
\begin{align*}
    \hat{\Theta} = \underset{\Theta}{\arg \min} \sum_{m=1}^M \sum_{i:\bm{x}^i \in R_m} L\left( \hat{y}_m, y^i \right). 
\end{align*}
That is, $\hat{\Theta}$ is chosen such that the target values of instances that fall into a region $R_m$ are close to the prediction $\hat{y}_m$ of the region, which is evaluated by a loss function $L$. The optimization problem can be broken down into two components -- finding the constants $\hat{y}_m$ and finding the regions $R_m$, which, however, depend on each other. 

Finding the constant $\hat{y}_m$ given a region $R_m$ can be specified by the input task. In a regression task, the constant is chosen as the arithmetic mean over all $y^i$ corresponding to the $n_m$ data instances falling in $R_m$. In contrast, for a classification task, the majority class is used. Let $p_{mk}=\frac{1}{n_m} \sum_{i:\bm{x}^i \in R_m} \mathds{1}(y^i =k)$ be the proportion of instances of class $k$ in $R_m$, then, $\hat{y}_{m} = \arg \max_k p_{mk}$. Several extensions have been proposed, e.g., for survival tasks, one could estimate the survival function with the Kaplan-Meier estimator in each leaf separately \citep{hothorn2004bagging}. 

To find the regions $R_m$, the algorithm builds the tree using recursive binary partitioning\index{recursive binary partitioning} or splitting. It starts at the root of the tree with all data and then chooses the best split into two partitions locally at each node. This splitting procedure follows the tree structure and does not consider the overall best split. Thus, it is a so-called top-down greedy approach\index{top-down greedy approach}. For a continuous or ordinal feature $\bm{x}_j$, the regions are divided by the split point $s\in\mathbb{R}$, such that instances with a feature value smaller or equal $s$ are assigned to the first, or left, child node and the remaining instances to the right child node:
\begin{align*}
    R_\text{left}(j,s) = \left\{ \bm{x} | \bm{x}_j \leq s \right\} \quad \text{and} \quad R_\text{right}(j,s) = \left\{ \bm{x} | \bm{x}_j > s \right\}. 
\end{align*}
The best split is found where, in both child nodes, the predicted target is close to the true target. To this end, we find the best pair $(j,s)$ for the split that minimizes the sum of the loss functions in the two child nodes:
\begin{align*}
    \underset{j,s}{\arg \min} \left(  \sum_{i:\bm{x}^i \in R_\text{left}(j,s)} L\left( \hat{y}_\text{left}, y^i\right) + \sum_{i:\bm{x}^i \in R_\text{right}(j,s)} L\left( \hat{y}_\text{right}, y^i \right) \right).
\end{align*}
For a categorical feature $\bm{x}_j$, we can assign a subset $S$ of its categories to the left child node and the remaining categories to the right child node:
\begin{align*}
    R_\text{left}(j,S) = \left\{ \bm{x} \left| \bm{x}_j \in S \right. \right\} \quad \text{and} \quad R_\text{right}(j,S) = \left\{ \bm{x} \left| \bm{x}_j \not \in S \right. \right\}.
\end{align*}
Note that we search for the best split $s$ or $S$ for all features $\bm{x}_j$ simultaneously.    

Depending on the task, different loss functions are chosen. For regression tasks, for instance, the mean squared error (MSE) is often utilized, whereas for classification tasks measures like cross-entropy or Gini coefficient are preferred. In principle and depending on the implementation, one could use any evaluation metric; see Sec.~\ref{sec:eval}, where evaluation metrics are discussed more generally and in more detail. In the context of decision trees, the concept of empirical risk or loss minimization is often used interchangeably with the concept of impurity reduction. For example, for a regression tree, minimizing MSE loss is equivalent to maximizing impurity reduction when the impurity is measured by the empirical variance.  

In theory, the described algorithm could be continued until no further split is possible. Intuitively, the deeper the tree, the more detailed the data are split into subsets. In the extreme case, each leaf contains only a single instance and the tree perfectly predicts the training data. However, in this case, the tree is probably fitted too well to the training data and for unseen data, the model could fail. In other words, the model overfitted (see Sec.~\ref{sec:eval}). 

Useful in the context of overfitting is the following perspective: Analogous to the above definition of a tree model $\hat{f}_\text{tree}$, a tree can also be represented via the sequence of splits, denoted by $T$. 
This intuitively allows for the definition of subtrees, which only contain a subset of splits until a certain iteration step, i.e., up to a certain tree depth. 

There are several techniques to avoid overfitting in trees. First, one typically defines a stopping criterion, such as a predefined minimum number of instances in a node or a maximum tree depth. Another option involves growing the tree as long as each split's reduction in the loss exceeds a predefined threshold. However, while this might perform better than a simple stopping criterion, it might still not result in an optimal tree due to the greedy nature of the algorithm. This is because a rather non-informative split could lead to a better split later on in the tree, i.e., a split that further reduces the loss. For this reason, the usual approach is to first grow a deep tree until a stopping criterion is reached and then \emph{prune} the tree, i.e., cut back the leaves.  

In \emph{cost-complexity pruning}\index{cost-complexity pruning}, or also called \emph{weakest link pruning}\index{weakest link pruning}, we want to find a tree with a good trade-off between high prediction performance and low model complexity. For a given tree $T$, the tree complexity can be measured by the number of leaves $|T|$, since more complex trees are deeper or wider, corresponding to a higher number of leaf nodes. Prediction performance can be measured by the loss function $L_m(T) = \sum_{i:\bm{x}^i \in R_m} L\left( \hat{y}_m, y^i \right)$ of a tree $T$ in the leaf $m = 1, \dots, |T|$, where a lower loss indicates a higher prediction performance. The cost complexity function for a given tree $T$ is given as 
\begin{align*}
    C_\alpha(T) = \sum_{m=1}^{|T|} n_m L_m(T) + \alpha |T|,
\end{align*}
where $n_m$ is the number of instances in leaf $m$ and $\alpha \geq 0$ is a predefined complexity parameter, which adjusts the trade-off between prediction performance and model complexity. In cost complexity pruning, we obtain the optimal subtree $T_\alpha \subseteq T$ among all possible subtrees by minimizing $C_\alpha$. 
Large $\alpha$ values indicate smaller trees and $\alpha = 0$ refers to the unpruned tree $T_0=T$. We obtain the best $\alpha$ by, for instance, cross-validation (Sec.~\ref{sec:eval}) and take the subtree $T_\alpha$ that minimizes the cross-validation error.    

The beauty of decision trees is that they are intuitive and easy to interpret. Yet, compared to other more sophisticated machine learning models, their prediction performance is often rather poor. Decision trees suffer from high variance, i.e., if we randomly sample the training dataset and fit a decision tree to each sample, we could get very different results. The good news is that if we combine many trees on such samples, the performance improves a lot. We will present some methods in the following that use this idea. 

\subsubsection{Bagging and Boosting}

To overcome the high variance of simple learners such as decision trees, \emph{ensemble} methods\index{ensemble method} combine many of such learners into a more powerful one. The simple learners in an ensemble are also called \emph{weak learners}\index{weak learner} since they would lead to mediocre performance on their own. The two most popular ensemble methods for decision trees are \emph{bagging} and \emph{boosting}, which will be described in the following. 

\textbf{Bootstrap aggregating (bagging)}\index{bagging} builds multiple models on bootstrapped datasets and aggregates the predictions of those models. The idea of the bootstrap stems from the following: If $Z_1,\dots, Z_B$ are $B$ independent and identically distributed random variables with variance $\sigma^2$, then the variance of the mean $\bar{Z}$ is $\sigma^2/B$. Analogously, if we had many independent training samples from a population and averaged the resulting prediction models, we would reduce the variance compared to only using one training sample. Since it is challenging and sometimes impossible to obtain multiple independent training samples, the idea of bootstrapping is to take repeated samples (with replacement) from the single training dataset we have. See Sec.~\ref{sec:Bootstrap} for more details on bootstrapping.
For bagging, first $B$ different bootstrap samples are generated from the training data and a prediction model, e.g., a decision tree, is fitted on each of the bootstrap samples. Let $\hat{f}_\text{tree}^b(\bm{x})$ be the prediction of the decision tree fitted on the $b$th bootstrap sample. Then the bagged prediction is defined as
\begin{align*}
    \hat{f}_{\text{bag}}(\bm{x}) = \frac{1}{B} \sum_{b=1}^B \hat{f}_\text{tree}^b(\bm{x}).
\end{align*}
For a classification task, one would, for example, take the majority vote among all the $B$ tree predictions. In contrast to single decision trees, the trees are typically grown deep, for each tree to have a high variance but low bias. Since averaging the trees reduces the variance, no additional pruning is required. The parameter $B$ can be chosen sufficiently large since it does not lead to overfitting. A crucial advantage of bagging is that, because of sampling with replacement, on average only $1 - \exp{(-1)} \approx 2/3$ of the instances are used for model fitting on each bootstrap sample. The remaining 1/3 of instances are called \emph{out-of-bag} (OOB). OOB instances are very useful when evaluating the generalization error, which is described in more detail in Sec.~\ref{sec:Bootstrap}. 

However, compared to a single tree, the improvement in prediction accuracy comes at the cost of interpretability (see Sec.~\ref{sec:iml}). 
Another disadvantage of bagging is that the trees grown on different bootstrap samples may still be quite similar, i.e., highly correlated, and consequently the effect of large variance reduction may vanish. Similar trees are often created when one or multiple strong features are present because they would be used for splits close to the root node in every tree. Mathematically speaking, if the random variables $Z_1,\dots, Z_B$ are identically distributed but not necessarily independent, the variance of their mean $\Bar{Z}$ is $\rho \sigma^2 + \frac{1-\rho}{B} \sigma^2$, where $\rho$ is the positive pairwise correlation. For $B \to \infty$, the variance reduces to the first term $\rho \sigma^2$. Therefore, the variance can be reduced, by reducing $\rho$, which leads to the idea of random forests. 

\textbf{Random forests}\index{random forest} \citep{breiman2001random} improve bagging by \emph{decorrelating} the trees, i.e., reducing the pairwise correlation $\rho$ and thereby the variance of the resulting ensemble. The method extends bagging by considering only a randomly sampled subset of size $q \leq p$ of all $p$ features for each split in a tree. Typical values for $q$ are $q\approx p/3$ or $q \approx \sqrt{p}$. The problem of having a few strong features described above is circumvented by this approach, since on average it is only considered for $(p-q)/p$ splits. These hyperparameters, the number of features considered for splitting at each node $q$ and the number of trees $B$, depend on the task but can be tuned (see Sec.~\ref{sec:tuning}). Generally, random forests have the advantage of relatively easy and fast training and require little tuning. Compared to single decision trees, the performance is often drastically improved and consequently, random forests are one of the most popular machine learning methods used in practice. 
 
\textbf{Boosting}\index{boosting} grows the trees sequentially on modified datasets, in contrast to the above methods, which grow trees in parallel. 
In boosting, the construction of each successive tree in the ensemble depends on the previous tree to use the knowledge gained from previous failures in an iterative learning process. Generally, in boosting, the ensemble consists of a weighted average of the individual trees: 
\begin{align*}
    \hat{f}_{\text{boost}}(\bm{x}) = \sum_{b=1}^B \omega_{b} \hat{f}_\text{tree}^b(\bm{x}),
\end{align*}
where $\omega_{b}$ are the weights and each tree $\hat{f}_\text{tree}^b(\bm{x})$ depends on the previous tree $\hat{f}_\text{tree}^{b-1}(\bm{x})$. 

\textbf{AdaBoost}\index{AdaBoost}, as introduced by \cite{freund1997decision}, uses stumps as weak learners, i.e., trees with only one split. In the first step, a stump is fitted to the original training dataset. Second, it is assessed for each instance whether it is correctly classified by the stump. In the next step, weights are assigned to all instances, whereby misclassified instances are assigned higher weights, while correctly classified instances are assigned lower weights. Then, another stump is fitted to the weighted dataset. By that, the second stump focuses on correcting the errors of the first stump. The algorithm proceeds iteratively, where each stump uses the weights obtained from the previous iteration. To obtain a final prediction, a weighted average of the individual stumps' predictions is calculated, where each stump is weighed depending on its performance. AdaBoost can be easily extended to other tasks such as regression and the use of deeper trees than just stumps.  

\textbf{Gradient boosting}\index{gradient boosting} \citep{friedman2001greedy} works in a similar iterative manner, but is based on the partial derivative, or gradient, of the loss function. For $L_2$ loss, the algorithm starts by fitting a single, typically not very deep, decision tree to the training data, obtaining a prediction $\hat{f}_\text{tree}^1(\bm{x})$. Second, the residuals $y - \hat{f}_\text{tree}^1(\bm{x})$ of this tree are calculated and used as the target to fit a second tree with prediction $\hat{f}_\text{tree}^2(\bm{x})$. The algorithm proceeds iteratively for $b=1,...,B$, where each tree is fitted to the residuals of the previous tree. Thereby, the algorithm puts its focus gradually on those parts, where the previous trees did not perform well, i.e., where the loss was large. Similarly to AdaBoost, the final prediction is a weighted average of the individual trees. However, for gradient boosting, the weights decrease in each iteration by a \emph{learning rate}, which is a hyperparameter that can be tuned. Note that the negative gradient of the $L_2$ loss is proportional to the residual vector. Gradient boosting can be generalized to other (differentiable) loss functions by using the gradient of the negative loss function instead of the residuals. 

A popular variant of gradient boosting is \textbf{extreme gradient boosting (XGBoost)}\index{XGBoost} \citep{xgboost_chen2016}. 
It extends gradient boosting by further regularization, e.g., by introducing an additional regularization term and shrinkage of tree weights, and column feature subsampling similar to random forests. XGBoost is computationally fast and often delivers state-of-the-art performance. As a result, it is as popular as random forests.

\subsubsection{Data Example}
\begin{figure}
    \centering
    \includegraphics[width=0.9\textwidth]{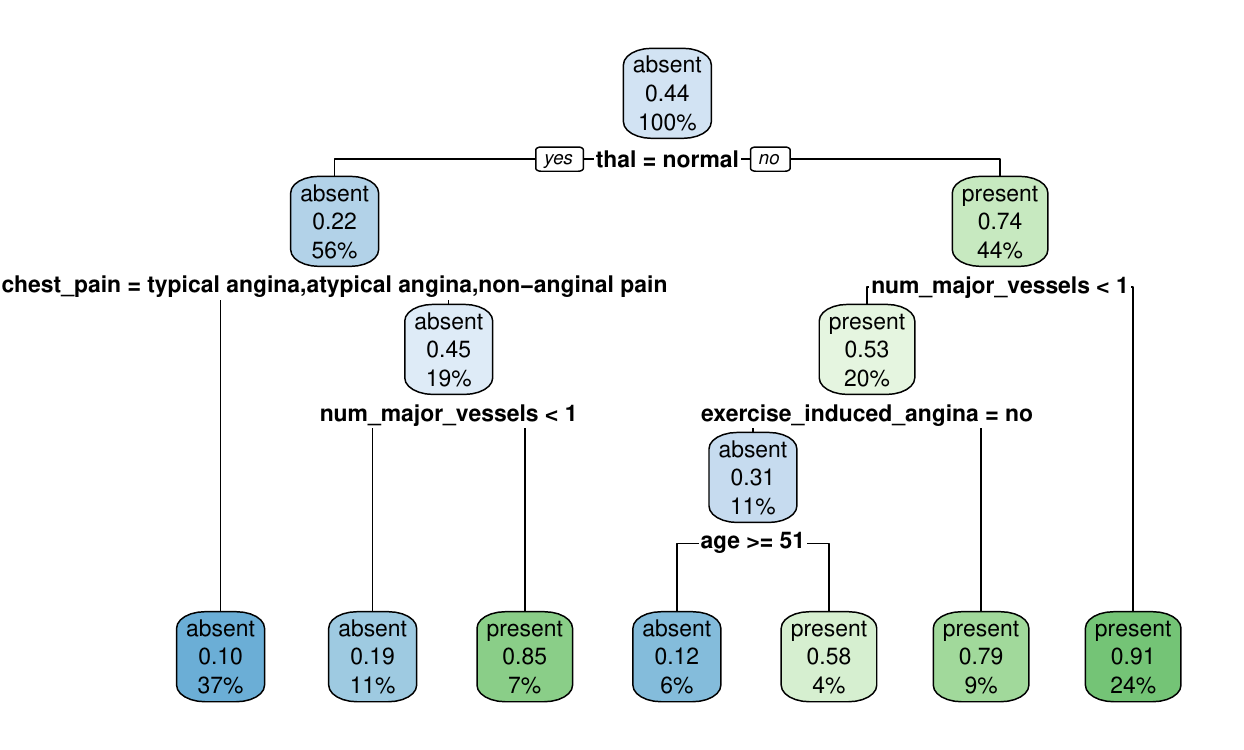}
    \caption{A decision tree fitted on the heart disease dataset, visualized with the \texttt{rpart.plot} package. It predicts whether heart disease is present or absent in a patient, based on the features in the dataset, e.g., the results of a thallium stress test (\texttt{thal}). Each node contains the following information: 1) absent/present prediction, 2) proportion of patients with present heart disease, 3) size of the node as percentage of total sample size. The splitting criteria are denoted below the nodes. If the criteria apply, we follow the left path, otherwise the right path.}
    \label{fig:Tree_all}
\end{figure}

Many software implementations are available for building decision trees. In \texttt{R}, we use the \texttt{rpart} package to fit a decision tree to the heart disease data described in Sec.~\ref{sec:intro}. In Fig.~\ref{fig:Tree_all}, we show the resulting tree.

\begin{verbatim} 
library(rpart)
library(rpart.plot)

tree <- rpart(heart_disease ~ ., heart)
rpart.plot(tree)
\end{verbatim}

Assume a doctor na\"{i}vely wants to predict whether to diagnose a new patient with heart disease on the basis of this tree. The doctor would then first perform a thallium stress test. If, for example, the results are not normal (right path), the next step is to check whether the number of major vessels colored by fluoroscopy is less than one. If that is not the case (right path), the doctor would diagnose a heart disease. 

As described in the algorithm above, regions are sequentially divided into smaller regions. This is made apparent by the information in the nodes. The root node in Fig.~\ref{fig:Tree_all}, for example, contains the following information: absent, 0.44, 100\%. This means that this node contains 100\% of the training data with a prevalence of 44\%. Since this is the minority, every new instance in this node would be classified as absent. The child node to the left contains all patients with a normal thallium stress test, which amounts to 56\% of the training data with a prevalence of 22\%. The remaining 44\% of the training data, i.e., those with a non-normal thallium stress test, are in the right child node, where heart disease is present for 74\% of patients. Further down in the tree, the nodes become smaller but improve in purity, i.e., contain mostly patients with or without heart disease.

\begin{figure}[t]
    \centering
    \includegraphics[width=0.9\textwidth]{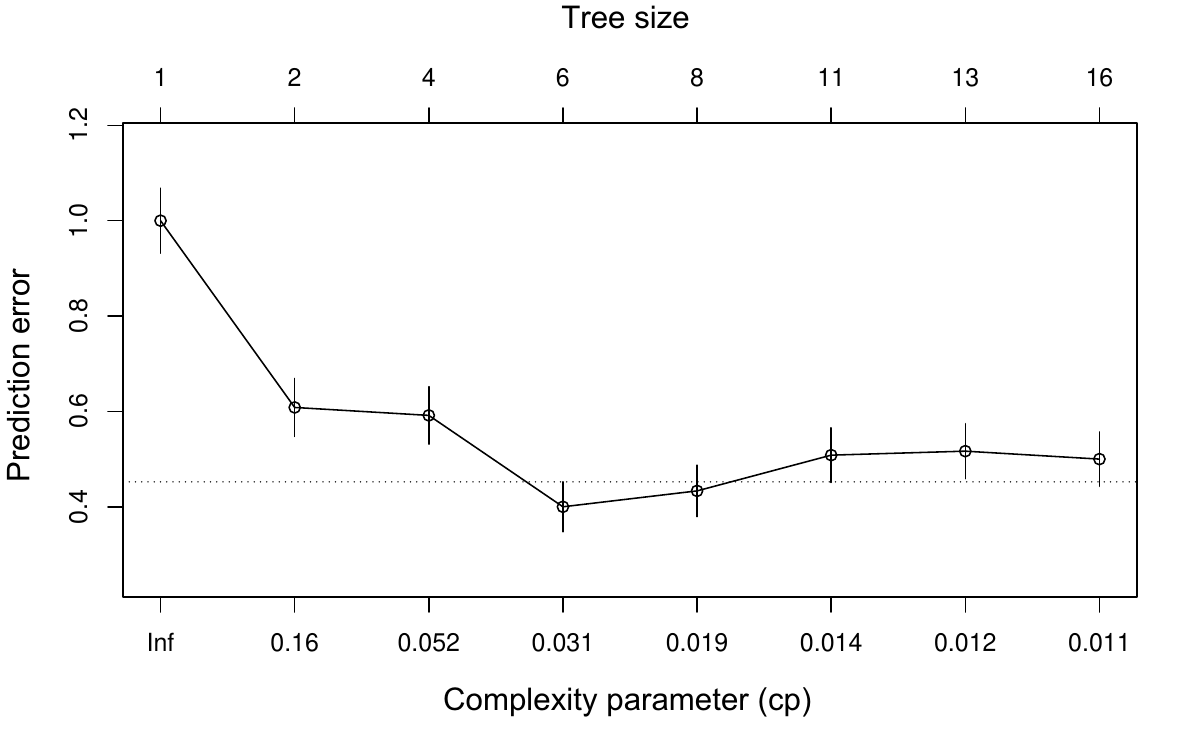}
    \caption{A visual representation of the cross-validation results of the cost-complexity pruning. The $x$-axis shows the complexity parameter \texttt{cp}. At the top, the number of leaf nodes corresponding to each complexity parameter is denoted. The $y$-axis represents a prediction error measure, based on cross-validation. The lowest error is reached at \texttt{cp} of 0.031. Figure created with the \texttt{rpart} package and slightly modified.}
    \label{fig:Tree_alpha}
\end{figure}

To see if we can improve upon the tree above, we now fit a deeper tree and perform cost-complexity pruning. A deeper tree can be created by reducing the minimum number of instances in a node to perform splitting: 
{\samepage
\begin{verbatim}
    tree_deep <- rpart(heart_disease ~ ., heart, 
                       control = rpart.control(minsplit = 2))
\end{verbatim}
}
In our example, this results in a tree of size (number of leaves) 16. We can plot an estimate of the generalization error (based on cross-validation) against the tree size by 
\begin{verbatim}
    plotcp(tree_deep)
\end{verbatim}
which outputs Fig.~\ref{fig:Tree_alpha}. 
We observe that the optimal tree has six leaves, which corresponds to a complexity parameter (denoted as $\alpha$ before) \verb+cp+ of 0.031. Hence, we prune the tree by 
\begin{verbatim}
    pruned_tree <- prune(tree_deep, cp = 0.031)
\end{verbatim}
which results in a tree as in Fig.~\ref{fig:Tree_all} where we prune back by cutting the last split at \texttt{age} $\geq 51$. 

\subsection{Artificial Neural Networks}
\label{sec:neuralnets}

Artificial neural networks, also known just as \emph{neural networks}\index{neural network}, are the fundamental component of the machine learning subfield of deep learning \citep{lecun2015deep,goodfellow_2016}. Neural networks are characterized by an extraordinarily flexible and parameter-rich transformation. This enables them to learn highly complex and non-linear functional relationships between features and target variables. However, due to the often extremely high number of model parameters, a sophisticated optimization procedure and regularization techniques are essential to achieve good performance on the data used for training as well as to generalize on unseen data, i.e., preventing overfitting. In the following sections, we discuss the possible architectures of neural networks and how their flexibility is achieved. In addition, we explore the training and regularization process of these models and conclude with an example of constructing and training a neural network in \texttt{R} using the heart disease dataset.

\subsubsection{Architecture}
\label{sec:nn_architecture}

\begin{figure}[t]
    \centering
    \includegraphics[width=0.9\textwidth]{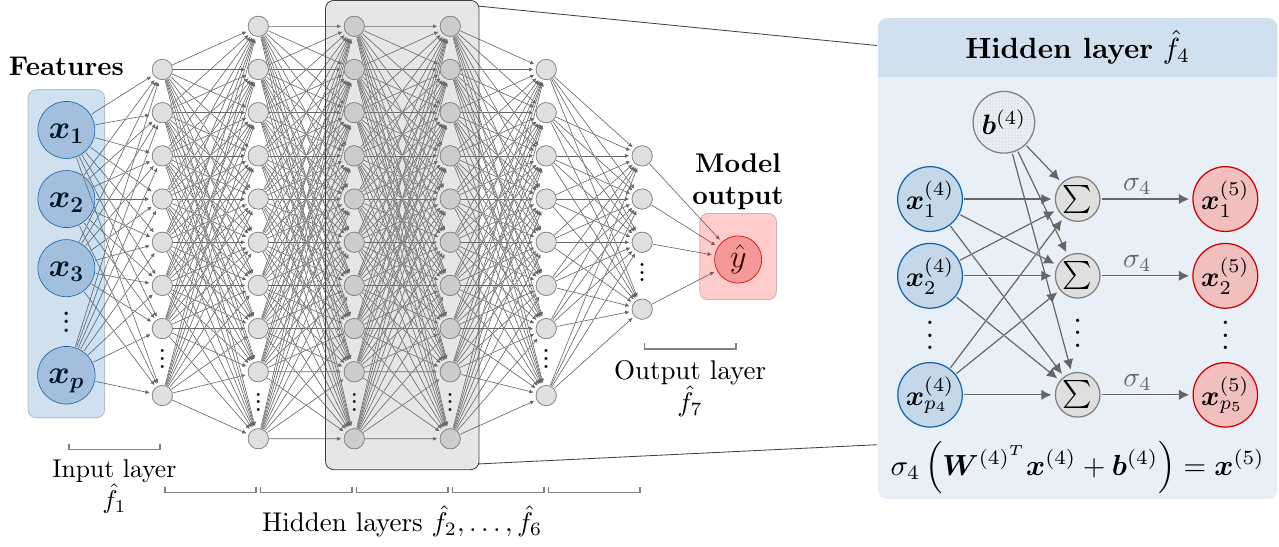}
    \caption{Model architecture of a sequential neural network with seven dense layers, generating predictions $\hat{y}$ from the input $\bm{x}$. On the right, the model's fourth layer is magnified, mapping the previous layer's output $\bm{x}^{(4)}$ to the next layer's input $\bm{x}^{(5)}$ through an affine transformation followed by a non-linear function, denoted as $\sigma_4$.}
    \label{fig:supervised_nn}
\end{figure}

In the context of neural networks, the basic model is commonly described as an \emph{architecture}\index{neural network architecture}. This term is appropriate because the user acts as an architect equipped with an extensive toolbox of modules and hyperparameters to construct a model tailored to their individual problem. The fundamental building blocks are the \emph{layers}\index{layer}, which essentially constitute simple parameterized non-linear transformations. The key to the flexibility of neural networks is that multiple layers $\hat{f}_1, \ldots, \hat{f}_K$ can be combined as long as their dimensions align, resulting in a highly powerful transformation as a model \citep{goodfellow_2016}. In most of the standard cases, the layers are arranged sequentially leading to the following mathematical representation with a single input layer $\hat{f}_1$  and output layer $\hat{f}_K$ depending on the features and type of target:
\begin{align*}
    \hat{f}_\text{NN}(\bm{x}) = \hat{f}_K \circ \hat{f}_{K-1} \circ \ldots \hat{f}_1(\bm{x}) = \hat{y}, \quad \bm{x} \in \mathbb{R}^p.
\end{align*}
The layers located between a neural network's input and output layer are commonly referred to as \emph{hidden layers}\index{hidden layer}, i.e., the layers $\hat{f}_2$, \ldots, $\hat{f}_{K-1}$ for the sequential model $\hat{f}_\text{NN}$ defined above. As an example, we illustrate a sequential architecture with $K=7$ layers in Fig.~\ref{fig:supervised_nn}.

Although increasingly complex types of layers have been invented in recent years, they all more or less rely on the same basic principles. This is captured by the earliest and most fundamental layer type of neural network architecture, known as the \emph{dense}\index{dense layer} or \emph{fully connected}\index{fully connected layer} layer. The layer's name and commonly used visualization, e.g., as shown in Fig.~\ref{fig:supervised_nn}, have significantly influenced the term "neural networks" and its association with the human brain. A dense layer connects each output variable of the preceding layer with every input of the subsequent layer, thus weaving a dense net of connections, in which the variables are frequently referred to as \emph{neurons}\index{neuron}. In addition, each connection contains a weight to regulate the influence of the corresponding neuron. The output neurons are then computed as the sum of the weighted input neurons, followed by a pointwise non-linear \emph{activation function}\index{activation function}. From a mathematical perspective, this layer type can be described as an affine transformation followed by a pointwise non-linearity. To be more precise, for a layer $\hat{f}_k:\mathbb{R}^{d_1} \to \mathbb{R}^{d_2}$, the layer's input $\bm{x}^{(k)}$ is multiplied by a weight matrix $\bm{W} \in \mathbb{R}^{d_2 \times d_1}$, then shifted by a vector $\bm{b} \in \mathbb{R}^{d_2}$, and afterwards passed to a pointwise non-linear function $\sigma_k$, resulting in the input for the succeeding layer $\bm{x}^{(k+1)}$, i.e.,
\begin{align*}
    \hat{f}_k\left(\bm{x}^{(k)}\right) = \sigma_k \left(\bm{W}^T\bm{x}^{(k)} + \bm{b} \right) = \bm{x}^{(k + 1)}.
\end{align*}

\begin{figure}[h]
    \centering
        \begin{subfigure}[b]{0.42\textwidth}
            \includegraphics[width=\textwidth]{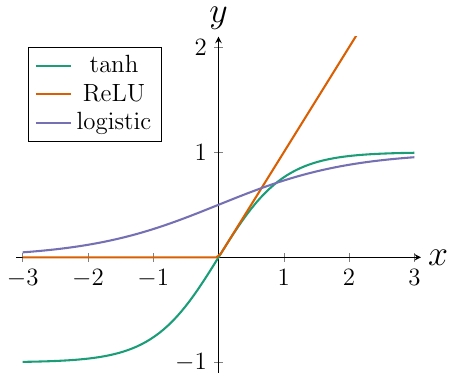}
            \caption{Activation functions}
            \label{fig:supervised_activations}
        \end{subfigure}%
        \hspace{0.5cm}
        \begin{subfigure}[b]{0.42\textwidth}
            \includegraphics[width=\textwidth]{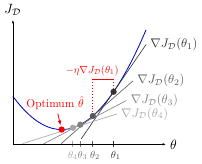}
              \caption{Gradient descent}
              \label{fig:supervised_sgd}
        \end{subfigure}
        \caption{(a) The graphs of typical activation functions: hyperbolic tangent (green), rectified linear unit (ReLU) (orange), and logistic function (blue) used for probability outcomes. (b) Illustration of the gradient descent technique for learning the optimum $\bm{\hat{\theta}}$ by iteratively updating the current parameter $\bm{\theta}_t$ for $-\eta \nabla J_\mathcal{D}(\bm{\theta}_t)$ units based on the negated tangent's slope at $\bm{\theta}_t$ on the loss function and the learning rate $\eta$.}
\end{figure}

While a dense layer in a neural network bears a resemblance to a generalized linear model, the terminology used in deep learning can be initially confusing for people with a statistical background. For example, the shift vector $\bm{b}$, which is known as the intercept $\bm{\beta}_0$ in linear regression, is referred to as the bias vector in the context of neural networks, and the statistical link function $\sigma$ (not to be confused with the standard deviation) is called the activation function. Moreover, the weights in neural networks are similar to the coefficients $\bm{\beta}$ in generalized linear models but do not have the same interpretation due to the layer-wise stacking. Initially, the hyperbolic tangent was predominantly employed as activation function. However, with the rise of deep learning, the \emph{rectified linear unit}\index{rectified linear unit} (ReLU) has replaced it almost entirely, allowing the elimination of unnecessary information for further forward propagation efficiently. Fig.~\ref{fig:supervised_activations} shows the graphs of three typical pointwise activation functions. The choice of activation function is once again left to the user as a hyperparameter. 

Aside from the activations in the hidden layers, the activation function in the final layer of a neural network plays a crucial role. For example, in regression tasks, the activation function should be capable of mapping to any likely real-valued number. In contrast, the activation should output probabilities ranging between 0 and 1 in classification tasks. Thus, an appropriate activation function for the final layer must be selected based on the nature of the task at hand:
\begin{itemize}
    \item For \textbf{regression tasks}, the neural network should be capable of learning any continuous real-valued target variable. Therefore, an activation function like ReLU, which cuts off all negative values, would not be a suitable choice. In such cases, the final layer uses a linear activation, meaning no activation function is applied.
    \item In a \textbf{binary classification task}, the target variable takes on values between 0 and 1, representing probabilities. In this case, the logistic function is used to constrain the prediction in the desired range (see Fig.~\ref{fig:supervised_activations}). While, mathematically, the logistic function belongs to the class of sigmoid functions\index{sigmoid function}, the terms "logistic" and "sigmoid" are mostly used interchangeably for the logistic function in the context of deep learning.
    \item In \textbf{classification tasks} involving more than two classes, the \emph{softmax}\index{softmax function} function is usually used. This vector-valued function converts a vector of real numbers to class-specific probabilities that sum up to one. This makes it a suitable choice for multiclass classification problems. 
\end{itemize}

So far, we have only introduced the fundamental dense layers, but there are many other types of layers that build upon this basic idea and are used in different applications. For example, \emph{convolutional}\index{convolutional layer} layers \citep{krizhevsky_2012, lecun2015deep} are the most popular choice in image processing and computer vision tasks. They learn to detect relevant patterns in an image, using a kernel operating similarly to a dense layer, but running in a sliding-window fashion over the image. Typically, convolutional layers are combined with pooling layers to drastically decrease the intermediate values' dimensions and, thereby, enhance the computational efficiency of the model. In recent years, \emph{residual}\index{residual layer} layers \citep{he_2016} have also been used in this context to obtain meaningful training impulses despite having a large number of layers. They address the vanishing gradient problem, which refers to the gradients becoming very small and less informative with increasing depth of the network, as described in Sec.~\ref{sec:nn_training}. Residual layers solve this issue by skip connections, which store earlier intermediate values and add them back at a later point. Furthermore, \emph{embedding}\index{embedding layer} layers \citep{bengio_2003} are another commonly used type of layer and the preferred choice in natural language processing or when dealing with discrete or factor-rich categorical tabular data. These trainable look-up tables transform discrete or categorical features into a compact and real-valued representation, avoiding the high dimensionality that would result from traditional encoding approaches. Usually, embedding layers are only used as input layers and are connected to a following dense network. Due to the efficient dimension reduction of large amounts of discrete or categorical features, they improve the network's performance. However, they are also recently applied in so-called self-attention modules as hidden layers within a transformer architecture (see Sec.~\ref{sec:unsupervised_generative_methods} for details). Another layer class commonly used for natural language processing tasks, time series analysis, and other input data where the order and context are crucial for accurate predictions is a \emph{recurrent}\index{recurrent layer} layer \citep{rumelhart1985RNN}. A recurrent neural network consisting of multiple recurrent layers is designed to process sequential data by maintaining hidden states that capture information from previous steps. However, they are hard to train due to the vanishing gradient problem. This issue is addressed by \emph{long short-term memory} (LSTM) layers\index{long short-term memory layer} \citep{hochreiter1997LSTM}.
These layers incorporate a memory cell and gating technique, allowing them to selectively retain and forget information over multiple time steps. This memory effect makes LSTMs particularly effective in capturing long-term dependencies in sequential data.

\subsubsection{Training of a Neural Network}
\label{sec:nn_training}

The most challenging aspect of neural networks is not only designing the model architecture. Rather, it is finding the appropriate values for the thousands to millions of parameters for the resulting model to closely approximate the underlying data-generating process. For example, a model with ten input variables for the features, one output variable for the target, and two hidden dense layers with 100 neurons each, has a total of 11,301 parameters that can take on arbitrary real values. This mammoth task is accomplished through a complex optimization procedure, typically referred to as learning or training in machine learning. The centerpiece of this process is a loss function that evaluates how well the model's predictions match the actual target values in the training data. For regression tasks, for example, the mean squared error (MSE), and for classification problems, binary or categorical cross-entropy are commonly used, which are explained in detail in Sec.~\ref{sec:eval-measures}. Regardless of the specific choice of the loss function $L$, it provides a real-valued metric of the model's performance, taking into account the actual target values $y^{i}$ of the training data $\mathcal{D} = \left\{(\bm{x}^{i}, y^{i}) \right\}_{i=1}^n$. Therefore, the fundamental optimization goal during the training process for a neural network model $\hat{f}_\text{NN}(\cdot\,; \bm{\theta})$ with parameters $\bm{\theta}$ is to minimize the loss function for all instances in $\mathcal{D}$, which is also referred to as empirical risk minimization:
\begin{align}\label{eq:supervised_nn_erm}
    \bm{\hat{\theta}} = \underset{\bm{\theta}}{\arg \min}\ \frac{1}{n} \sum_{i=1}^n L\left(\hat{f}_{\text{NN}} \left(\bm{x}^{i};\bm{\theta} \right), y^{i} \right) = \underset{\bm{\theta}}{\arg \min}\ J_{\mathcal{D}}(\bm{\theta}).
\end{align}
The term training loss $J_{\mathcal{D}}(\bm{\theta})$ is commonly used to refer to the average value of the instance-wise loss function over the training dataset $\mathcal{D}$, i.e., $J_{\mathcal{D}}(\bm{\theta}) = \frac{1}{n} \sum_{i=1}^n L\left(\hat{f}_\text{NN}(\bm{x}^{i};\bm{\theta}), y^{i} \right)$.

In order to receive optimal parameter values for the problem in Eq.~\eqref{eq:supervised_nn_erm}, neural networks leverage their partial differentiability combined with the chain rule to compute the model parameters' gradients efficiently. The technique for efficient and fast gradient calculation is the famous \emph{(error) backpropagation}\index{backpropagation} algorithm and one of the crucial reasons why training neural networks with millions of parameters becomes feasible. The whole optimization procedure applies an iterative process known as \emph{gradient descent}\index{gradient descent}, which updates the parameters of the neural network in the opposite direction of the gradient of the training loss until a minimum is reached (see Fig.~\ref{fig:supervised_sgd}). The gradient with respect to the parameters $\bm{\theta}$, denoted as $\nabla_{\bm{\theta}}$, provides information about the tendency of the steepest local descent, which points in the direction of the most rapidly decreasing training loss. During training, we update the parameters by taking small steps in this direction, multiplied by a small value known as the learning rate $\eta \in \mathbb{R}^+$, i.e.,
\begin{align*}
    \bm{\theta}_{t+1} \leftarrow \bm{\theta}_t - \eta\ \nabla_{\bm{\theta}_t} J_{\mathcal{D}}(\bm{\theta}_t).
\end{align*}
In practice, the variant \emph{stochastic gradient descent}\index{stochastic gradient descent} (SGD) is applied, which randomly samples a mini-batch $\tilde{\mathcal{D}} \subseteq \mathcal{D}$ of training instances and then computes the gradient of the training loss with respect to the parameters. This batch-wise evaluation is more computationally efficient and less memory intensive than calculating the gradient over the entire training set $\mathcal{D}$. Moreover, adding more randomness also prevents the optimization procedure from overfitting or getting stuck in local minima \cite[for more information on optimization procedures, see][]{bottou2018}. This optimization step is then usually repeated for several iterations, also known as epochs\index{epoch}, and terminates when the loss value no longer decreases or when a predefined maximum number of epochs is reached. The mini-batch size\index{mini-batch size} and the learning rate\index{learning rate} are additional hyperparameters that need to be manually tuned for the optimization procedure. There are also several other extensions of the SGD optimizer, which incorporate a momentum or use a parameter-adaptive learning rate, e.g., AdaGrad or Adam \citep{kingma2014}. 

In addition to the raw optimization process, \textbf{regularization}\index{regularization} methods play a vital role in successful model training. They accelerate the overall training procedure and lead to a more robust and generalized model that is less prone to overfitting. Typically, there are three areas where we can apply regularization methods: First, preprocessing the input data can be beneficial during training. This involves standardizing or normalizing the data, such as scaling numerical features to lie between 0 and 1 or normalizing them based on the dataset's mean and variance. In particular, \emph{augmentations}\index{augmentation} are recommended for image data, which increase the training data diversity through random rotations, added noise, or cropping. Regularization methods also touch the model's architecture: Additional layers like \emph{dropout layer}\index{dropout layer} or \emph{batch normalization}\index{batch normalization} can be inserted between the regular layers. A dropout layer randomly sets a fraction of neurons to zero during training, while batch normalization normalizes intermediate values by their mean and variance. Lastly, regularization can also be applied during the optimization process: For instance, $L_1$ or $L_2$ regularization terms can be incorporated into the loss function to keep the model's parameters small, where $L_2$ regularization is usually called weight decay in the deep learning context. From a statistical point of view, this corresponds to \emph{least absolute shrinkage and selection operator} (LASSO) and \emph{ridge} regression. Moreover, \emph{early stopping}\index{early stopping} or a \emph{learning rate scheduler}\index{learning rate scheduler} can be employed when the model's improvement plateaus or starts to overfit, typically monitored using validation data (see Sec.~\ref{sec:tuning} for details). For a more detailed overview of the most common regularization techniques, we refer to \cite{goodfellow_2016}.

\subsubsection{Data Example}
\label{sec:nn_example}

As a data example, we train a neural network with dense layers and ReLU activations on the heart disease dataset using the \texttt{keras} \texttt{R} package \citep{keras_R}. The architecture comprises three layers, with a dropout layer added between each of the three dense layers for regularization, using a dropout rate of 0.4. Since this task involves binary classification, we employ the logistic function -- in \texttt{keras} denoted as \texttt{sigmoid} -- as the final activation to obtain the probability of a patient being diagnosed with a heart disease. Furthermore, after applying one-hot encoding\footnote{One-hot encoding converts categorical features into a numerical representation analogous to dummy encoding but with $N$ instead of $N -1$ features for $N$ categories, i.e., dummy encoding without a reference category.}, the dataset consists of 22 features. Therefore, we pass this number to the argument \texttt{input\_shape} in our sequential model. Next, we compile the model, using Adam as the optimizer with a learning rate of 0.002 and binary cross-entropy as the loss function. Additionally, we include accuracy as another metric, which will be explained in more detail in Sec.~\ref{sec:eval-measures}. We proceed to train the model on the heart disease dataset for 50 epochs, using a mini-batch size of 32 instances. We allocate 20\% of the training data for validation purposes and stop the training if the loss on the validation data does not improve for ten epochs, i.e., early stopping. With this approach, we achieve an accuracy of 100\% on the test data. These steps are summarized in the following \texttt{R} code:
%
\begin{verbatim}
library(keras)

# Build model architecture
model <- keras_model_sequential(input_shape = c(22)) %>%
  layer_dense(units = 64, activation = "relu") %>%
  layer_dropout(rate = 0.4) %>%
  layer_dense(units = 32, activation = "relu") %>%
  layer_dropout(rate = 0.4) %>%
  layer_dense(units = 1, activation = "sigmoid")
  
# Compile model
model %>% compile(
  optimizer = optimizer_adam(learning_rate = 0.002),
  loss = "binary_crossentropy",
  metrics = "accuracy")
  
# Train model
model %>% fit(train_x, train_y,
  epochs = 50, batch_size = 32, validation_split = 0.2,
  callbacks = callback_early_stopping(
    patience = 10, restore_best_weights = TRUE))
\end{verbatim}

\section{Model Evaluation and Resampling}
\label{sec:eval}

A central step in the machine learning methodology is the evaluation of one or more models based on a suitable metric while making efficient use of the usually limited data available.
It is crucial to determine what constitutes an appropriate metric for evaluating model performance. Relying solely on high accuracy as a measure of a classification model's efficacy may not suffice. Additionally, it is important to ascertain whether accuracy measurements pertain to the same dataset used for model training or extend to the evaluation of model performance on previously unseen data instances.
There are numerous approaches to model evaluation, therefore it is infeasible to comprehensively explain every possible pitfall here. This section aims to give a concise overview of common terminology and the basic approach to model evaluation. Sec.~\ref{sec:eval-measures} introduces the most common evaluation metrics for continuous and binary outcomes. Sec.~\ref{sec:resampling} explains the concept of estimating the generalization performance. Within this section, several methods of resampling are presented, consisting of cross-validation in Sec.~\ref{sec:crossvalidation}, subsampling in Sec.~\ref{sec:subsampling} and bootstrapping in Sec.~\ref{sec:Bootstrap}. 
For further information on model evaluation, see e.g., \cite{japkowicz2011evaluating} for a thorough overview in classification settings.
On the topic of resampling strategies in particular, \cite{bischl2012resamplingmethodsa} provide concrete recommendations.
\cite{gerds2021medicalrisk} additionally provide a thorough overview with a focus on medical settings.

\subsection{Evaluation Metrics}
\label{sec:eval-measures}


In general, an evaluation metric (or measure)\index{evaluation metric}\index{evaluation measure} quantifies the difference between the true target $y$ and the predicted value $\hat{y}$.
In linear regression, this is a straightforward choice: Most frequently the \emph{mean squared error}\index{mean squared error} (MSE) or in some cases the \emph{mean absolute error}\index{mean absolute error} (MAE) are used, which measure the squared or absolute distance, respectively. 
In machine learning terminology, both MAE and the more common MSE are examples of  \emph{empirical risk}\index{empirical risk} functions  $\frac{1}{n} \sum_{i=1}^n L(\hat{y}^i, y^i)$, which aggregate the respective loss functions 
\begin{align*}
    L_1\left(\hat{y}^i, y^i\right) = \left|\hat{y}^i-y^i\right| \quad \text{and} \quad L_2\left(\hat{y}^i, y^i\right) = \left(\hat{y}^i-y^i\right)^2.
\end{align*}

In statistics, the modeling process is typically framed in terms of maximizing likelihood given the observed data.
Machine learning, in contrast, tends to approach this matter from the opposing direction: An error or risk function\index{risk function} that depends on the data and the parameters of the learning algorithm is the key element. The objective is to minimize this risk through the model fitting (or training) process.

The estimation of this risk shares commonalities with the estimation of other statistical quantities, exhibiting a trade-off between bias and variance\index{bias-variance tradeoff} of the estimate.
We will take a closer look at this phenomenon in the following section. For the present, it is sufficient to acknowledge that a model is conventionally trained on a \emph{training set}\index{training set} and evaluated on an independent sample that was not part of the training process, referred to as the \emph{test set}\index{test set}.

The first step in any model evaluation is to choose a suitable performance metric.
This is typically determined by the type of task, e.g., regression or classification, and the specific criteria of the prediction problem.
In the regression example, the choice is between MSE and MAE depending on whether robustness needs to be emphasized, whereas in binary classification there are a multitude of different measures, which emphasize different aspects of predictive performance.
In the following, we will introduce some common metrics for classification tasks.

\subsubsection{Binary Classification Measures}

We begin this section with a motivating example for disease diagnosis, underscoring the necessity of transcending the concept of \emph{classification accuracy}\index{classification accuracy} when assessing binary classification predictions: If model A predicts the correct diagnostic status of 80\% of patients, and model B is correct for 90\% of patients, then model B, at first intuition, seems to be the superior model. Yet, it could be possible that the dataset consists of only ten patients in total, with nine healthy patients ($y = 0$) and one diseased patient ($y = 1$), for whom model A predicts a probability of $\hat{\pi} = 0.49$ and model B predicts $\hat{\pi} = 0.51$. In this scenario, it does not seem appropriate to declare one model better than the other, purely based on predictive accuracy as defined by the proportion of correct classifications. Evidently, accuracy and its complement, the \emph{classification error}\index{classification error} (CE), serve as valuable initial metrics. However, it is imperative to recognize that they provide an incomplete picture when it comes to the comprehensive evaluation of a classification model.

\begin{figure}
\centering
  \includegraphics[width=.6\linewidth]{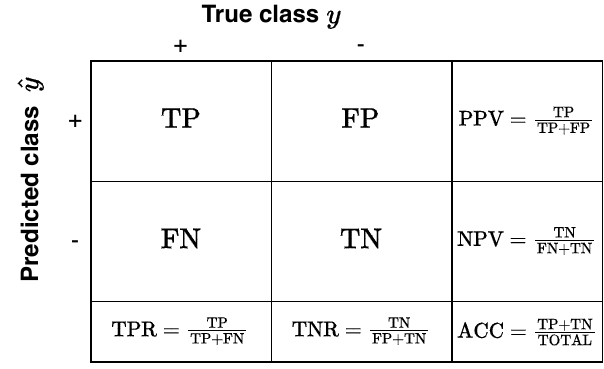}
  \caption{A confusion matrix showing the categorization of true and predicted class labels into true positives (TP), false positives (FP), false negatives (FN) and true negatives (TN). Common statistics derived from these values are the positive predictive value (PPV) and the negative predictive value (NPV), the accuracy (ACC) and its complement, the classification error (CE), as well as the true positive rate (TPR) and true negative rate (TNR).}
  \label{fig:confmat}
\end{figure}

The same principles employed in diagnostic testing can be applied, utilizing various measures constructable around a \emph{confusion matrix}\index{confusion matrix}.
Notably though, machine learning has historically used terminology deviating from terminology, e.g., familiar from medical settings.
While the concept of true and false positives and negatives remains the same (see Fig.~\ref{fig:confmat}), we note some commonly used measures and their more common aliases in machine learning literature:

\begin{itemize}
    \item{Sensitivity}\index{sensitivity} or \emph{recall}\index{recall} or true positive rate (TPR)\index{true positive rante}
    \item{Specificity}\index{specificity} or true negative rate (TNR)\index{true negative rate}
    \item{Positive predictive value}\index{positive predictive value} (PPV) or \emph{precision}\index{precision}
\end{itemize}

Regardless of how the correct and incorrect classifications are related, however, the actual classifications themselves depend on the threshold applied to the probability prediction of the learner, i.e., a value of 0.51 may either be considered large enough to warrant this a prediction of the positive class, or a higher (or even lower) threshold may be applied.
To analyze the behavior of a classification model across different thresholds, the \emph{receiver operating characteristic}\index{receiver operating characteristic curve} (ROC) curve is a commonly used tool.

\begin{figure}[htb]
\centering
  \includegraphics[width=.81\linewidth]{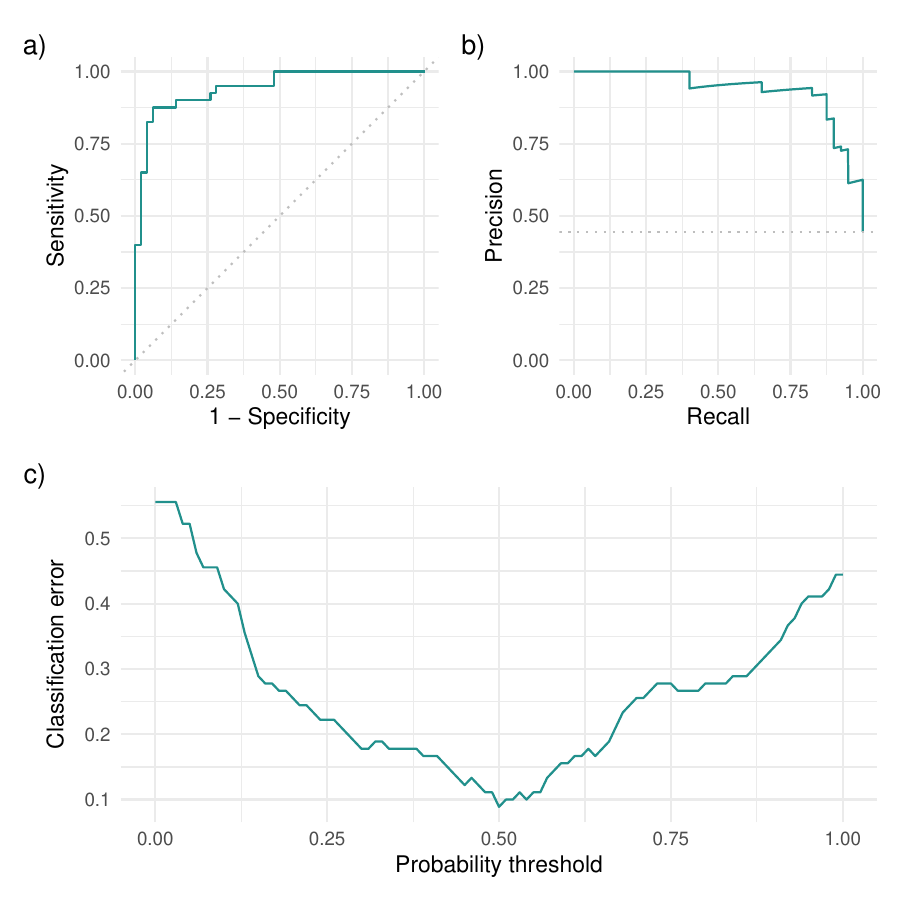}
  \caption{ROC curve (a), precision-recall curve (b), and threshold curve (c) of a random forest classifier applied to the heart disease task as generated by \texttt{mlr3}.
  Depending on the task, each visualization can provide valuable insight into the model at hand and provide necessary context to the model evaluation process.}
  \label{fig:evalcurves}
\end{figure}

Fig.~\ref{fig:evalcurves}\,(a) visualizes sensitivity (or TPR) on the $y$-axis with the false positive rate (FPR, also $1 - \text{specificity}$) on the $x$-axis, offering a more comprehensive representation of the classifier's performance than a basic accuracy score.
If the curve corresponding to a classifier resides within the upper-left quadrant of the plot, the classifier's performance is not inferior to random chance (i.e., a coin flip).
For the purpose of extracting a quantitative measure of prediction quality from the ROC curve, the \emph{area under the curve}\index{area under the curve} (AUC) can be computed.
The resulting value will be 0.5 for a random classifier and 1 for a perfect classifier, regardless of the threshold.
An alternative visualization is the \emph{precision-recall curve}\index{precision-recall curve} in Fig.~\ref{fig:evalcurves}\,(b), which is functionally similar to the ROC curve but plotting precision (PPV) against recall (TPR).
A sometimes overlooked yet often informative visualization is the \emph{threshold curve}\index{threshold curve}, which relates the classification error to the classification threshold (Fig.~\ref{fig:evalcurves}\,(c)). 

The code below illustrates how to create the graphics in Fig.~\ref{fig:evalcurves} using \texttt{mlr3}, including displaying a confusion matrix and computing associated measures:

\begin{verbatim}
# Choose a learner (random forest)
learner <- lrn("classif.ranger", predict_type = "prob")
# Partition data into train/test sets
split <- partition(heart_task)
# Train on the training data
learner$train(heart_task, row_ids = split$train)

# Predict on the test data
pred <- learner$predict(heart_task, row_ids = split$test)
# Retrieve the confusion matrix for these predictions
pred$confusion
#>         truth
#>response  present absent
#>  present      35      3
#>  absent        5     47

# Enumerate mlr3 measures to use for evaluation
msr_ids <- c("classif.tpr", "classif.tnr", 
  "classif.ppv", "classif.npv", "classif.acc")

# Score predictions accordingly
pred$score(msrs(msr_ids))
#> classif.tpr classif.tnr classif.ppv classif.npv classif.acc 
#>  0.8750000   0.9400000   0.9210526   0.9038462   0.9111111 

# Create the three different plots
autoplot(pred, type = "roc")
autoplot(pred, type = "threshold")
autoplot(pred, type = "prc")
\end{verbatim}

While these classification metrics are certainly useful, there exist alternative metrics that prioritize quantifying the degree of concordance between the predicted scores and the true class, extending beyond the sole evaluation of binary class predictions.
The \emph{Brier score}\index{Brier score} (BS) was originally developed to quantify the accuracy of weather prediction \citep{brier1950verification}, consider for instance the rain probability prediction $\hat{\pi}$ and the target $y = 1$, indicating rain:

\begin{equation*}\label{eq:brier}
    \mathrm{BS} = \frac{1}{n} \sum_{i=1}^n \left(\hat{\pi}^i - y^i\right)^2.
\end{equation*}

The smallest scores are achieved when the predicted probability is closest to the true target, and larger deviations are penalized quadratically.
In this regard, the Brier score evaluates both calibration and discrimination.
Calibration evaluates how close the predicted probability for an outcome such as heart disease is to the true outcome, discrimination evaluates whether a patient with a higher underlying risk also receives a higher probability \citep{gerds2021medicalrisk}.
Especially in risk prediction contexts, individual predicted probabilities should be considered with caution if the underlying model is poorly calibrated.
It is common to report both the AUC and the Brier score for a given model, as the former is primarily a discrimination measure.

The logistic loss
\begin{equation*}\label{eq:logloss}
    \mathrm{LogLoss} = \frac{1}{n} \sum_{i=1}^n \left(-y^i \log\left(\hat{\pi}^i\right) - \left(1 - y^i\right) \log\left(1 - \hat{\pi}^i\right)\right)
\end{equation*}
is commonly employed in logistic regression and the binomial likelihood.
In machine learning, it is also known as the log-loss, the binomial loss, or the binary cross-entropy.

In addition to the binary classification measures discussed here, many have extensions to multiclass settings, such as the multiclass Brier score, categorical cross-entropy, or various multiclass extensions of the AUC.

\subsection{Resampling and Generalization Performance}\label{sec:resampling}

When evaluating a model, the primary interest invariably centers on its predictive performance as evaluated on data the learner has not encountered during the training process.
Given the potential utilization of a model for diagnostic decision-making or future trend forecasting, the \emph{generalization performance}\index{generalization performance} or \emph{generalization error}\index{generalization error} is among the most important qualities of a machine learning model in a predictive context.
Regardless of the metric chosen to quantify the model performance,  it is imperative to obtain a reasonable estimate.

When training and evaluating a model on the same set of data $\mathcal{D}$, the estimated generalization error will be biased, with the model frequently \emph{overfitting}\index{overfitting} the training data. 
This problem is of course not specific to machine learning. The same phenomenon can be observed in classical statistical modeling as well.
However, many machine learning algorithms are highly flexible, usually surpassing traditional statistical models in the capacity to effectively memorize the patterns within their training data.
The prediction error on these data will appear very low, while the model is unable to provide accurate predictions for novel and unseen data.
Many machine learning algorithms have inherent mechanisms to combat overfitting, such as the depth of decision trees, or weight regularization or dropout for neural networks (see Secs.~\ref{sec:treebased} and \ref{sec:neuralnets}).

In the general case, resampling strategies are employed for model evaluation, specifically for the purpose of detecting overfitting and to adjust models accordingly. 
A simple approach is to randomly split the data $\mathcal{D}$ into two disjoint sets $\mathcal{D}_{\text{train}}$ and $\mathcal{D}_{\text{test}}$. A train-to-test ratio of 2/3 is a good rule of thumb.
Subsequently, a model is fitted on $\mathcal{D}_{\text{train}}$, the target of the test set $\mathcal{D}_{\text{test}}$ is predicted.
The resulting \emph{test error}\index{test error} will exhibit minimal bias as it has been assessed on previously unseen data, albeit at the cost of reducing the volume of data available for model training.
On the other hand, a significantly improved estimate of the generalization error has been acquired through the data partitioning, as opposed to the situation where the data remained undivided.
This approach is known as the \emph{holdout}\index{holdout} method. 
It is employed in our code examples due to its simplicity. Yet, in practical applications it is rarely considered optimal, primarily due to the use of a single test set, yielding merely one performance estimate for the generalization error.

\begin{figure}
\centering
  \includegraphics[width=.8\linewidth]{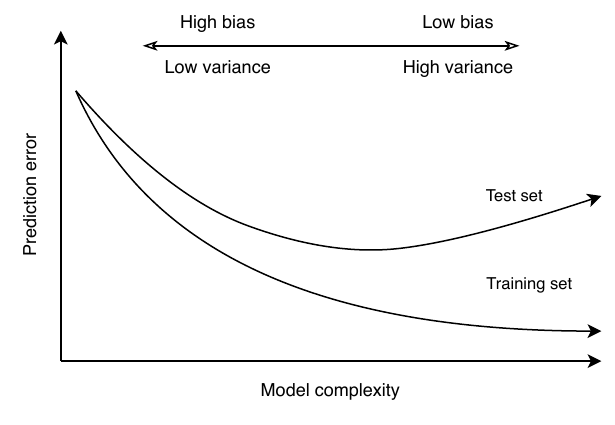}
  \caption{The bias-variance trade-off characterizes the common phenomenon of the prediction error decreasing with higher model complexity up to the point at which it begins to increase again. This is based on the fact that low model complexity typically implies a high bias, whereas high complexity tends to imply higher variance.}
  \label{fig:bias-variance}
\end{figure}

As is common in statistics, a trade-off between the bias and the variance of a model's generalization performance occurs (see Fig.~\ref{fig:bias-variance}).
Although the bias can be reduced by evaluating a larger training dataset, this will invariably lead to an increase in variance and vice versa.
Here, our trade-off is induced by the relative proportion of instances used for training and testing, respectively. The trade-off effect is comparable to that under model complexity, as described in Fig.~\ref{fig:bias-variance}.
Generally speaking, finding a balance between the bias and variance of the generalization error estimate can be challenging.
For more information on the topic, see \cite{Hastie_elem_stat_learning_2017_long}.

The second motivation for resampling is fair model comparison.
This can refer to either multiple versions of the same learner with different hyperparameter configurations, as discussed in Sec.~\ref{sec:tuning}, or to the comparison of multiple competing learners in a benchmarking experiment.
In the general case, however, they tend to follow the same principle:

\begin{enumerate}
    \item Split original dataset $\mathcal{D}$ into (smaller) datasets $\mathcal{D}_{b}$ with $b = 1,\ldots, B$.
    \item On each dataset $\mathcal{D}_{b}$:
      \begin{enumerate}
        \item Train learner.
        \item Estimate performance on $\mathcal{D}_{b}^{\ast}=\mathcal{D}\setminus \mathcal{D}_{b}$ with suitable performance measure.
      \end{enumerate}
    \item Aggregate performance estimates, e.g., with the arithmetic mean.
\end{enumerate}

In the case of the simple holdout strategy, $\mathcal{D}$ is split into two disjoint sets, but naturally, $\mathcal{D}$ can be split into an arbitrarily large number of subsets, or repeat the process multiple times and aggregate the results.
In the following, we present the most popular resampling strategies and illustrate their application with \texttt{mlr3}, beginning with this use of a conventional train-test split:\\

\begin{verbatim}
# Use a random forest learner
learner <- lrn("classif.ranger",  num.trees = 100, 
              predict_type = "prob")

# Create a simple train-test split with a 2/3 ratio
split <- partition(task, ratio = 2/3)

# Train the learner on the train set
learner$train(heart_task, row_ids = split$train)

# Evaluate on both train set and test set
learner$predict(heart_task, split$train)$score(msr("classif.auc"))
#> classif.auc 
#>   0.994625 
learner$predict(heart_task, split$test)$score(msr("classif.auc"))
#> classif.auc 
#>    0.9445 
\end{verbatim}

\subsubsection{Cross-Validation}\label{sec:crossvalidation}
\emph{Cross-validation}\index{cross-validation} (CV) is one of the most common resampling strategies.
For cross-validation, the full dataset is split into $k$ disjoint subsets $\mathcal{D}_1, \ldots, \mathcal{D}_k$, which is also why this approach is often referred to as $k$-fold cross-validation.
\begin{figure}
    \centering
    \includegraphics[width=0.9\textwidth]{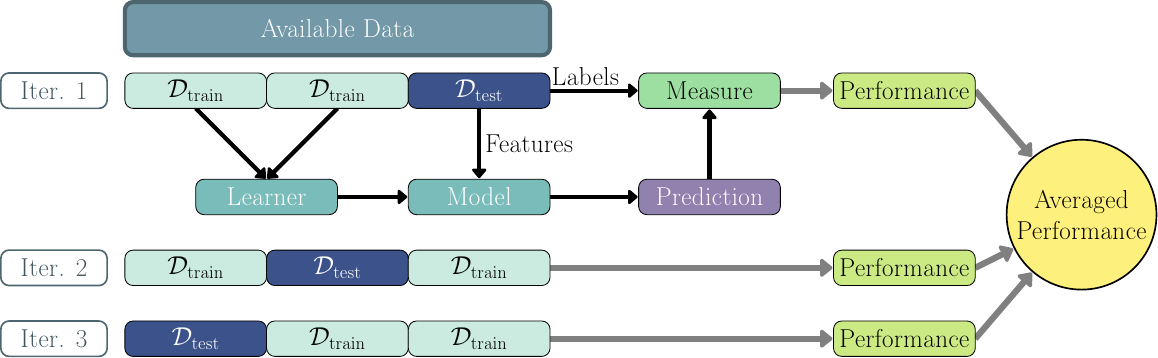}
    \caption{3-fold cross-validation splits the data into three disjoint sets. In each iteration, two thirds of the data serve as training set while the remaining fold serves as the test set for performance estimation. The results of the folds are averaged for a final performance estimate. Reprinted with permission from \citet[Chap.3]{Bischl2024}.}
    \label{fig:resamp-cv}
\end{figure}
For an example of $k = 10$, the data are split into ten equally sized subsets. In the next step, the learner is trained on the combined data of nine of them, using the remaining dataset as test set for prediction.
This process is then repeated nine times until each of the original ten sets was used as the test set exactly once.
Within each iteration, a new model is fit, used for prediction, and then discarded.
The ten performance scores are then averaged, resulting in our estimate for the generalization error.
This process is frequently employed due to its efficient utilization of data, ensuring that each original data instance is guaranteed to have been employed for evaluation purposes.
It also allows to adjust $k$ to balance computational complexity (larger $k$ implies fitting more models) and bias in the performance estimate. The bias increases for smaller $k$, since the size of each training set will be $\frac{k-1}{k}$.
In practice, 3-, 5- and 10-fold cross-validation are common choices.
For small datasets, this procedure can be applied repeatedly, averaging the resulting performance estimates across iterations to get a more stable estimate.
Using \texttt{mlr3}, cross-validation can be applied using three folds as illustrated in this example:

\begin{verbatim}
# Resample using 3-fold CV on the same learner as before
rr_cv3 <- resample(heart_task, learner, rsmp("cv", folds = 3))

# Evaluate on each fold...
rr_cv3$score(msr("classif.auc"))
#> task_id     learner_id resampling_id iteration classif.auc
#> 1:  heart classif.ranger            cv         1   0.9291101
#> 2:  heart classif.ranger            cv         2   0.8853695
#> 3:  heart classif.ranger            cv         3   0.8750000
#> Hidden columns: task, learner, resampling, prediction

# ...and aggregate the results
rr_cv3$aggregate(msr("classif.auc"))
#> classif.auc 
#>  0.8964932 
\end{verbatim}

A special case is \emph{leave-one-out cross-validation}\index{leave-one-out cross-validation} (LOO-CV), which sets $k = n$, i.e., the data are split such that each instance of the original data becomes the test set in one iteration of the procedure, whereas $n-1$ instances are available for training.
This alternative is appealing owing to the substantial size of the training sets. However, it possesses less desirable qualities in specific contexts, such as imbalanced classification problems.
The LOO-CV estimate requires training a learner $n$ times, which can be computationally expensive for complex learners.

\subsubsection{Subsampling}\label{sec:subsampling}
Instead of splitting the data once into disjoint sets, an alternative approach involves the repeated draw of subsamples in a specified ratio, such that each $\mathcal{D}_{b}$ is an independent draw.
This is referred to as \emph{subsampling}\index{subsampling} or \emph{repeated holdout}\index{repeated holdout}, as it is exactly that: A repeated application of the holdout strategy we introduced earlier, repeated $b$ times to result in $b$ performance estimates.
Subsampling is mostly recommended for small dataset sizes with a large number of iterations (e.g., 100, 1000, or more), to reduce the variance in the performance estimate.
Common values for the sampling ratio are 2/3 and 9/10, resulting in training sets of the same size as 3- and 10-fold cross-validation.
Due to the use of repeated random sampling, this strategy is also sometimes referred to as "Monte-Carlo cross-validation" in the literature.
Applying this strategy in \texttt{mlr3} is very similar to the cross-validation example we provided earlier:

\begin{verbatim}
resample(heart_task, learner, rsmp("subsampling", ratio = 2/3, 
                                   repeats = 100))
\end{verbatim}

\subsubsection{Bootstrapping} \label{sec:Bootstrap}
While subsampling draws samples without replacement, \emph{bootstrapping}\index{bootstrapping} randomly draws samples with replacement. This results in sets $\mathcal{D}_b$ of the same size as $\mathcal{D}$, in which instances may appear more than once.
On average, $1 - e^{-1} \approx 63.2\%$ of the original data constitute the training set, referred to as \emph{in-bag}\index{in-bag}.
The remaining instances used as test set are referred to as \emph{out-of-bag}\index{out-of-bag}.
This iterative process is repeated numerous times, with the resultant performance estimates being aggregated in a manner akin to previously introduced resampling techniques.
Since training sets contain duplicate observations, extra caution is warranted to achieve unbiased performance estimates, particularly when aiming for valid confidence intervals and when performing nested resampling (see Sec.~\ref{sec:model_building}).
As discussed previously in Sec.~\ref{sec:treebased}, the random forest learner is renowned for its incorporation of bootstrap-aggregation (bagging) of decision trees, with the out-of-bag prediction error playing the key role in the algorithm.
The application in \texttt{mlr3} is similar to previously introduced strategies:

\begin{verbatim}
resample(heart_task, learner, rsmp("bootstrap", repeats = 40))
\end{verbatim}
 
\section{Hyperparameter Tuning}
\label{sec:tuning}

Machine learning models are comprised of two different types of parameters, \emph{model parameters}\index{model parameters} and \emph{hyperparameters}\index{hyperparameters}. Model parameters are internal parameters, determined in the learning process by training the model on data. Examples include the coefficients $\bm{\beta}$ in regression models. In a neural network, the model parameters that are learned during training are the weights and biases assigned to each neuron in the network. In a decision tree, model parameters correspond to the decision rules that are used to split the data and the values, labels, or functions assigned to each leaf node. In contrast, hyperparameters are configurations of the learners that can be set manually by the user within a certain range and influence the learning process. While simple models like linear regression do not have hyperparameters, more complex machine learning models, such as tree-based methods or neural networks often have multiple hyperparameters. Some of the crucial hyperparameters in tree-based methods include the number of trees, tree depth, learning rate in boosted trees or the minimum number of instances in a leaf node. In neural networks, there is the number of layers, the number of neurons per layer, activation functions and learning rate as well as many hyperparameters to consider, which may significantly impact the model performance. The following section provides an in-depth description of the general process of determining an adequate configuration of hyperparameters, irrespective of the specific learner under consideration. It is accompanied by a practical example of tuning a specific learner for the use on the heart disease dataset. Sec.~\ref{sec:hyperparameter_optimization} introduces the concept of hyperparameter optimization. Sec.~\ref{sec:tuning_approaches} describes different approaches to hyperparameter tuning, beginning with exhaustive search (Sec.~\ref{sec:all_possible_combos}), grid search (Sec.~\ref{sec:grid_search}), random search (Sec.~\ref{sec:random_search}) and model-based optimization (Sec.~\ref{sec:model_based}). Finally, in Sec.~\ref{sec:model_building} the pieces are put together in the full model building process. 







\subsection{Hyperparameter Optimization}\label{sec:hyperparameter_optimization}
Hyperparameters are essential in machine learning, as they play a crucial role in balancing model complexity with overfitting, relating closely to the bias-variance trade-off discussed in Sec.~\ref{sec:resampling} and illustrated in Fig.~\ref{fig:bias-variance}. In this chapter, we use gradient boosting to predict the presence of heart disease as an introductory example of how hyperparameter tuning can work in practice. Gradient boosting combines weak learners, e.g., decision trees, into a strong learner iteratively by fitting new models to the residuals of the previous models using gradient descent (see Sec.~\ref{sec:treebased}). First, we instantiate the learner and check the settings for some of the hyperparameters. We are using the \texttt{xgboost} implementation based on \cite{xgboost_chen2016}. The \texttt{nrounds} hyperparameter specifies the number of boosting iterations, \texttt{max\_depth} defines the maximum depth of the trees, \texttt{eta} controls the learning rate, and \texttt{lambda} controls the $L_2$ regularization terms of the tree leaf weights. The \texttt{mlr3tuning} package specifies the upper and lower limits of these hyperparameters as well as default values. Default values for these hyperparameters are typically set based on empirical observations of how they perform on a wide range of datasets. By training \texttt{xgboost} using the default hyperparameter values, a classification accuracy of 78.9\% is obtained using 10-fold cross-validation and 67.8\% for a simple train-test split. This performance could potentially be improved by choosing different hyperparameters. Systematically exploring the space of possible \emph{hyperparameter configurations}\index{hyperparameter configurations} (HPC) to find the optimal hyperparameters is crucial for achieving the best possible model performance. This is what is generally referred to as \emph{hyperparameter tuning}\index{hyperparameter tuning}. 

\begin{verbatim}
# Instantiate learner
learner = lrn("classif.xgboost")

# Description of hyperparameters the learner has with ranges,
# defaults and current values stored in param_set
as.data.table(learner$param_set)[c(31, 24, 12, 20),
 .(id, class, lower, upper, nlevels, default)]
#>           id    class lower upper nlevels        default
#> 1:   nrounds ParamInt     1   Inf     Inf <NoDefault[3]>
#> 2: max_depth ParamInt     0   Inf     Inf              6
#> 3:       eta ParamDbl     0     1     Inf            0.3
#> 4:    lambda ParamDbl     0   Inf     Inf              1
\end{verbatim}




The terms hyperparameter tuning and \emph{hyperparameter optimization}\index{hyperparameter optimization} (HPO) are often used interchangeably since the tuning process can be viewed as an optimization problem. The set of hyperparameters $\Lambda$ to be tuned can be defined as optimization variables, and the tuning process can be formulated as a minimization problem of the generalization error, estimated by the population loss, as denoted by \cite{bartz2023hyperparameter}

\begin{equation*}
\bm{\hat{\lambda}} = \underset{\bm{\lambda} \in \Lambda}{\arg \min} \,\mathbb{E} \left[L(\mathcal{A}_{\lambda}(\mathcal{D}_{\text{train}})(\bm{x})), y\right].
\end{equation*}

Formally, by optimizing the generalization error on the underlying data distribution with respect to the hyperparameter set $\bm{\lambda} \in \Lambda$, the learning algorithm $\mathcal{A}$ can estimate the model $f$ with $\hat{f}$ on the training set $\mathcal{D}_{\text{train}}$. However, in practice, only a sample of the full population is available. An unbiased estimate of the generalization error can only be obtained from data unseen by the learner during the training process, as established in Sec.~\ref{sec:eval}. Similarly, if the same data are used for model or hyperparameter selection and model evaluation, the actual performance estimate of the model might be severely biased. To obtain an unbiased performance estimate, \emph{nested resampling}\index{nested resampling} should be used. Nested resampling provides a more reliable performance estimate and avoids information leakage that happens when using the same data to tune a model and evaluate its performance, by using a series of training, validation\index{validation set}, and test splits. For a detailed explanation of nested resampling refer to Sec.~\ref{sec:model_building}. Splitting the available data sample into training, validation, and test sets can be considered a special case of nested resampling with holdout. The validation set $\mathcal{D}_\text{val} \subseteq \mathcal{D}$ is used to compare model performance for different hyperparameter configurations to minimize the validation error. The optimization problem can be reformulated by replacing the expected value with the Monte-Carlo estimator:

\begin{equation}\label{eq:tuning2}
\bm{\hat{\lambda}} \approx \underset{\bm{\lambda} \in \Lambda}{\arg \min} \frac{1}{{|\mathcal{D}_{\text{val}}|}} \sum_{(\bm{x}, y) \in \mathcal{D}_{\text{val}}} L\left(\mathcal{A}_{\bm{\lambda}}(\mathcal{D}_{\text{train}})(\bm{x}), y\right).
\end{equation}

However, as discussed previously in Sec.~\ref{sec:eval}, the single fold approach has several limitations. 
An effective solution to overcome these limitations is the use of nested resampling (Sec.~\ref{sec:model_building}). As the focus of this section is on hyperparameter tuning, we consider a simplified, special case of nested resampling with an outer loop involving one partition for training and testing (holdout) and $k$-fold cross-validation in the inner loop for tuning.
The basic hyperparameter tuning process optimizes for $\hat{\lambda}$ performing the following steps for every iteration $h$ within a predefined tuning budget:
\begin{enumerate}
    \item Choose a hyperparameter configuration $\bm{\lambda}_h$ from the space of hyperparameters $\Lambda$.
    \item Train the learner $\mathcal{A}_{\bm{\lambda}_h}$ using $\mathcal{D}_{\text{train}}$.
    \item Record final performance on $\mathcal{D}_{\text{val}}$.

\end{enumerate}

\subsection{Hyperparameter Tuning Approaches}\label{sec:tuning_approaches}

Various approaches exist for hyperparameter tuning, varying in how they select new sets of hyperparameters to evaluate in each iteration. In the following sections, we present and provide code examples for some of the commonly used methods.

\subsubsection{Exhaustive Search}\label{sec:all_possible_combos}

The most rigorous approach to hyperparameter tuning would be to evaluate all possible hyperparameter combinations. In practice, however, such an \emph{exhaustive search}\index{exhaustive search} is not feasible for most machine learning models due to the large number of hyperparameters and their feasible range of values. Although in some situations, researchers may have an intuition about good hyperparameter values and can choose them manually, this approach is rarely justified over comparing a larger set of hyperparameter configurations.

\subsubsection{Grid Search}\label{sec:grid_search}

\emph{Grid search}\index{grid search} is a common and straightforward method for hyperparameter tuning that involves discretizing the search space into a grid of possible hyperparameter value combinations, evaluating each by comparing performance metrics. 

To perform grid search using \texttt{mlr3tuning}, the data are first split into training and testing sets to ensure unbiased evaluation of the final model. Then the learner is instantiated, in combination with the search space of the chosen hyperparameters. Defining the \emph{search space}\index{search space} is straightforward for some hyperparameters such as \texttt{lambda} or \texttt{eta}. For hyperparameters like \texttt{nrounds}, a reasonable domain size such as [1,500] can be used initially and can be expanded if the optimal values are found to be near the domain's limits. The \texttt{tune()} function instantiates and executes a tuning instance in one step. The \texttt{Tuner} class specifies the hyperparameter optimization algorithm. The \texttt{resolution} is the number of distinct values tested per hyperparameter, while the \texttt{batch\_size} determines the number of configurations evaluated simultaneously and the frequency of terminator checks. The \texttt{Terminator} class in \texttt{mlr3} sets the \emph{tuning budget}\index{tuning budget}, which determines when to stop the tuning algorithm. For grid search, the size of the grid automatically determines the termination. Post-tuning, we can access the results and construct a new learner with the optimized hyperparameters. The resulting accuracy on the test set is 79.9\%. For more concrete information on hyperparameter tuning with \texttt{mlr3tuning}, please refer to \citet[Chap.4]{Bischl2024}. 

\begin{verbatim}
# Instantiate learner with search space
learner <- lrn("classif.xgboost",
  nrounds  = to_tune(1,500),
  max_depth = to_tune(1,20),
  eta = to_tune(1e-10,1),
  lambda = to_tune(1e-10,1)
)

# Instantiate tuning instance
tuning_instance_gs <- tune(
  tuner = tnr("grid_search", resolution = 10, batch_size = 20),
  task = task_train,
  learner = learner,
  resampling = rsmp("cv", folds = 10),
  measures = msr("classif.acc"),
  terminator = trm("none")
)

# Use tuned learner
learner_tuned <- lrn("classif.xgboost")
learner_tuned$param_set$values <- 
  tuning_instance_gs$result_learner_param_vals
learner_tuned$train(task, row_ids = split$train)

# Score of tuned model on test data
learner_tuned$predict(task, row_ids = split$test)$score(measure)
#> classif.acc 
#>       0.799
\end{verbatim}

\begin{figure}
\centering
    \begin{subfigure}[b]{0.42\textwidth}     
        \centering
        \includegraphics[width=\textwidth]{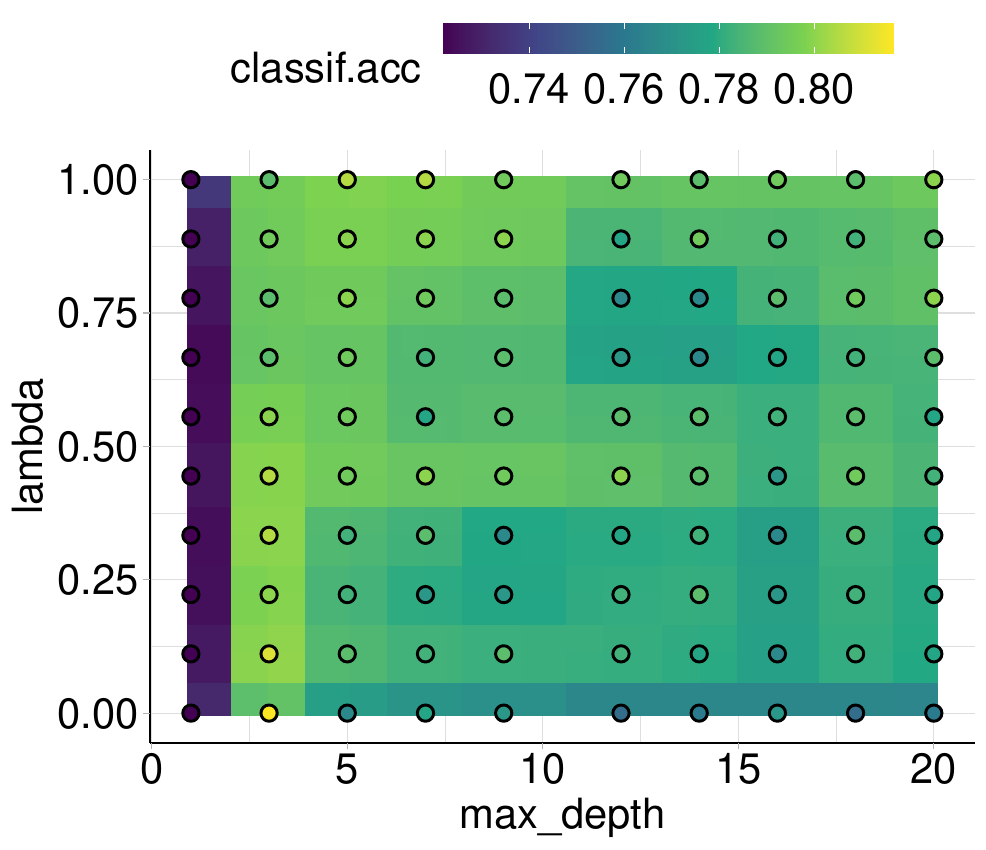}
        \caption{Grid search}
    \end{subfigure}%
    \hspace{5mm}
    \begin{subfigure}[b]{0.42\textwidth}
        \centering
        \includegraphics[width=\textwidth]{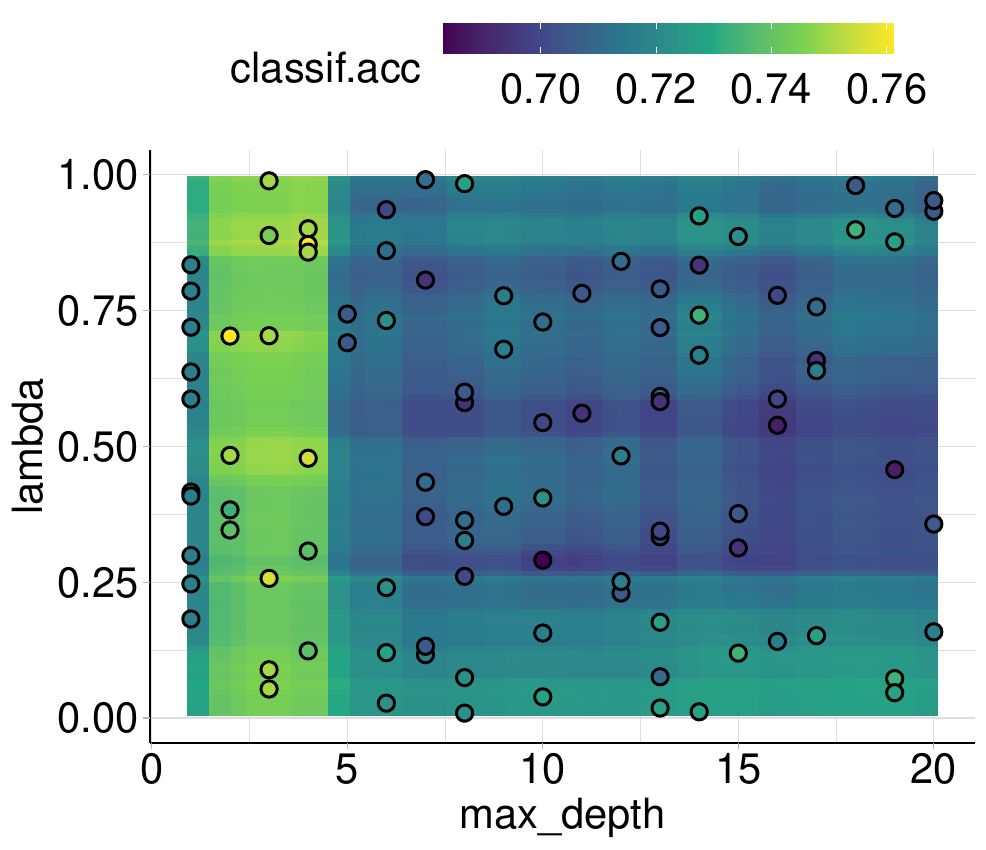}
        \caption{Random search}
    \end{subfigure}
    \caption{Illustration of (a) grid search and (b) random search for tuning two hyperparameters. The hyperparameters \texttt{lambda} and \texttt{max\_depth} are tuned within the search space \texttt{lambda} $\in (0, 1)$ and \texttt{max\_depth} $\in (0, 20)$. The dots with black outline are the respective (\texttt{max\_depth}, \texttt{lambda}) combinations for which the \texttt{xgboost} model is evaluated in every iteration of the (a) grid search or (b) random search algorithm. In (a), combinations are selected on an equally spaced grid within the borders of the search space. In (b), the combinations are selected randomly within the borders of the search space. A high cross-validation classification accuracy (classif.acc) is represented by yellow color coding, darker colors indicate points with lower accuracy.}\label{fig:grid_search_vs_random_search}
\end{figure}

Fig.~\ref{fig:grid_search_vs_random_search}\,(a) illustrates the grid search algorithm for tuning two hyperparameters. The evaluated hyperparameter combinations are indicated by points, while the colors represent the corresponding resampling accuracy. This figure highlights the strengths and weaknesses of the grid search method. On the one hand, grid search is systematic and reproducible, as it ensures that no potential optimal values located within the predefined grid are missed. It provides a comprehensive overview of the model's performance for different hyperparameter combinations, facilitating interpretability and an understanding of the relationship between hyperparameters and performance. On the other hand, it is computationally expensive and very rigid, as it evaluates many hyperparameter combinations in areas of poor performance due to the predetermined structure of the grid. Furthermore, potentially optimal values between or beyond the predefined grid points might be missed, particularly when the grid is sparsely defined and the hyperparameters to be optimized are continuous. Due to its interpretable, reproducible and deterministic nature, grid search is often recommended for models with two to three hyperparameters, but usually not for larger hyperparameter spaces \citep{bergstra2012random,bischl2023hyperparampaper}.

\subsubsection{Random Search}\label{sec:random_search}

Instead of evaluating only points in a predefined grid, \emph{random search}\index{random search} explores a random subset of points sampled from the entire search space. The optimization speed increases as the subset of sampled values decreases, but the accuracy may suffer if too few values are evaluated. Random search has been demonstrated to outperform grid search, particularly in cases where only a few hyperparameters significantly impact the final performance of the machine learning algorithm \citep{bergstra2012random,bischl2023hyperparampaper}.

To conduct random search using \texttt{mlr3tuning}, the \texttt{Tuner} and \texttt{Terminator} classes in the \texttt{tune()} function must be modified, as shown below. Unlike grid search, random search does not terminate automatically, and the user must define the tuning budget explicitly. A common approach is to set a limit on the number of evaluations or random draws from the subset. The resulting classification accuracy for the tuned example model is 82.5\%.

\begin{verbatim}
# Instantiate tuning instance
tuning_instance_rs <- tune(
  tuner = tnr("random_search"),
  task = task_train,
  learner = learner,
  resampling = rsmp("cv", folds = 10),
  measures = msr("classif.acc"),
  terminator = trm("evals", n_evals = 500)
)

...

# Score of tuned model on test data
learner_tuned$predict(task, row_ids = split$test)$score(measure)
#> classif.acc 
#>      0.825 
\end{verbatim}

In Fig.~\ref{fig:grid_search_vs_random_search}\,(b), a random search in a search space of two hyperparameters is depicted for illustrative purposes. 
It can be seen that the points are randomly sampled and do not form a grid-like structure. Although seemingly straightforward, random search remains a critical baseline for evaluating the performance of new hyperparameter optimization methods. 
It offers the key advantages of being more flexible and faster than grid search, enabling easy adaptation to various search spaces and the handling of numerous hyperparameters. 
Random search works well even in large hyperparameter spaces, if the effective dimension (i.e., number of hyperparameters with a large impact on the performance) is low, which is often the case in machine learning models \citep{bischl2023hyperparampaper}.  
Furthermore, similar to grid search, random search can be parallelized efficiently, substantially increasing computational speed. On the other hand, random search is less systematic than grid search and in particular for categorical hyperparameters, may lead to suboptimal results due to the potential omission of optimal values. 
Moreover, random search complicates interpretation of the relationship between hyperparameters and model performance and is not as easily reproducible as grid search.

\subsubsection{Model-Based Optimization}\label{sec:model_based}

Both random search and grid search fall under the category of \emph{model-free searches}\index{model-free search procedure}: hyperparameter optimization methods that tune the hyperparameters of a learner solely based on their performance on a given dataset. As a consequence, model-free approaches suffer from long run times, a sensitivity to the search space and search strategy as well as a lack of optimality guarantees for the tuning results. 
Most notably, model-free searches are inefficient, as they do not leverage prior knowledge of the problem or information gained about the hyperparameter configurations during the search process. 
\emph{Model-based search procedures}\index{model-based search procedure} aim to utilize this information to model the relationship between hyperparameter values and the learner's performance on the validation set, to guide the search for the optimal hyperparameter set. 
Central to model-based optimization methods are \emph{surrogate models}\index{surrogate models}, which are simple models intending to mimic highly complex relationships. In the context of model-based hyperparameter optimization, the surrogate model is an approximation of the actual objective function that is used in the optimization process with the intention to avoid computationally expensive calls to the objective function by estimating its behavior. Popular model-based methods include Bayesian optimization, tree-structured Parzen estimators (TPE), sequential model optimization (SMO), gradient descent-based algorithms or evolutionary algorithms. Since this chapter does not delve into the specifics of these methods, we refer to \cite{bergstra2011algorithms,bischl2023hyperparampaper} for exhaustive overviews. However, all model-based methods adhere to a basic high-level structure, which can be summarized in four steps.

\begin{enumerate}
    \item \textbf{Build surrogate model} \\
    A surrogate model is constructed, typically as a mathematical function that maps hyperparameter values to an estimate of the model's performance such as the (cross-)validation error. Typical examples of surrogate models are Gaussian processes, random forests and neural networks.
    \item \textbf{Optimize surrogate model} \\
    Find the hyperparameter values that (are likely to) yield the best model performance based on the surrogate model.
    \item \textbf{Evaluate machine learning model} \\
    Obtain performance estimate of the machine learning model with the hyperparameter values obtained from the surrogate model.
    This performance estimate is used to update the surrogate model and improve the accuracy of the model-based search process.
    \item \textbf{Repeat steps 1-3} \\
    Until a stopping criterion is met or the budget is exhausted.
\end{enumerate}

As an example of model-based search in \texttt{mlr3tuning}, we select Bayesian optimization. This strategy employs a surrogate function to model the posterior distribution of model performance given the observed data. A test set classification accuracy of 82.8\% is achieved with 30 iterations. The number of iterations is significantly reduced compared to the model-free methods since a considerable part of the evaluation burden can be shifted from evaluations of the actual objective function to surrogate evaluations. However, this comes at the cost of additional computational overhead for each iteration because the surrogate model must be iteratively fitted and updated. Thus, one faces a trade-off between needing fewer iterations, but each iteration requiring more computation time. Generally, model-based methods tend to win this trade-off \citep[][Chap.4]{bischl2023hyperparampaper,Bischl2024}. In summary, model-based search techniques offer several advantages, including higher efficiency and scalability, and can provide insights into the underlying structure of the data and model behavior through the surrogate function.

\begin{verbatim}
# Instantiate tuning instance
tuning_instance_bo <- tune(
  tuner = tnr("mbo"),
  task = task_train,
  learner = learner,
  resampling = rsmp("cv", folds = 10),
  measures = msr("classif.acc"),
  terminator = trm("evals", n_evals = 30)
)
  
...

# Score of tuned model on test data
learner_tuned$predict(task, row_ids = split$test)$score(measure)
#> classif.acc 
#>      0.828
\end{verbatim}

\subsection{Model Building Process}\label{sec:model_building}

To obtain a robust and reliable performance estimate, the best-practice approach is \emph{nested resampling}\index{nested resampling}. Nested resampling is a method that uses multiple levels of resampling to address the issue of overfitting in model evaluation by ensuring that the model fitted using the chosen hyperparameters generalizes well to new data. The first, or inner, level\index{inner resampling} of resampling is used for model selection or hyperparameter tuning, while the second, or outer, level\index{outer resampling} is used for model evaluation.

\begin{figure}[htb]
\centering
  \includegraphics[width=0.9\textwidth]{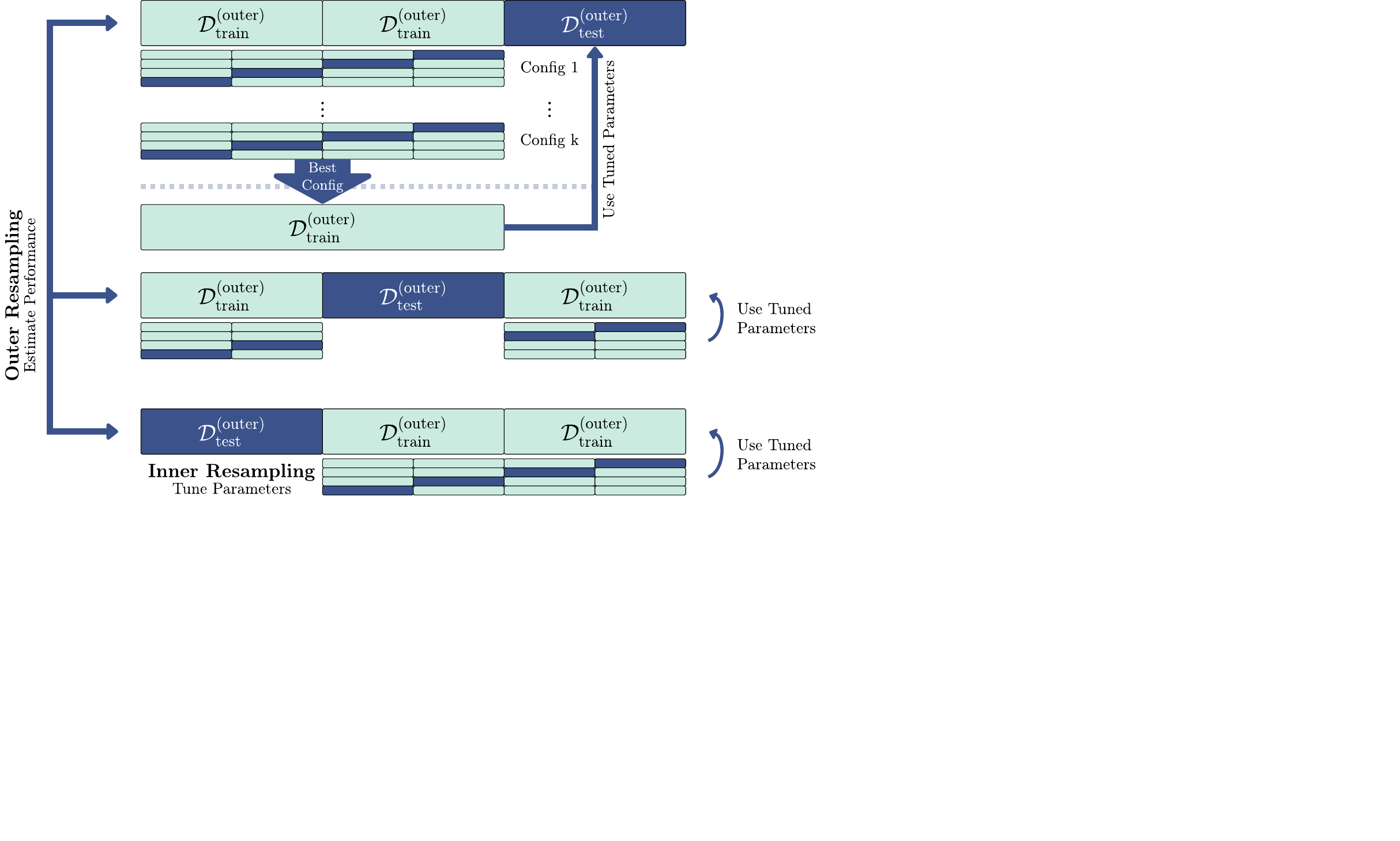}
  \caption{Illustration of the process of nested resampling. The larger blocks signify the use of 3-fold CV for evaluating the models in the outer resampling process, while the smaller blocks depict 4-fold CV for the inner resampling used in hyperparameter optimization. The training sets are represented by light blue blocks, whereas the test sets are shown by dark blue blocks. The thin blue arrows point to each step of the outer resampling process, in which a different portion of the data is used as a test set. Reprinted with permission from \citet[Chap.4]{Bischl2024}.}
  \label{fig:nested_resampling}
\end{figure}

A model-building process using nested resampling is comprised of the following steps: 

\begin{enumerate}
    \item \textbf{Outer loop resampling:} \\
The dataset is partitioned into multiple \emph{outer folds}\index{outer folds}, and the model is trained and evaluated through sequential partitioning into training and test sets.
     \item \textbf{Inner loop resampling:} \\
The training sets of each outer fold are further divided into multiple \emph{inner folds}\index{inner folds}. Model selection (i.e., hyperparameter tuning) is conducted on the inner folds.
     \item \textbf{Hyperparameter/Model selection:} \\
Based on the results of the inner loop resampling, the optimal hyperparameter configuration is determined for each outer fold. The selected model is trained on the complete training set of the respective outer fold.
     \item \textbf{Performance estimation:} \\
The selected models are evaluated on the outer fold's test set to obtain their respective performance estimates. Averaging over these outer loop performance estimates yields an aggregated performance measure.
\end{enumerate}
The process is visualized in Fig.~\ref{fig:nested_resampling}, which shows a nested cross-validation process for 3 outer and 4 inner folds. For a comprehensive overview of nested resampling, we refer to \cite{bischl2023hyperparampaper}. 

In \texttt{mlr3tuning}, the \texttt{AutoTuner} class can implement nested resampling automatically without constructing a new learner with optimal hyperparameters. It shares the same inputs as the \texttt{tune()} function, except for the \texttt{task}. The code example uses Bayesian optimization for hyperparameter tuning with cross-validation and 10 inner and outer folds. The inner cross-validation and tuning setup is specified with the \texttt{auto\_tuner()} function, it is then passed to \texttt{resample()} for the outer cross-validation process. The \texttt{extract\_inner\_tuning\_results()} function returns the optimal HPCs across all inner folds with their performance. The accuracy estimates on the inner folds can be compared to the classification accuracy estimated on the outer folds to detect potential issues of overfitting. The aggregated performance over all outer resampling iterations yields an unbiased estimate of the generalization error of 83.3\%. The outer resampling performance is slightly lower than the inner resampling performance, but neither substantially nor consistently, indicating no overfitting issues for the tuned models.

\begin{verbatim}
# Inner resampling with auto_tuner
at_nested <- auto_tuner(
  tuner = tnr("mbo"),
  learner = learner,
  resampling = rsmp("cv", folds = 10),
  measure = msr("classif.acc"),
  terminator = trm("evals", n_evals = 100)
)

# Outer resampling with rsmp()
outer_resampling <- rsmp("cv", folds = 10)

# Combine inner and outer resampling steps
rr <- resample(task, at_nested, outer_resampling, store_models = TRUE)

# Inner resampling performance with optimal configurations
extract_inner_tuning_results(rr)[1:4, .(iteration, nrounds, max_depth,
                                        eta, lambda, classif.acc)]
#>   iteration nrounds max_depth        eta    lambda classif.acc
#> 1:         1     468        13 0.96438727 0.2628022   0.8482143
#> 2:         2      32         1 0.01709498 0.8880480   0.8785714
#> 3:         3     167         8 0.22692057 0.1432204   0.8250000
#> 4:         4       2         1 0.66714936 0.4984954   0.8482143

# Outer resampling performance
rr$score(measure)[, .(iteration, classif.acc)]
#>  iteration classif.acc
#> 1:         1   0.6666667
#> 2:         2   0.8750000
#> 3:         3   0.8750000
#> 4:         4   1.0000000

# Aggregated performance over all outer folds
rr$aggregate(measure)
#> classif.acc 
#> 0.833 
\end{verbatim}

In conclusion, hyperparameter tuning or optimization is a crucial step in any machine learning pipeline when dealing with highly complex models. While the field offers proven search strategies, advanced algorithms, and sophisticated software implementations, it is crucial to acknowledge that this combination is not a panacea and does not guarantee to find the absolute best configuration for every problem. Although automated software solutions may suggest quick and easy optimal hyperparameter configuration, hyperparameter optimization might still involve a considerable amount of trial and error. In practice, fine-tuning the hyperparameters can be a time-consuming and iterative process, requiring multiple rounds of experimentation. In particular, for state-of-the-art deep learning models with vast high-dimensional hyperparameter spaces, the curse of dimensionality remains a challenge, and finding the optimal hyperparameter combination may still require substantial computational resources.

\section{Interpretable Machine Learning}
\label{sec:iml}

\begin{figure}[ht]
    \centering
    \includegraphics[width=0.9\textwidth]{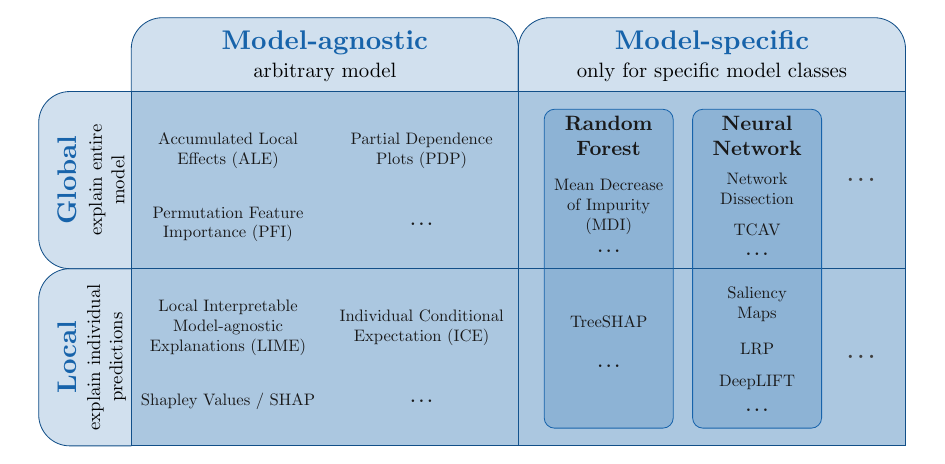}
    \caption{Overview of the most popular post-hoc interpretable machine learning methods.}
    \label{fig:iml_overview}
\end{figure}

One of the major challenges in the application of machine learning is the lack of interpretability and model transparency. The traditional statistical models used in epidemiology, such as linear regression, logistic regression, and Cox regression models, provide interpretable coefficients that help researchers understand the relationship between the features and the target. In contrast, more complex machine learning models, such as random forests and neural networks (see Sec.~\ref{sec:supervised}), are often considered "black boxes" as they lack transparency in their decision-making process. This lack of interpretability limits the potential of machine learning to inform decision-making, especially in epidemiology and healthcare, where decisions need to be explainable and transparent \citep{ahmad2018}. \emph{Interpretable machine learning}\index{interpretable machine learning} (IML) is an emerging field that aims to address the challenge of understanding machine learning models and revealing data insights. The most commonly used distinguishing criteria for IML methods are the following \citep{molnar2022}:
\begin{itemize}
    \item \textbf{Intrinsic vs. post-hoc:}\index{intrinsic interpretability method}\index{post-hoc interpretability method} Intrinsic methods involve building models that are interpretable by design, also known as white boxes. For instance, a linear regression model is highly interpretable as it directly models the relationship between the features and the target through its coefficients. In contrast, post-hoc methods typically analyze complex and opaque models after a completed training procedure, attempting to explain the predictions made by these models without modifying the model itself.
    \item \textbf{Local vs. global:}\index{local interpretability method}\index{global interpretability method} Another way to categorize IML methods is based on the level of interpretability the method provides. On the one hand, local methods focus on explaining the predictions made for individual or groups of instances, while global methods reveal insight into the model as a whole. Local explanations are particularly useful when it is necessary to understand the reasoning behind a specific prediction. On the other hand, global methods are applied for a more general and instance-independent understanding of the model behavior across the entire dataset and beyond.
    \item \textbf{Model-specific vs. model-agnostic:}\index{model-specific interpretability method}\index{model-agnostic interpretability method} IML methods can also be distinguished according to their applicability to machine learning models. Model-specific methods are designed to work with a specific model type, such as a random forest or neural network, by leveraging the model internals for the explanation. In contrast, model-agnostic approaches can be applied to arbitrary models, regardless of their type or complexity. This category of methods generates explanations based solely on the relationship between the features, the model's predictions and the target without relying on additional model information or external factors.
\end{itemize}
While intrinsic methods are often well-known and established statistical techniques, which are covered in detail elsewhere in this book, we focus on post-hoc methods. In the following sections, we describe how predictions from complex black box models can be explained at both local and global levels, highlighting the importance of these methods in the context of epidemiology. Fig.~\ref{fig:iml_overview} provides an overview of the distinguishing criteria of post-hoc methods, accompanied by examples for each criterion.

\subsection{Model-agnostic Interpretability Methods}\label{sec:model-agnostic-iml}

A large group of IML methods is model-agnostic in nature, i.e., they deliver explanations for arbitrary predictive models, regardless of their underlying algorithm or architecture. They do not have direct access to the internal model processes but derive their interpretations on the basis of how the model reacts to changes in the input data. In order to simplify the notation for changes only in specific features $\bm{x}_S$ with $S \subseteq \{1, \ldots, p\}$, we denote the evaluation of a model $\hat{f}$ at $\bm{x}_S$ filled up with the fixed feature values of a given point $\bm{x}_{\bar{S}}$ with $\bar{S} = \{1,\ldots, p\} \setminus S$  by $\hat{f}\left(\bm{x}_S; {\bm{x}_{\bar{S}}}\right)$.

\subsubsection{Local Model-agnostic Interpretability Methods}\label{sec:model-agnostic-local-iml}

Local model-agnostic interpretable machine learning methods specifically focus on explaining the model's prediction for a given instance, such as understanding the decisive factors influencing the predicted outcome of a disease for a particular patient. In this section, we will discuss three popular local model-agnostic interpretability techniques: Individual conditional expectations (ICE), local interpretable model-agnostic explanations (LIME), and Shapley values (SV).

\begin{figure}[ht]
    \centering
     \begin{subfigure}[b]{0.9\textwidth}
         \centering
         \includegraphics[width=0.48\textwidth]{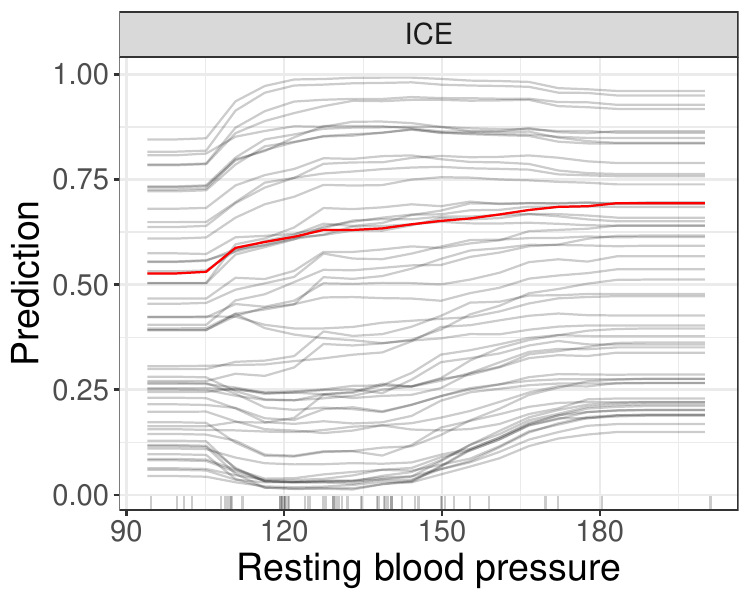}%
         \hspace{0.04\textwidth}%
         \includegraphics[width=0.48\textwidth]{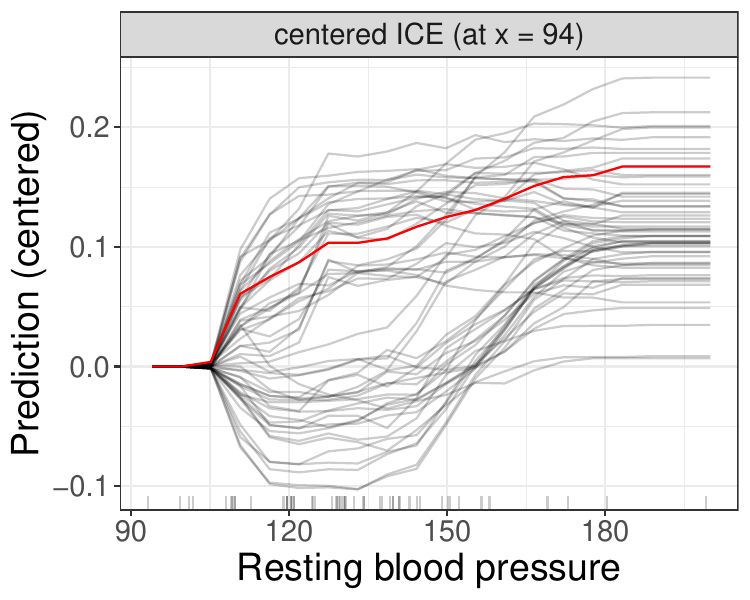}
         \caption{ICE plots}
     \end{subfigure}
     \begin{subfigure}[b]{0.9\textwidth}
         \vspace{1em}
         \centering
         \includegraphics[width=\textwidth]{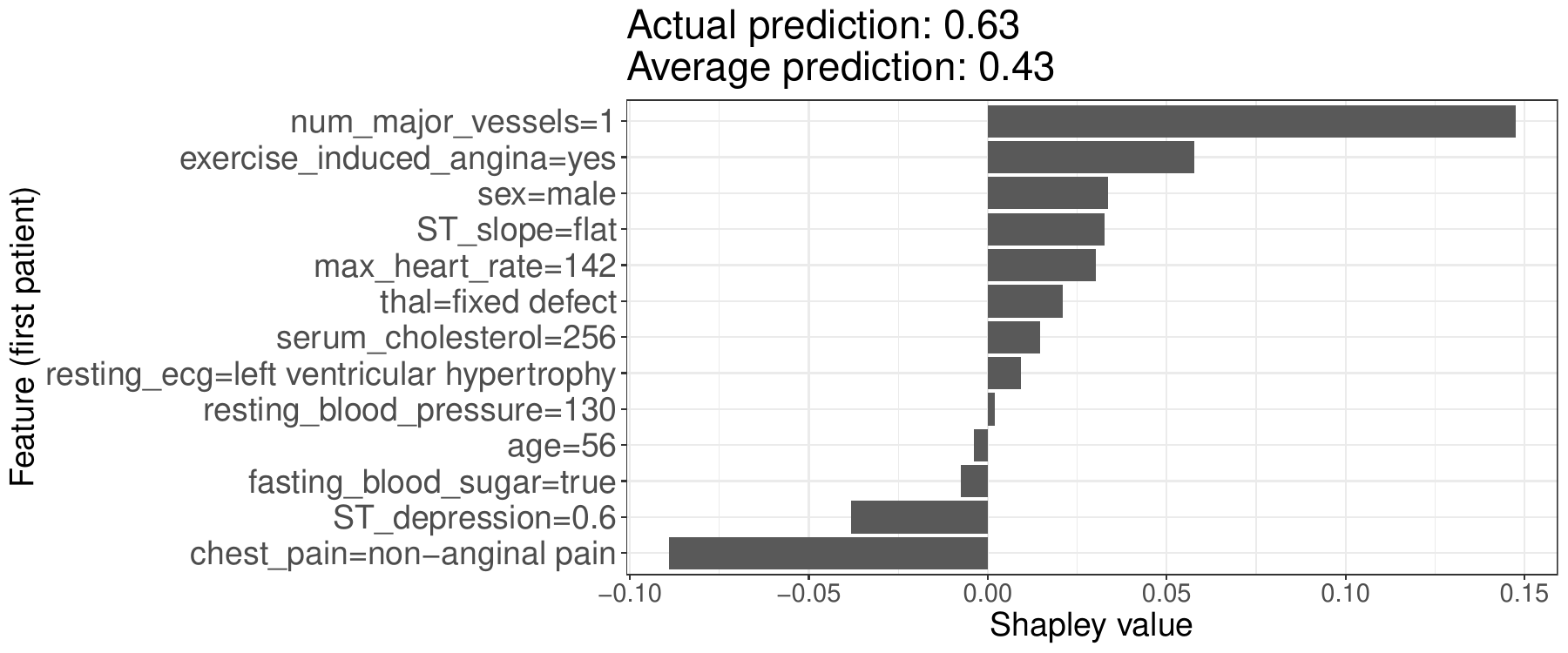}
         \caption{Shapley values}
     \end{subfigure}
     \caption{Illustration of the local model-agnostic IML methods ICE and Shapley values on the heart disease dataset: In (a), the left plot shows the ICE lines for the feature \texttt{resting\_blood\_pressure} and all test instances, while the right one displays the centered ICE plots at a blood pressure of 94. Additionally, the red highlighted line represents the first patient in the test data. In (b), the Shapley values of all features for the first patient are depicted.}
    \label{fig:iml_local_agnostic}
\end{figure}

\textbf{Individual conditional expectations (ICE)}\index{individual conditional expectations plot}, proposed by \cite{goldstein2015}, is a technique that visualizes the relationship between a single or a group of features of interest $S$ and the model's prediction while keeping all other features fixed. Consequently, an ICE plot reveals the conditionally expected effect on predictions when only the values of the features of interest $\bm{x}_S^{i}$ are modified for a specific instance $\bm{x}^{i}$ and the other $\bm{x}_{\bar{S}}^{i}$ remain unchanged. Mathematically, this relationship is represented by the graph of the function 
\begin{align*}
    \operatorname{ICE}_S \left(\bm{x}_S; \bm{x}^i\right) = \hat{f}\left(\bm{x}_S; {\bm{x}_{\bar{S}}^i}\right).
\end{align*}
Usually, this method focuses on analyzing a single feature (i.e., $|S| = 1$) resulting in a line plot (see Fig.~\ref{fig:iml_local_agnostic}\,(a)). Hence, we can display and compare multiple ICE plots for different instances within a single view. When dealing with two features $|S| = 2$, we can explore interactions between them using an area plot for a single instance. However, due to visualization limitations in higher dimensions, comparing these plots across instances or creating ICE plots involving more than three features in a single plot becomes challenging.

Due to the instance-specific range of the predictions and, thus, the wide variety of the line's vertical shifts, comparing the individual ICE plots for a given feature becomes difficult (see Fig.~\ref{fig:iml_local_agnostic}\,(a)). This clutter complicates the interpretation and perception of heterogeneity within the data. However, to address this issue, the centered ICE (c-ICE) plot was proposed. This approach selects an anchor point $\bm{a}$ of the features $S$ where all the ICE plots for the selected instances cross, and the individual lines are centered around this reference point (see Fig.~\ref{fig:iml_local_agnostic}\,(a)):
\begin{align*}
    \operatorname{c-ICE}_S \left(\bm{x}_S; \bm{x}^i \right) = \hat{f}\left(\bm{x}_S; \bm{x}_{\bar{S}}^{i}\right) - \hat{f}\left(\bm{a}; \bm{x}_{\bar{S}}^{i}\right).
\end{align*}

Nevertheless, both variants face the problem that in the presence of correlated features, instances that are unlikely in the dataset are generated, potentially leading to biased or even misleading interpretations \citep{molnar2022}. This issue can be tackled by shrinking the domain of the feature of interest conditioned on the remaining features \citep{Hooker2021}. In particular, accumulated local effect (ALE) plots leverage this conditioned perspective to reveal global effects even for correlated features (see Sec.~\ref{sec:model-agnostic-global-iml} for details).

The \textbf{local interpretable model-agnostic explanations (LIME)}\index{local interpretable model-agnostic explanation} method proposed by \cite{ribeiro2016} uses an interpretable model to emulate the black box locally. It selects neighboring data points of the instance of interest and perturbs these samples. The model's predictions for these perturbed instances are leveraged to train a locally interpretable model that approximates the behavior of the black box. During training, the loss value is weighted by the proximity of the perturbed data point to the instance of interest, forcing a direct focus on the relevant neighborhood. When the approximation of the black box predictions is sufficiently accurate, this local surrogate model, such as the linear model, offers an understanding and interpretation of the prediction by design.

LIME stands out among other model-agnostic approaches as it can be applied to various data types, including images, text, or tabular data. It enables the interpretation of instances from complex black box models using established statistical techniques like linear models. However, determining a suitable neighborhood is very difficult, and due to the perturbation, there is a potential risk of training the local surrogate on unlikely data points. Additionally, \cite{alvarez2018} provided evidence of LIME's limited robustness.

The local model-agnostic \textbf{Shapley values}\index{Shapley value} \citep{shapley1953} are derived from cooperative game theory and provide a fundamental approach to attribute the contribution of features to the prediction of interest. Figuratively speaking, each instance's prediction represents a game's payout, where all features are the players. Shapley values present a theoretical framework for fairly distributing the payout based on the contributions made by each player, taking into account all possible players' interactions and collaborations. In doing so, the Shapley value of a feature quantifies its average marginal contribution across all possible combinations of the remaining features, i.e., over all $2^{p-1}$ possibilities. In game theory, these combinations are often referred to as \emph{coalitions}\index{coalition} of players. Mathematically, the Shapley value of feature $j$ of instance $\bm{x}^{i}$ is defined as
\begin{align*}
\phi_j^{i} = \sum_{S \subseteq \{1, \ldots, p\} \setminus \{j\}} \frac{|S|!(p-|S|-1)!}{p!} \left(v(S \cup \{j\}) - v(S)\right), 
\end{align*}
where $v$ is the value function, $p$ is the total number of features and $S$ represents a set of feature's indices, i.e., a coalition. The marginal contribution of the selected feature $\bm{x}_j$ to a coalition $S$ is measured by taking the difference in the value function $v$ when the feature $\bm{x}_j$ is part of the coalition compared to when it is not included. 
A common choice of a value function is the marginal expectation over the data distribution $X_{\bar{S}}$ restricted to the out-of-coalition features $\bar{S}= \{1, \ldots, p\} \setminus S$ while fixing the in-coalition features $S$ with values of the instance of interest $\bm{x}^{i}$, i.e.,
\begin{align*}
v(S) = \mathbb{E}_{X_{\bar{S}}}\left[\hat{f}\left(X_{\bar{S}};\bm{x}_S^{i}\right)\right], \quad S \subseteq \left\{1, \ldots, p \right\}.  
\end{align*}
In addition to the expected value over the marginal distribution, there are many other approaches for defining a value function depending on the required accuracy or user application. For example, sampling can also be done from the conditional distribution conditioned on $\bm{x}_S$, or values can be set without sampling according to a predefined baseline value \citep{Chen2023}.

Despite the acclaimed theoretical properties of Shapley values, considering all possible coalitions and marginalizing them incurs a computationally expensive and often infeasible task, which grows exponentially with an increased number of features (players).
For this reason, numerous approximation methods have been developed: \cite{strumbelj2014} introduced an approach using a Monte Carlo approximation for the Shapley values. 
However, this permutation-based method can lead to the incorporation of unrealistic data instances when dealing with correlated features and is still very computationally intensive. 
\cite{lundberg2017} unified the approximation of Shapley values for various machine learning models by introducing a universally applicable framework called Shapley additive explanations\index{Shapley additive explanations} (SHAP). 
In particular, their variant KernelSHAP\index{KernelSHAP} combines the interpretability approaches of LIME and Shapley values. By leveraging a combination of sampling and regression techniques, KernelSHAP enables a fast approximation of the underlying true Shapley values. 
Due to the significantly shorter computation time of approximately accurate Shapley values, using dataset-aggregated Shapley values as a global measure of feature importance has become feasible. 
For the global IML method, the average absolute Shapley value of a feature is calculated as
\begin{align*}
 \phi_{j} = \frac{1}{n} \sum_{i = 1}^n \left|\phi_j^{i}\right|.
\end{align*}

\noindent
\textbf{Data Example}\newline

\noindent In this section, we have presented several local model-agnostic IML methods, out of which we will exemplify the use of ICE plots and Shapley values on the heart disease dataset. We trained a random forest on 80\% of the data, using the remaining 20\% for interpretation.

Considering the ICE plots in Fig.~\ref{fig:iml_local_agnostic}\,(a), we explain the marginal effect of the feature \texttt{resting\_blood\_pressure} on the prediction for each instance from the test dataset. In the notation of ICE plots, we have chosen $|S| = 1$, thus obtaining a single line for each patient in the test dataset. This line describes the patient-specific predicted probability of heart disease based on different blood pressure levels. In other words, this line indicates how the risk of heart disease for a patient -- based on the model -- changes when they have lower or higher blood pressure values. For example, the first patient in the test data (red line in Fig.~\ref{fig:iml_local_agnostic}\,(a)) generally has an increased risk of heart diseases, which would further increase with higher blood pressure. For the other test data instances, the centered ICE plot reveals two trends: First, patients whose risk decreases from a blood pressure of 100, and second, patients whose risk increases, as is the case with the first patient in the test data. Nevertheless, the risk grows for the majority of patients above a blood pressure of 130. Again, it must be mentioned that these explanations and inferred hypothetical behavior are based on the trained model and do not necessarily reveal causal relationships in the data. 

In contrast to the ICE plots, the Shapley values from Fig.~\ref{fig:iml_local_agnostic}\,(b) describe the feature-wise contribution to the first patient's probability of 66\% for a heart disease. For instance, \texttt{num\_major\_vessels} with a value of 1 has the strongest contribution to the prediction, thus favoring the risk of heart diseases. On the other hand, the negative Shapley values of \texttt{chest\_pain} and \texttt{ST\_depression} provide evidence that these values mitigate the patient's probability. Additionally, neither the age of 56 nor the resting blood pressure of 130 significantly influenced the prediction. However, in combination with the ICE plots from Fig.~\ref{fig:iml_local_agnostic}\,(a), it can be speculated that the influence of blood pressure could become more relevant if the patient's value increased.

\subsubsection{Global Model-agnostic Interpretability Methods}\label{sec:model-agnostic-global-iml}

Global model-agnostic interpretability methods attempt to explain the average behavior and predictions of machine learning black box models without relying on the internal model structure. The most popular techniques to date include partial dependence plots (PDPs), accumulated local effects plots (ALEs) and permutation feature importance (PFI), which all will be explained in more detail in the following. For more methods such as feature interaction effects or functional decomposition and a comprehensive review, we refer to \cite{molnar2022}. 

\textbf{Permutation feature importance (PFI)}\index{permutation feature importance} quantifies the importance of a specific feature $\bm{x}_j$ as the increase in prediction error from replacing the dataset's feature values $\bm{x}_j^1, \ldots, \bm{x}_j^n$ with a permuted version of itself, which is used to train the learner.\footnote{Permutation refers to the act of rearranging the order or sequence of a vector.} As a consequence of permutation, the relationship between the target variable $y$ and $\bm{x}_j$ is broken. The idea of PFI is straightforward: A feature may be considered important if the predictive accuracy suffers after severing its association with the target variable. PFI was originally developed by \cite{breiman2001random} for random forests, later \cite{fisher2019all} developed a model-agnostic version. Their proposed algorithm is:

\begin{enumerate}
    \item For a given model $\hat{f}$, estimate the original model error $e = \tfrac{1}{n} \sum_{i = 1}^n L(\hat{f}(\bm{x}^i), y^i)$.
    \item For a selected feature $\bm{x}_j$, permute the feature values across the whole dataset to obtain the permuted dataset $\{(\bm{\tilde{x}}^i, y^i)\}_{i = 1}^n$. 
    \item Make a prediction for every instance in the permuted dataset $\hat{f}(\bm{\tilde{x}}^i)$, then estimate the permuted model error $\tilde{e}_j = \tfrac{1}{n} \sum_{i = 1}^n L(\hat{f}(\bm{\tilde{x}}^i), y^i)$.
    \item Calculate the absolute PFI as difference $\operatorname{PFI}_j = \tilde{e}_j - e$ or the relative PFI as ratio $\operatorname{PFI}_j = \frac{\tilde{e}_j}{e}$.
    \item Repeat steps 2-4 for all other features $j=1,...,p$.
\end{enumerate}

Since PFI relies on the model error, it is usually preferable to estimate it on test data. PFI provides an easy to understand and intuitive way of quantifying and displaying feature importance. Yet, it is generally expensive to compute and it may produce misleading results for highly correlated features. This is again due to the generation of unrealistic data points as described in \ref{sec:model-agnostic-local-iml}. A potential solution for the problem of unrealistic data points is the use of conditional sampling instead of traditional marginal sampling techniques. Furthermore, PFI does not allow for quantifying the direction and magnitude of the feature effects.

A \textbf{partial dependence plot (PDP)}\index{partial dependence plot} is a graphical representation of the marginal effect of one or more features on the target variable in a machine learning model \citep{friedman2001greedy}. The marginal effect refers to the average predicted target across a range of values for one or more features, while holding all other features constant. Therefore, the plot elucidates the nature of the relationship between target and feature. Mathematically, the partial dependence function is expressed by
\begin{equation*}
\operatorname{PDP}_S\left(\bm{x}_S\right) = \mathbb{E}_{X_{\bar{S}}}\left[\hat{f}\left(\bm{x}_S;X_{\bar{S}}\right)\right],
\end{equation*}
where $\bm{x}_S$ represents the feature or set of features for which the partial dependence is calculated and $X_{\bar{S}}$ describes the data distribution restricted to the remaining features $\bar{S}$. Due to the marginalization over $X_{\bar{S}}$ the PDP depends only on the features $\bm{x}_S$. 

In practice, the PDP can be estimated with simple averages according to the Monte-Carlo method, as described by
\begin{equation*}
\operatorname{PDP}_S\left(\bm{x}_S\right) \approx \frac{1}{n} \sum_{i=1}^{n} \hat{f}\left(\bm{x}_S;\bm{x}_{\bar{S}}^i\right).
\end{equation*}
Here, $\bm{x}_{\bar{S}}^{i}$ are actual feature values from the dataset, and $n$ is the number of data samples. The partial dependence is computed by taking the average of the ICE curves at each feature value, introduced in Sec.~\ref{sec:model-agnostic-local-iml}. This averaging process smooths out the variations of individual instances and provides a consolidated view of the feature's effect on the predicted target. Note that for classification tasks with binary targets, the machine learning model predictions as well as the PDP correspond to class probabilities. 

\begin{figure}
    \centering
    \begin{subfigure}{0.8\textwidth}
        \includegraphics[width=\linewidth]{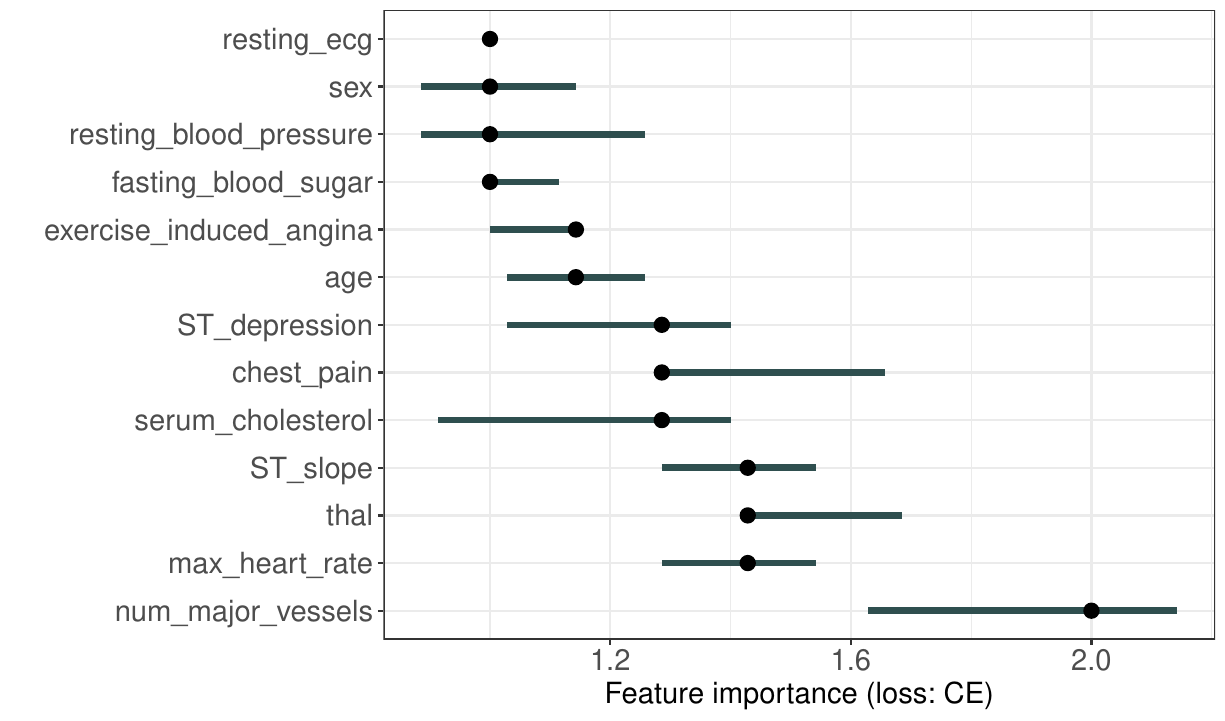}
        \caption{PFI}
    \end{subfigure}%

        \begin{subfigure}{0.8\textwidth}
        \includegraphics[width=\linewidth]{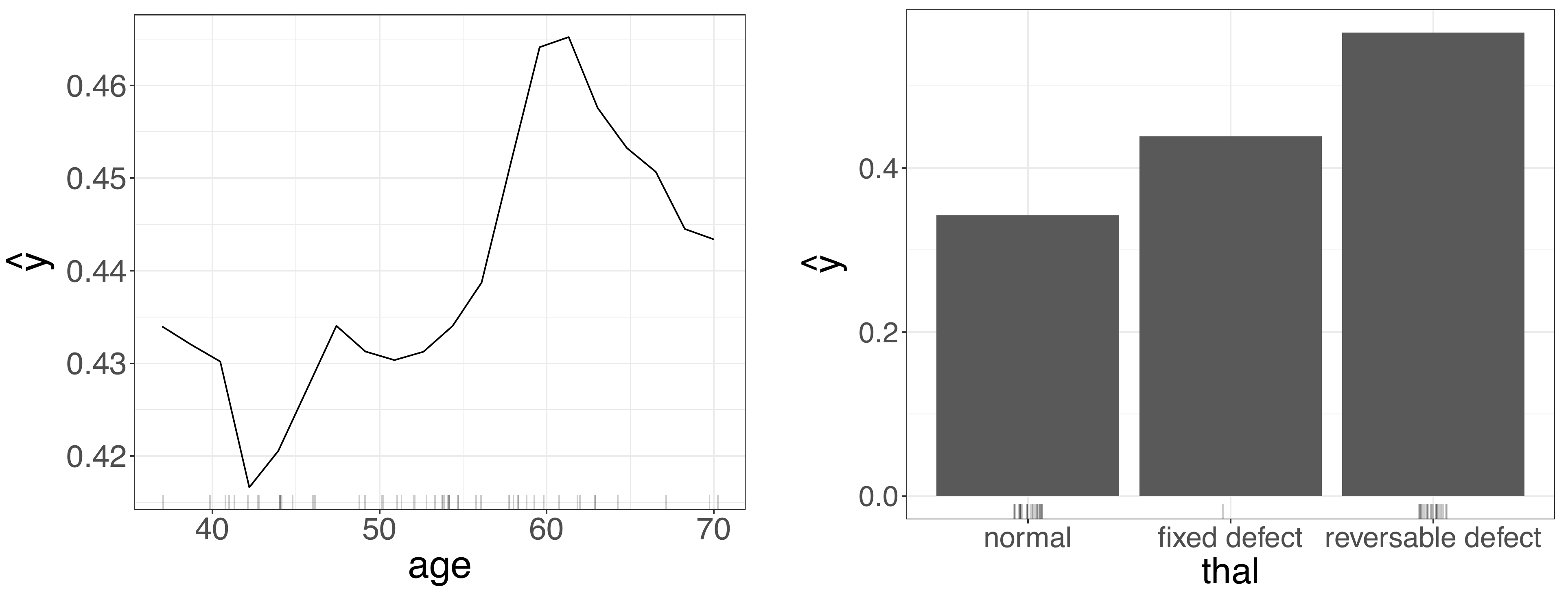}
        \caption{PDP}
    \end{subfigure}

    \begin{subfigure}{0.8\textwidth}
        \includegraphics[width=\linewidth]{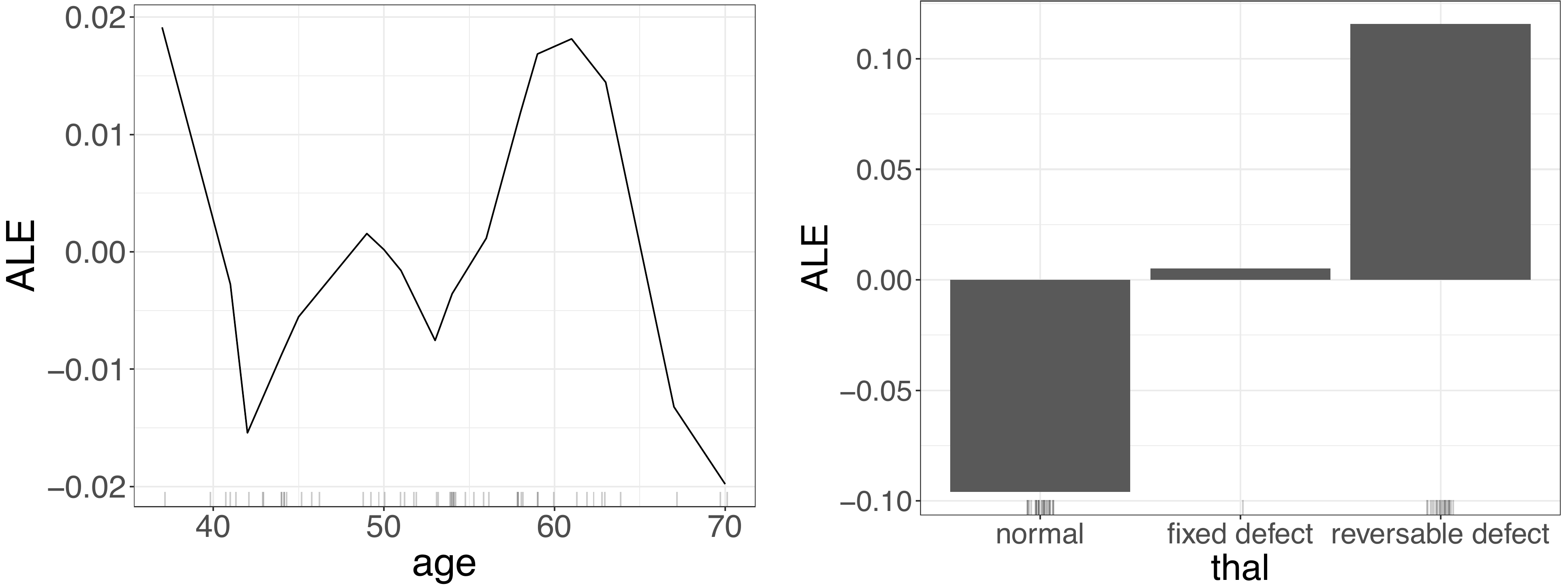}
        \caption{ALE}
    \end{subfigure}

    \caption{Illustration of the global model-agnostic IML methods PFI, PDP and ALE plots. All three methods are calculated on a test dataset based on a random forest classifier trained on the heart disease dataset. Plot (a) highlights the ordered relative PFI computed using the classification error (CE). The PDPs in (b) show how \texttt{age} (left plot) and \texttt{thal} (right plot) affect the average model predictions while holding all other features constant. In (c) ALE plots describe how changes in \texttt{age} (left plot) and \texttt{thal} (right plot) affect the average model predictions while taking into account the effects of other features.}
    \label{fig:iml-global-model-agnostic}
\end{figure}

PDPs are a popular interpretability tool since they are intuitive to compute and easy to implement. In case the features are not correlated, PDPs perfectly capture the average effect of one or multiple features on the prediction. However, this implicit assumption of independence constitutes the main limitation of PDPs. For instance, in the given example of the heart disease dataset (see Fig.~\ref{fig:iml-global-model-agnostic}), the maximum heart rate achieved during a cardiac stress test (\texttt{max\_heart\_rate}) and age are negatively correlated. A higher maximum heart rate achieved during a stress test is generally considered a positive indicator of cardiovascular fitness and should generally reduce the risk of heart disease. For the computation of the PDP at an older \texttt{age} (e.g., around 70), the average over the marginal \texttt{max\_heart\_rate} distribution is calculated, including very high \texttt{max\_heart\_rate} values, which are unrealistic to be achieved by elderly people, misrepresenting the true underlying effect of \texttt{age}. 

\textbf{Accumulated local effects (ALE)}\index{accumulated local effects} plots provide an alternative solution to PDPs for estimating unbiased feature effects in case features are correlated \citep{apley2020visualizing}. The ALE plot shows how the predicted target of a model changes on average as one or multiple features are varied, while controlling for the influence of all other features. The main idea behind the calculation of ALE is to remove unwanted effects of other features by first taking partial derivatives of the prediction function $\hat{f}$ with regard to the feature of interest $\bm{x}_S$ and then integrating them with respect to that same feature. Computing the local effects in terms of the partial derivative of $\hat{f}$ with regard to $\bm{x}_S$ removes the main effects of any other features, integrating or accumulating again with respect to $\bm{x}_S$ recovers its original main effect. 
The conceptual calculation of ALE plots involves three main steps:
\begin{enumerate}
    \item Estimating local effects $\frac{\partial \hat{f}(\bm{x}_S;\bm{x}_{\bar{S}})}{\partial \bm{x}_S}$ via finite differences.
    \item Averaging local effects over the conditional distribution of $X_{\bar{S}} \mid X_S = \bm{x}_S$ instead of the marginal distribution of $X_{\bar{S}}$ to avoid the extrapolation issue of PDPs.
    \item Integrating averaged local effects from a starting value $\bm{x}_\text{min}$ up to $\bm{x}_S$ to estimate the global main effect of $\bm{x}_S$. This avoids omitted variable bias issue since other unwanted main effects were removed in Step 1.
\end{enumerate}
The formula is given by
\begin{equation*}
\operatorname{ALE}_S(\bm{x}_S)= \int_{\bm{x}_{\text{min}}}^{\bm{x}_S} \mathbb{E}_{X_{\bar{S}}|X_S=\bm{z}_S} \left[\frac{\partial \hat{f}(\bm{z}_S; X_{\bar{S}})}{\partial \bm{z}_S}\right]\,d\bm{z}_S - C,
\end{equation*}
where $C \in \mathbb{R}$ is chosen to center the plot vertically.
To approximate the local effects or partial derivatives in practice, particularly for non-differentiable models, the feature is partitioned into many intervals, for which the differences in predictions are obtained. The intervals are usually defined by using the quantiles of the feature distribution. Each interval difference is equivalent to the effect the feature of interest has for a specific instance within a particular interval. All effects of all instances within each interval are then summed up and divided by their total number to compute the average change of predictions for the interval. In the final step, the average effects are accumulated by summing up the local effects across all intervals. For further details on the estimation process, we refer to \cite{apley2020visualizing} and \cite{molnar2022}.

Unlike numerical features, categorical features usually do not have a natural order, which is required for calculating the directional differences across intervals. An established approach is to arrange categories in order of similarity, which is determined based on the remaining features. The measure of similarity between two categories is calculated as the cumulative distance over all other features. Subsequently, multi-dimensional scaling is applied to the inter-category distances to reduce the original distance matrix to one dimension, from which a similarity-based order of the feature categories can be derived. For a more comprehensive explanation and further mathematical details, also on the computation of second-order ALE plots, we refer to \cite{apley2020visualizing} and \cite{molnar2022}.\\

\noindent\textbf{Data Example}\newline

\noindent In this section, we have presented several global model-agnostic IML methods. As for the local methods, practical examples of PFI, PDP and ALE plots will be provided using a random forest as our machine learning model of choice, trained on 80\% of the data, using the remaining 20\% for interpretation.

Fig.~\ref{fig:iml-global-model-agnostic}\,(a) displays the ordered relative PFI values for each feature. The most important features are the number of major vessels colored by fluoroscopy (\texttt{num\_major\_vessels}), which is associated with a relative increase in the classification error of around 87\% percent after permutation, the type of chest pain experienced (\texttt{chest\_pain}), and the thallium stress test results (\texttt{thal}). From a medical perspective, it is reasonable to assume that the number of major coronary arteries that show significant narrowing or blockage when viewed through fluoroscopy is an important indicator for the presence of heart disease. The same holds for the type of chest pain experienced and the thallium imaging stress test, which assesses blood flow to the heart muscle.

In Fig.~\ref{fig:iml-global-model-agnostic}\,(b), the PDP for \texttt{age} shows that the probability of presence of heart disease increases most drastically between the ages of 52 and 60. In contrast, it decreases between the ages of 30 and 43 according to the PDP plot. This is not very coherent, since younger people in general should have a lower risk of heart disease. There are two apparent reasons for this phenomenon. 
For one, not many data points are available before the age of 40 and after 70, so the PD estimates are not reliable in those areas. 
Additionally, higher rates of disease are observed for the younger people in this study than in the general public. 
This is most likely due to selection bias in the study. For the categorical feature denoting the results of the thallium stress test (\texttt{thal}), the PDP is a barplot. 
As expected, the absence of significant abnormalities or defects in the blood flow to the heart muscle (\texttt{thal=normal}) is associated with the lowest marginal effect on the probability of heart disease being present, whereas having some kind of defect (\texttt{thal=fixed defect} or \texttt{thal=reversable defect}) is associated with a higher probability of heart disease.

The $y$-axis values in the ALE plots in Fig.~\ref{fig:iml-global-model-agnostic}\,(c) represent the main effect of the feature at a specific value, relative to the mean prediction of the dataset. 
For instance, an ALE approximation of 0.018 at an \texttt{age} of 60 indicates that the probability of a heart disease being present increases by 0.018 percent points when an individual is 60 years old, compared to the average predicted probability of a heart disease being present. For both ALE plots, only small changes in shapes compared to the PDP plots are observed, despite the strong correlation of \texttt{age} and \texttt{thal} with other features in the dataset. 
ALE and PDP plots may show similar patterns despite the presence of correlated features when the correlated features have similar effects on the target variable and do not interact strongly with each other. Yet, it could also be indicative of the model not adequately capturing the feature correlation.

\newpage
\subsection{Model-specific Interpretability Methods}\label{sec:model-specific-iml}

Alongside model-agnostic methods, several model-specific techniques have been developed, specially designed for certain model classes. These approaches not only leverage the relationships between features and predictions but also incorporate internal components of the model, such as learned weights in a neural network. A closer look inside the black box can reveal additional information and insights that model-agnostic methods may overlook and bypass time-consuming estimation or optimization procedures. The following two sections will explain model-specific IML methods designed for tree ensemble learners and neural networks.

\subsubsection{Model-specific Interpretability Methods for Tree Ensemble Learners}\label{sec:model-specific-iml-trees}

While individual decision trees lack accuracy, they are convincing due to their ease of interpretability. Tree ensemble learners improve accuracy, but it is much harder to explain their individual predictions. As they are among the most popular machine learning models used in practice, several global and local model-specific interpretability methods have been developed for tree ensemble learners. Global methods include surrogate decision trees, tree ensemble attribution, representative trees, and feature importance, with the latter two being the most popular. Less attention has been paid to local methods in the past, yet with the development of TreeShap and related methods, they have risen significantly in popularity over the last few years. Both local and global tree ensemble learner interpretability methods remain an active area of research. In the following, a brief introduction to the most widespread methods is given.

\textbf{Variable importance measures (VIMs)} are a standard approach to gain insight into a random forest model. The most widely used importance measures are the \emph{impurity importance}\index{impurity importance} or \emph{mean decrease of impurity}\index{mean decrease of impurity} (MDI) and the \emph{permutation feature importance} (PFI) or \emph{mean decrease of accuracy}\index{mean decrease of accuracy} (MDA) \citep{breiman2001random}. MDI adds up the weighted impurity decreases for all nodes in which a feature is used for splitting and averages them over all trees in the forest. The impurity weights are the proportions of samples reaching the corresponding internal nodes. In simple terms, it measures how much each feature contributes to the overall reduction in impurity of the decision tree nodes when making splits during the training process.

MDI measures are simple and fast to compute, however, they are biased in favor of features with many possible split points and balanced category frequencies. Hence, despite being computationally expensive in comparison, MDA methods are more popular. MDA and PFI have been originally developed for random forests but are also applicable to a large variety of models; thus, their computation is explained in Sec.~\ref{sec:model-agnostic-global-iml}. Albeit prevalent due to their simplicity, VIMs exhibit non-negligible drawbacks, including their inability to capture the context and directionality of feature effects or feature interactions. 

\textbf{Representative trees}\index{representative tree} simplify a complex ensemble of decision trees to a few or one representative tree, making it easier to observe common tree structures, the importance of specific features and interactions. The fundamental concept underlying representative tree algorithms typically encompasses assessing tree similarity with a distance metric. Distance measures may cover different aspects of similarity such as the similarity of predictions, clustering in the terminal nodes, the selection of splitting features or the level and frequency at which features are selected in the trees and ensembles. When selecting a singular tree as the representative, it is typically the one exhibiting the highest mean similarity with respect to all other trees. Alternatively, clustering algorithms may be employed to identify multiple representative trees from distinct clusters \citep{laabs2023identification}.

The \textbf{TreeSHAP algorithm}\index{TreeSHAP} enables direct measurement of local feature interaction effects, which can be aggregated to facilitate the understanding of the global model structure \citep{lundberg2020local}. 
It is a variant of the Shapley values algorithm (Sec.~\ref{sec:model-agnostic-local-iml}), specifically designed for the computation of feature attributions in tree-based models. 
To compute the feature attribution for a given prediction and feature, the algorithm recursively traverses the tree from the root based on the feature values until a leaf node is reached. 
At each node, the algorithm calculates the contribution of the current feature to the prediction by comparing the output of the left and right child nodes, weighted by the proportion of samples that passed through that node. 
The algorithm continues recursively down the tree to compute the contributions of other features until reaching the leaf node. 
Upon computing the contributions of all features for the prediction, the Shapley values algorithm is applied to fairly attribute the contribution of each feature to the prediction of interest. TreeSHAP and its adjacent methods are powerful and flexible due to their ability to provide comprehensive feature attributions that are fast to compute. 

\subsubsection{Model-specific Interpretability Methods for Neural Networks}

Due to their size and complexity, neural networks are the most challenging to interpret among all machine learning models. As described in Sec.~\ref{sec:neuralnets}, they are hard to train due to their tangled and parameter-rich structure, consisting of multiple interconnected layers. While this complexity provides them with impressive predictive power, it also results in longer evaluation times than other machine learning models. This complicates the application of model-agnostic methods that rely on estimation or optimization procedures, making them unfeasible and too time-consuming. For this reason, numerous IML methods have been developed specifically for neural networks, which we will discuss in the following.

\begin{figure}[htb]
    \centering
    \includegraphics[width=0.9\textwidth]{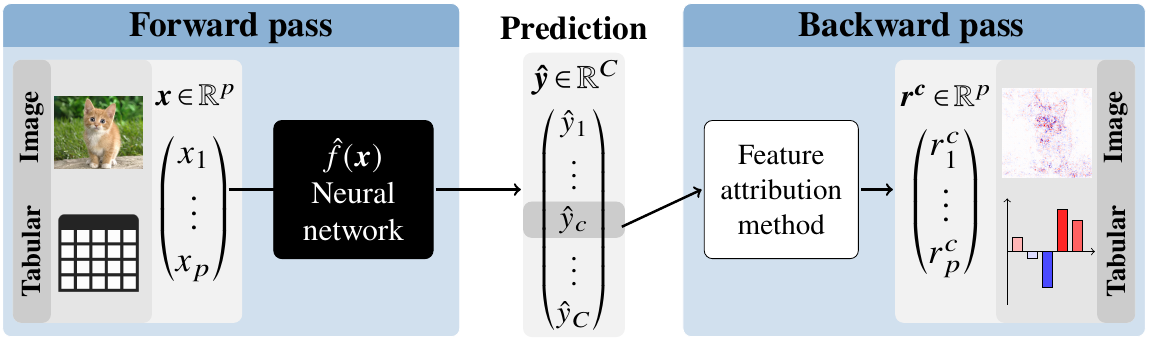}
    \caption{Basic approach of feature attribution methods.}
    \label{fig:iml-feature-attribution}
\end{figure}

A group of IML methods specifically designed for neural networks and -- due to the feature-wise findings and insights -- particularly relevant for epidemiologists are \emph{feature attribution}\index{feature attribution} methods. Feature attribution methods for neural networks summarize several local IML methods that assign to each feature the contribution or impact to a chosen model prediction. For example, suppose an input instance $\bm{x} \in \mathbb{R}^p$ is fed forward through a neural network $\hat{f}$ resulting in an output $\hat{f}(\bm{x}) = \bm{\hat{y}} \in \mathbb{R}^C$ with $C \in \mathbb{N}$ classes or regression outputs. In this case, a feature attribution method assigns relevance scores $\bm{r}_1^c, \ldots, \bm{r}_p^c$ with $\bm{r}^c \in \mathbb{R}^p$ to each of the features $\bm{x}_1, \ldots, \bm{x}_p$ of $\bm{x}$ on a chosen prediction $\hat{f}(\bm{x})_c =\hat{y}_c$ to be explained. Generally, these methods involve only a single forward pass of the model to make predictions, followed by a backward pass to generate the feature-wise explanations. This fundamental principle is illustrated in Fig.~\ref{fig:iml-feature-attribution}. In the following, we briefly present the most well-known representatives of feature attribution methods.

The \textbf{gradient} method, also known as \textbf{vanilla gradient} or \textbf{saliency maps}\index{saliency map} in the context of images, was introduced by \cite{simonyan2013} and is one of the initial and most intuitive feature attribution methods. It simply computes the gradients of the chosen prediction with respect to the input features, providing a feature-wise tendency of whether and to what extent a slight change in the input variable influences the prediction. Mathematically, this method can be described by calculating the partial derivative of the model prediction with respect to the feature $\bm{x}_j$:
\begin{align*}
    \operatorname{Grad}(\bm{x})_j^c = \frac{\partial \hat{f}(\bm{x})_c}{\partial \bm{x}_j} = \frac{\partial \bm{\hat{y}}_c}{\partial \bm{x}_j}.
\end{align*}

A straightforward modification of the vanilla gradient method results in the \textbf{gradient $\times$ input (G$\times$I)}\index{gradient $\times$ input} method introduced by \cite{shrikumar2016}. Despite its simplicity, this interpretation method leads to an approximative feature-wise decomposition of effects based on the first-order Taylor decomposition. 
It is calculated by multiplying the mathematical derivative of the chosen output prediction with the corresponding input values. When these effects are summed, they approximate the prediction, i.e.,
\begin{align*}
    \text{G$\times$I}(\bm{x})_j^c = \frac{\partial \hat{f}(\bm{x})_c}{\partial\, \bm{x}_j}\, \bm{x}_j\quad \text{with}\quad \hat{f}(\bm{x})_c \approx \sum_{j = 1}^p \text{G$\times$I}(\bm{x})_j^c.
\end{align*}

Another extension of the vanilla gradient method is \textbf{smoothed gradients (SmoothGrad)}\index{SmoothGrad} proposed by \cite{smilkov2017}, which compensates and smooths potential fluctuations or abrupt changes in the gradients arising from multiple non-linear activation functions. The method calculates gradients from randomly perturbed input copies and averages them to obtain the mean gradient. The variance of the induced noise and the number of samples are hyperparameters and can be used to adjust the neighborhood considered for the averaged gradient or to obtain a more accurate estimate. Analogous to the G$\times$I method, the gradients of individual perturbations can also be multiplied with the perturbed inputs before averaging, resulting in the \emph{SmoothGrad $\times$ input} method.

The \textbf{layer-wise relevance propagation (LRP)}\index{layer-wise relevance propagation} method proposed by \cite{bach2015} pursues a different approach than the previous gradient-based methods. In LRP, the upper-layer relevance is redistributed layer by layer, starting from the output layer to the preceding layers based on the weights and intermediate values. This redistribution continues until the input layer is reached, assigning relevance values to each feature. By taking the sum, the LRP relevances approximate the prediction, similar to the previously discussed method G$\times$I. 
The redistribution can be defined using various rules. The original simple rule, also known as LRP-z, distributes the relevances to the lower layer according to the intermediate values and weights. In addition to this rule, there are other rules or combinations of rules to distribute the upper-layer relevances to the previous layer depending on the user's focus. The most well-known rules include the $\varepsilon$-rule for sparser explanations, the $\alpha$-$\beta$-rule for a different weighting of positive or negative relevances, and the composite rule, which uses different rules depending on the position of the layer. We refer to the overview paper by \cite{Montavon2019} for a deeper look at this approach.
 
The \textbf{deep learning important features (DeepLIFT)}\index{DeepLIFT} method introduced by \cite{shrikumar2017} -- to some extent -- echoes the idea of LRP by applying rules in a layer-by-layer backpropagation fashion from the prediction back to the features. However, it incorporates a reference value $\bm{\tilde{x}}$ to compare the relevances with each other. Hence, the relevances of DeepLIFT represent the relative effect of the outputs of the instance to be explained $\hat{f}(\bm{x})_c$ and the output of the reference value $\hat{f}(\bm{\tilde{x}})_c$, i.e., $\hat{f}(\bm{x})_c - \hat{f}(\bm{\tilde{x}})_c$. Additionally, there are two rules to propagate relevances through the activation part of individual layers: the rescale rule and the reveal-cancel rule. The rescale rule simply scales the contribution to the difference from the reference output according to the value of the activation function. The reveal-cancel rule considers the average impact after adding the negative or positive contribution, showing similarities to the model-agnostic approach of Shapley values.

\textbf{Other approaches: }%
In addition to feature attribution methods, other techniques also exist that reveal insights from black box models differently without delivering feature-wise relevance values. 
However, these methods are often tailored and developed for image data and convolutional neural networks. 
One such method is Grad-CAM, which utilizes the learned patterns and high-level features in the last convolutional layers. 
Similarly to the vanilla gradient method, the gradients are computed for the last convolutional layer instead of propagating them back to the input pixels to create a class-weighted feature map. This feature map is then scaled to the original image size, highlighting connected regions rather than individual pixels. Other methods like occlusion, prediction difference analysis or meaningful perturbations also rely on this structure of convolutional layers. 
These local methods are perturbation-based approaches that aim to make single or small regions of pixels uninformative by masking, altering, or conditional sampling and afterwards quantify the resulting change in prediction probability. Consequently, they reveal prediction-sensitive regions in the input image. 
So far, only local methods have been presented, but there are also some global IML methods for neural networks. 
For example, the feature visualization approach examines which inputs -- as close as possible to the training dataset -- lead to the strongest activation in individual neurons or even entire layers within a neural network. 
These inputs are generated using various optimization and regularization techniques, not aiming to explain the entire model but rather individual components within the model or, in the context of image processing, what the model uncovers in a hidden layer.
Lastly, the network dissection and TCAV approach are both global techniques used to assess the correspondence between human-labeled concepts in images and the abstract features in the final convolutional layers of a model. The network dissection approach compares the highly activated areas in the final convolutional layers to human-understandable concepts like objects, parts, or colors. 
The TCAV method goes one step further by training a binary classifier to measure the model's sensitivity to specific concepts and comparing it with the gradients from the model's output. This allows them to evaluate and even define a test for the overall conceptual sensitivity of an entire class. It is important to note that both methods require a dataset with concept annotations, i.e., specially prepared datasets, and image data to perform the analysis. For a more detailed overview of the described methods and other approaches, we refer to \cite{samek2021explaining}.

\section{Unsupervised Learning and Generative Modeling}
\label{sec:unsupervisedgen}

While Secs.~\ref{sec:supervised}-\ref{sec:iml} introduce the principles of supervised learning, including approaches for evaluation, hyperparameter tuning and interpretation, this section presents the main ideas of unsupervised learning and the closely related field of generative modeling. Many routines for model training, evaluation and tuning cannot be easily transferred from the supervised to the unsupervised setting, so that new solutions need to be considered, which often makes these tasks more challenging.

This section starts with a general introduction to unsupervised learning, including an overview of frequently used methods and examples of their applications in epidemiology (Sec.~\ref{sec:unsupervised}). In Sec.~\ref{sec:unsupervised_generative}, the basic ideas of generative modeling are explained, including use cases, frequently used methods and evaluation approaches.

\subsection{Unsupervised Learning}
\label{sec:unsupervised}

As already introduced in Sec.~\ref{sec:supervised}, supervised learning uses data with predictive features $\bm{x} \in \mathbb{R}^p$ and a corresponding target variable $y \in \mathbb{R}$. Opposed to that, in \emph{unsupervised learning}\index{unsupervised learning}, models are trained on data with no target variable $y$ or treating $y$ the same as the features $\bm{x}$ without exploiting its supervising character. For the sake of simplicity, we also denote $\bm{x}$ as the input data in the latter case, which then already include $y$. Additionally, we assume throughout this section that all data follow the distribution of a random variable $X$ with density $p(\bm{x})$.

Methods for unsupervised learning are frequently applied during exploratory data analysis and offer valuable insights or facilitate subsequent analysis and decision-making. They can also function as a preprocessing stage for supervised learning tasks by extracting meaningful data representations. The underlying goal of unsupervised learning is to reveal the inherent structure, similarities, or interdependencies within the data by directly inferring properties of the joint density $p(\bm{x})$ without any prior knowledge or guidance. As the dimension of $X$ can be very high, these tasks can often be more challenging than supervised ones, where the main interest lies in finding properties of the usually low-dimensional conditional density $p(y|X = \bm{x})$. On top of that, without the supervision by a “teacher” $y$, unsupervised learning lacks obvious choices of evaluation measures in contrast to classification or regression in supervised learning.

With the definition we provided above for supervised and unsupervised learning, we present only one possible distinction between these terms. However, it is important to note that the terminology used in the literature is not consistent \citep{dangeti2017slusl_label, Hastie_elem_stat_learning_2017_long}. For instance, both terms are often used interchangeably with \emph{discriminative}\index{discriminative modeling} and \emph{generative modeling}, respectively \citep{babcock2021discgen_label, ng2001supgen}. 
In other cases, supervised and unsupervised learning is purely defined by the existence of labels and their utilization for training, whereas discriminative modeling and generative modeling refer to the type of modeled underlying density, i.e., the conditional density $p(y|X = \bm{x})$ or the joint density $p(\bm{x})$ \citep{bishop2006slusllabel_discgendistr}. This way, also supervised-generative models, e.g., naïve Bayes \citep{ng2001supgen}, and unsupervised-discriminative models \citep{dosovitskiy2014unsupdiscr} exist.

There are various techniques used in unsupervised learning to detect relationships between instances or between features. In the following, we briefly explain the most popular methods \citep{Hastie_elem_stat_learning_2017_long,dangeti2017slusl_label,bishop2006slusllabel_discgendistr}:\\

\textbf{Clustering} \index{clustering} algorithms group similar data points together based on their proximity in the dataset with respect to their features. A popular clustering algorithm is \emph{\mbox{$k$-means}}\index{k-means}, which aims to partition the data into a predefined number $k$ of clusters. It starts by randomly initializing $k$ cluster centroids and iteratively updates them to minimize the distance between data points and centroids. After that, each data point is assigned to the nearest centroid based on Euclidean distance.
In epidemiology, disease outbreak investigation is one of numerous use cases for clustering to identify geographical clusters of cases in order to find potential sources of infection and guide control measures \citep{hussein2021epiCluster}.

\textbf{Dimensionality reduction}\index{dimensionality reduction} methods aim to reduce the number of input features while preserving the essential information. This can have several advantages, including computational efficiency, alleviating the curse of dimensionality, noise reduction, and facilitating data visualization. However, by reducing the dimensionality, some information may be lost, and the interpretation of the transformed features can be challenging. \emph{Principal component analysis}\index{principal component analysis} (PCA) is one of the most widely used dimensionality reduction techniques. It identifies a new set of orthogonal variables, called \emph{principal components}, that capture most of the variance in the data. These components are ranked in order of their explained variance, allowing for a reduction in dimensionality while retaining the most important information.
Dimensionality reduction techniques are for instance applied for biomarker identification: Researchers often measure multiple biomarkers to assess disease risk or progression. As a result of dimensionality reduction approaches, patterns or components that capture the highest variations in the biomarker profiles are identified \citep{taguchi2013epiPCA}.

\textbf{Anomaly detection}\index{anomaly detection} is a technique used to reveal rare or unusual data points or patterns that deviate significantly from the norm or expected behavior. Anomalies can represent data points that are of interest due to their unique properties, as well as potentially indicating errors in the data collection process. \emph{One-class support vector machines}\index{one-class support vector machine} \citep{scholkopf19991SVM} and \emph{isolation forests}\index{isolation forest} \citep{liu2008isolationForest} are methods specifically designed for this purpose, which learn the normal behavior of the data and flag data points that do not conform to the learned patterns as anomalies.
Both methods have been applied by \cite{nagata2021epiAD} to detect overdose and underdose prescriptions and hence prevent life-threatening events and diminished therapeutic effects by wrong medication.

\textbf{Density estimation}\index{density estimation} approaches explicitly model the joint density $p(\bm{x})$, while the previously described techniques try to infer certain properties of this density. Besides traditional parametric and non-parametric approaches such as \emph{maximum likelihood estimation} and \emph{kernel density estimation}\index{kernel density estimation}, respectively, there are popular modern methods such as \emph{Guassian mixture models}\index{Gaussian mixture model}, \emph{normalizing flows} \citep{rezende2015NF}, and \emph{variational diffusion models} \citep{kingma2021variationalDiff}, which show better performance in higher dimensions than the classical approaches.
An example of classical density estimation in epidemiological research can be found in \cite{bithell1990epiKDE}, where kernel density estimation is applied to estimate a relative risk function over geographical regions for childhood leukemia. \cite{ausset2021epiNF} demonstrate the utility of normalizing flows for conditional density estimation tasks in survival analysis which is of particular use in individualized medicine.

\subsection{Generative Modeling}
\label{sec:unsupervised_generative}

Generative modeling\index{generative modeling} is an important application of density estimation. Generative models either explicitly or implicitly approximate the underlying data distribution and are able to draw realistic samples from this distribution. \emph{Explicit} generative models directly learn the data distribution $p(\bm{x})$ and are usually trained by maximizing the likelihood of the observed data. As opposed to that, \emph{implicit} generative models do not explicitly model the data distribution and hence do not allow for explicit likelihood calculations. Instead, they are trained to generate new samples $\bm{x}^\ast \sim X$ by learning a mapping from a random noise vector to the data space with an objective function that indirectly encourages realistic sample generation.

\subsubsection{Use Cases for Generative Modeling}
\label{sec:unsupervised_generative_usecases}

Generative modeling can be applied to many different modalities such as image, audio, text or tabular data, and has an enormous amount of different modality-specific use cases. Recently, especially chatbot and image generation applications such as ChatGPT \citep{OpenAI2023GPT4} and DALL-E \citep{Ramesh2022DALLE} have gained worldwide popularity.

In epidemiology, health records are typically of tabular nature or consist of image data, e.g., X-rays, CT or MRI scans. Applications of particular interest for this field are:
\begin{itemize}
    \item \textbf{Missing data imputation}: Manually gathered data from examinations or surveys are prone to including erroneous and missing data entries. Generative models can sample likely values for these cases according to the learned underlying distribution so that these data points do not need to be excluded from the dataset. 
    \item \textbf{Data augmentation and data balancing}: Deep learning methods are state-of-the-art in applications for image data such as cancer detection or organ segmentation but often require a vast amount of data to perform well. Augmenting the training data with synthetic samples can help increasing the model performance and robustness.
    When dealing with rare outcome events, a common issue for many supervised methods is class imbalance. This can be tackled by rebalancing the dataset using synthetic data.
    \item \textbf{Privacy-preserving data synthesis}: There is a high demand for research in health-related disciplines, but personal health data in many cases are hardly accessible due to strict data protection laws. Synthetic data with privacy-preserving guarantees can be a way to allow data analysis and scientific research without access to original patient records.
\end{itemize}

\subsubsection{Methods for Generative Modeling}
\label{sec:unsupervised_generative_methods}

There is a wide variety of different methodical frameworks in generative modeling, all with different strengths and weaknesses. Especially with the rise of deep learning, methods such as variational autoencoders \citep{kingma2014VAE} and generative adversarial networks \citep{goodfellow2014GAN} on image generation and language processing tasks have gained immense popularity and public attention also outside of machine learning research. However, artificial neural networks have not yet been able to achieve the same overwhelming success in generative modeling of tabular data as for other modalities such as image or text generation. Like in supervised tasks, tree-based methods perform well and often even outperform deep learning methods while keeping the required amount of data and tuning efforts low \citep{grinsztajn2022why, borisov2022DLTabular, watson2023ARF}.

It is very common for many types of generative models to include some kind of action-counteraction pair like "encoding and decoding networks" \citep{kingma2014VAE, vaswani2017Transformer}, "normalizing and inverse flows" \citep{rezende2015NF}, "diffusion and denoising" \citep{ho2020DPM} or even adversarial training with a "generator-discriminator" pair \citep{goodfellow2014GAN, watson2023ARF} in order to learn a $d$-dimensional latent space representation $\bm{z} \in Z \subseteq \mathbb{R}^d$ of the data instance $\bm{x}$ and a mapping from latent $Z$ to original data space $\mathbb{R}^p$. Like this, data can be sampled using a very basic distribution in the latent space and then be transformed to realistic synthetic copies $\bm{x}^\ast$ in the data space.

\begin{figure}
    \centering
    \includegraphics[width=0.9\textwidth]{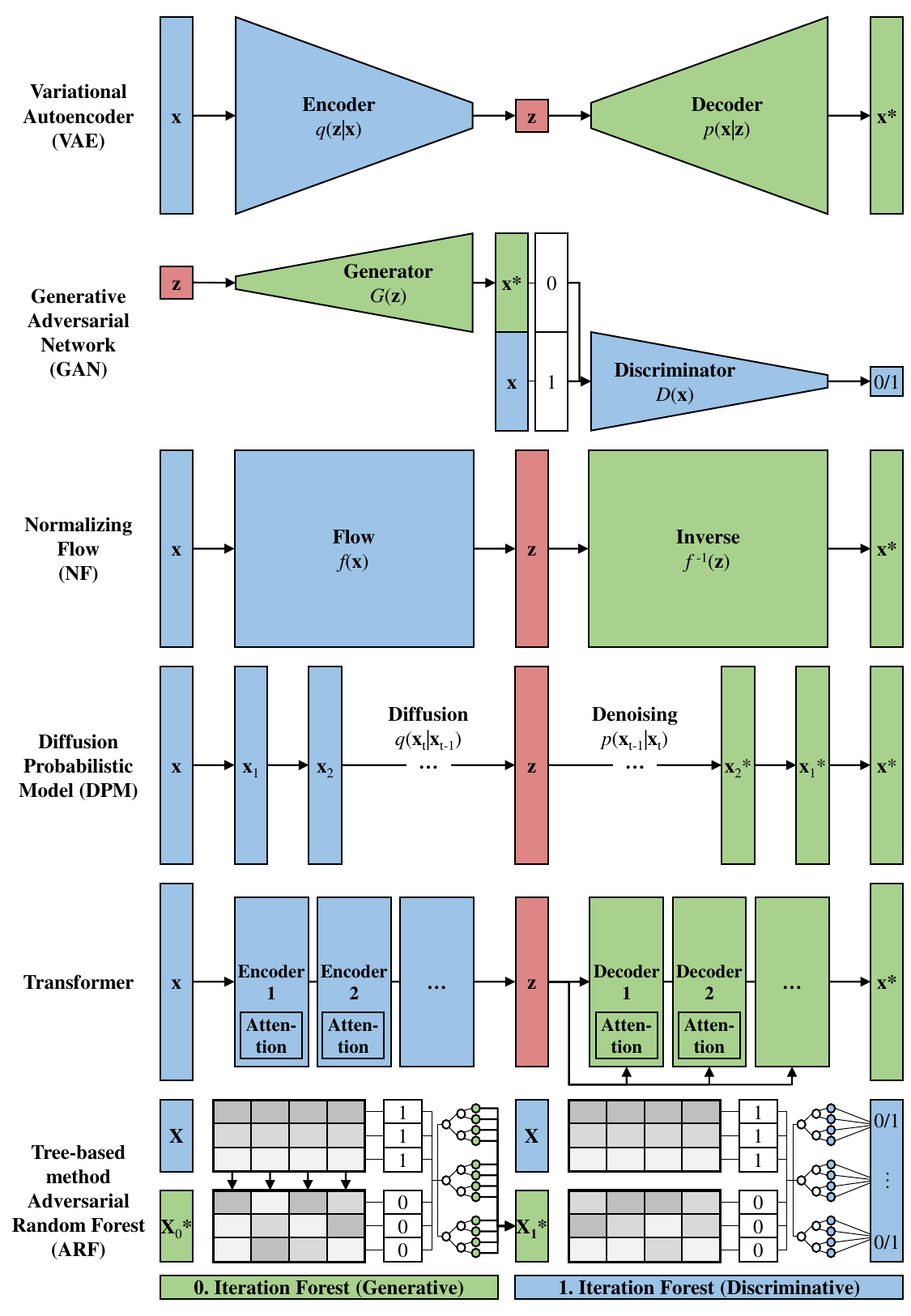}
    \caption{Visual comparison of different methods for generative modeling. Action-counteraction pairs are colored in green and blue, respectively, latent space representations in red.}
    \label{fig:genModels}
\end{figure}

A basic overview covering the most important fundamental methods is given in the paragraphs below. In addition, Fig.~\ref{fig:genModels} provides a visual comparison and highlights the action-counteraction pairs of each method. For a more comprehensive overview, we refer to \cite{foster2023overviewGEN1}.

\textbf{Variational autoencoders (VAEs)} \index{variational autoencoder} \citep{kingma2014VAE} learn to generate realistic synthetic data by approximating the underlying data distribution, maximizing a log-likelihood lower bound, also called evidence lower bound (ELBO). They combine elements from both autoencoders and probabilistic modeling. An autoencoder is a type of neural network that learns to encode input data into a lower-dimensional latent space representation and then decode it back to reconstruct the original input. However, in a VAE, the encoded inputs no longer correspond to points in the latent space but rather to the parameters of a latent space distribution from which an element is then sampled. In practice, the latent space distribution is typically assumed to be multivariate Gaussian. In order to generate synthetic data with VAEs, samples are drawn from the latent space distribution and transformed by the decoder into the final output. VAEs are known for performing well on image synthesis tasks, especially due to their great capability to capture the diversity of the original data.

\textbf{Generative adversarial networks (GANs)} \index{generative adversarial network} \citep{goodfellow2014GAN} are implicit generative models that are able to generate high quality new samples that resemble the training data, making them well-suited for tasks such as image synthesis, text generation, and more. At a high level, GANs consist of two main components: a generator network and a discriminator network. The generator's role is to generate synthetic samples from random latent space noise that mimic the training data, while the discriminator acts as a judge to distinguish between real and generated samples. The two networks are trained together in a competitive manner, which is also called adversarial training or minimax game: As the training progresses, the generator learns to produce samples that are increasingly realistic, trying to fool the discriminator, which in turn becomes better and better at distinguishing real from synthetic samples. This process results in a zero-sum game between the two networks, where they both improve iteratively. The training of GANs can be challenging, as it involves finding a Nash equilibrium, where the generator produces samples that are indistinguishable from real data for the discriminator. Often, well-trained GANs can achieve great results in terms of quality but struggle to recover the original dataset's diversity due to mode collapse.

\textbf{Normalizing flows (NFs)}\index{normalizing flow} \citep{rezende2015NF} explicitly model the underlying data distribution by learning a transformation from a simple probability distribution to the target distribution through a composition of invertible mappings using likelihood maximization. The simple initial latent space distribution acts as a source of randomness and serves as a starting point for sample generation. Usually, a standard Gaussian distribution is used in practice. In theory, NFs have the advantage of being fully invertible, which means that both sampling and density estimation are possible. As inversion tasks can be difficult to compute, the requirement for tractable inverses is often weakened and only one direction of NFs is directly modeled depending on the purpose: the normalizing direction for density estimation, the denormalizing direction for generative modeling.

\textbf{Diffusion probabilistic models (DPMs)}\index{diffusion probabilistic model} \citep{ho2020DPM} are currently state-of-the-art for image synthesis with respect to both quality and diversity of the generated data. Moreover, DPMs have shown great performance in pure density estimation tasks. They consist of a diffusion (forward) and a sampling or denoising (backward) process and are trained maximizing a lower bound of the data likelihood (like VAEs). During the diffusion process, a Markov chain gradually adds noise to the original data until they are indistinguishable from pure random noise. The model learns a Markov chain of denoising steps to recover the original data after diffusion has been applied to them. A drawback of DPMs is considerably slow sampling due to the sequential nature of the denoising process where often hundreds to thousands of steps are needed in order to generate high fidelity samples. However, there has been progress to speed up sampling significantly, e.g., as described in \cite{nichol2021DPM}.

\textbf{Transformers}\index{transformer} \citep{vaswani2017Transformer} are a type of deep learning architecture that has gained remarkable popularity in the field of generative modeling, particularly in language processing tasks where transformer-based large language models (LLMs) such as GPT-4 \citep{OpenAI2023GPT4} are state-of-the-art, but also in image generation. Transformers consist of stacked encoder-decoder blocks. The encoders take the input data and transform it into a set of abstract representations, while the decoders generate output based on those representations. One of the significant advantages of transformers is their ability to model long-range dependencies effectively, as opposed to traditional convolutional neural networks (CNNs) \citep{krizhevsky_2012} and recurrent neural networks (RNNs) \citep{hochreiter1997LSTM}. The key components equipping transformers with this ability are the attention mechanisms self-attention and encoder-decoder-attention, which allow the model to weigh the importance of different parts of the input when generating output. Like this, the model is able to capture dependencies across features, for example between pixels in an image or words in text, regardless of their position. Transformer architectures typically require a lot of data and a high amount of parameters to perform well and hence need a very powerful infrastructure for training.

\textbf{Tree-based generative methods} \citep{correia2020GF, watson2023ARF} have demonstrated the ability to perform well on tabular data and even outperform current deep learning methods in many cases without the need for extensive hyperparameter tuning. Adversarial random forests (ARFs)\index{adversarial random forest} \citep{watson2023ARF} are a type of explicit tree-based generative models. They use unsupervised random forests \citep{shi2006URF}\index{unsupervised random forest} to learn dependencies across features by attempting to distinguish real data from naïvely constructed synthetic data with independent features. This process can be iterated with fresh naïve synthetic data drawn from the forest's leaves, leading to an adversarial approach where the higher iteration forest (discriminating iteration) evaluates the previous one (generating iteration), respectively, until synthetic data cannot be distinguished from real data, i.e., the accuracy of the discriminating forest falls under 50\%. At this stage, local independence with regard to the features can be assumed for the original data within the boundaries of the generating forest's leaves, since the corresponding discriminating forest is not able to distinguish them from the naïve synthetic data. This assumption then allows for feature-wise independent density estimation and data synthesis and hence breaks down a high-dimensional problem into several one-dimensional ones within the leaves. 

\subsubsection{Evaluation of Generative Models}
\label{sec:unsupervised_generative_evaluation}

As already stated in Sec.~\ref{sec:unsupervised}, evaluating unsupervised methods in general and generative models in particular is not straightforward, since there is no "obvious" target as for classification or regression tasks in supervised learning. Usually, there are two different properties one is interested in: The quality of the synthetic samples, also called fidelity, and the diversity of the generated output, which measures how successfully the variation within the original data has been recovered.

There exists a variety of different measures assessing quality and diversity for different types of models and data modalities, all with different advantages and drawbacks. Research in this field has been very active within the last years and new measures are proposed frequently. However, none of these truly stand out from all the others yet. Visual inspection of generated outputs is still a valid and integral assessment method for image synthesis. For tabular data, there is no such innate option for generative model assessment.

The following paragraphs aim to provide a non-comprehensive overview of currently used measures for synthetic tabular and image data:\\

\textbf{(Log-)likelihoods}: This measure is only available for explicit generative models since they are able to output likelihoods for given data points. This can be used for likelihood-based model comparisons as in the field of density estimation. However, \cite{theis2016evaLL} show that higher quality in data generation does not necessarily imply higher likelihoods and vice versa.

\textbf{Precision-recall-based measures}: This concept provides two metrics to assess the performance of generative models and is inspired by the conforming measures for binary classification evaluation introduced in Sec.~\ref{sec:eval-measures}: \emph{Precision} measures the fidelity of synthetic data, \emph{recall} their diversity. Early approaches quantify these metrics by solely comparing the empirical supports of the true and the modeled distribution \citep{sajjadi2018evaPR}: Precision is defined as the proportion of the support of the generated data distribution that also lies in the support of the true data distribution. Conversely, recall is defined as the proportion of the support of the true distribution that is in the support of the generated distribution as well. This aligns with definitions given in Sec.~\ref{sec:eval-measures} when the intersection of the supports of both distributions are regarded as the true positives. The pure reliance on support-based comparisons is highly affected by outliers and can moreover lead to perfect scores for very different distributions for real and synthetic data. More recent work on this topic has therefore attempted to mitigate this by also considering how densely regions of the supports are packed \citep{alaa2022evaalphabetaPR}. As the true distribution is usually not available in real-world applications and the generated data distribution is only available for explicit models, nearest neighbor-based methods are often used in practice to approximate these measures.

\textbf{Classifier 2-sample test (C2ST)}\index{classifier 2-sample test} \citep{lopez2017C2ST}: C2ST is quantified by the ability of a binary classifier to distinguish real data and synthetic data. The chosen classifier should be high-performing on the modality of the data in order to have trustworthy results. Gradient-boosted trees and convolutional neural networks (CNNs) are reasonable choices for tabular and image data, respectively. First, the dataset is split into a train and test set. The generative model is trained on the training data and used to generate a synthetic test set of the same size as the original test set. These two test sets are provided with labels "real" and "synthetic" and then merged. The resulting labeled dataset is again split into a train and test set in order to train and evaluate the selected binary classifier. The performance of this discriminant classifier, often measured by the area under the curve (AUC), is used as the C2ST statistic.

\textbf{Machine learning efficacy}\index{machine learning efficacy} \citep{choi2017evaMLeffic}: This measure, also referred to as \emph{machine learning utility}\index{machine learning utility}, is only applicable for labeled datasets and if a connected supervised task, i.e., classification or regression, is available for the data. The dataset is first split into a training and test set. After that, the generative model is applied to training data to produce synthetic copies. A supervised task is then performed by the same predefined set of learners both on the synthetic and the original training dataset. The performances of both learned models are evaluated on the original test data. The model learned using the original data serves as a performance upper bound. This evaluation method does not necessarily reflect high generative performance, as \cite{zein2022evaXGBOOST} demonstrate.

\textbf{Fréchet inception distance (FID)}\index{Fréchet inception distance} \citep{heusel2017evaFID}: FID is only available for image data and very popular in this domain. It measures the similarity between the distributions of generated samples and real data based on the activations of a pre-trained inception network \citep{szegedy2016evaInception} with an optimal value of 0 for identical distributions. FID is a combined score for quality and diversity, lower FID scores indicate better performance of the evaluated model.

\subsubsection{Remarks on the Privacy of Synthetic Data}
\label{sec:unsupervised_generative_privacy}

It is a fallacy to assume that synthetic data always guarantee privacy preservation and that no implications on the training data or re-identifications can be made. \emph{Membership inference attacks} (MIA) \index{membership inference attack} attempt to identify whether some individual's data was used for model training and have already been performed successfully against generative models \citep{chen2020MIAGAN}. Especially for medical records and personal health data, it is essential to prevent any kind of patient-level information leakage.

One of the main frameworks to mitigate the risk of MIAs is called \emph{differential privacy} \index{differential privacy} \citep{dwork2014DP}. An algorithm is said to be differentially private if the difference of its outputs' probabilities remains within a predefined budget, regardless of whether any single individual's data are included in or excluded from the dataset. This is often achieved by the introduction of random noise to the computations which is calibrated based on the desired privacy budget. However, there is a trade-off between privacy protection and utility of the analysis results: While adding a higher amount of noise lowers the risk of information leakage, the resulting synthetic data might not be useful anymore for further analysis. Finding this balance is challenging, particularly for high dimensional data.

Distance-based measures based on nearest neighbor computations are often used to assess the privacy of generative models. The \emph{distance to closest record} \index{distance to closest record} (DCR) \citep{park2018DCR} measures the distance between any synthetic record and its closest corresponding real neighbor in the training data. The \emph{nearest-neighbors distance ratio} (NNDR) \index{nearest-neighbors distance ratio} \citep{lowe2004NNDR} measures the ratio between the distance for the closest and second closest real neighbor in the training set for any synthetic record. This approach puts more emphasis on the protection of outliers, as these individuals are most vulnerable for MIAs.
These measures are usually both calculated for the synthetic data and a test set with real data that was not used for model training. When comparing the outcomes for both datasets, the synthetic data should not be closer to the training data than the test data.

\subsubsection{Data Example}
To conclude this section, we show the application of generative modeling for tabular data using the ARF method and the heart dataset. As stated in Sec.~\ref{sec:unsupervised_generative_methods}, tree-based methods such as ARF are well-suited for this task and often directly applicable. In order to create a synthetic heart dataset of the same size like the original one, three simple steps need to be performed: Growing an adversarial random forest, fitting distributions locally in its leaves, and sampling from the resulting global mixture distribution. Note that the column names of the final output are shortened.

\label{sec:unsupervised_generative_example}
\begin{verbatim}
library(arf)

# Grow adversarial random forest
arf <- adversarial_rf(heart)
#> Iteration: 0, Accuracy: 57.91%
#> Iteration: 1, Accuracy: 38.39%

# Calculate density parameters
params <- forde(arf, heart)

# Create synthetic data
syn_heart <- forge(params, n_synth = nrow(heart))
head(syn_heart)
#>    age    sex rbp  sc   fbs mhr eia nmv              thal
#> 1:  45 female 137 215 false 129  no   0            normal
#> 2:  55   male 118 257  true 141 yes   1      fixed defect
#> 3:  57   male 100 207  true 145  no   1            normal
#> 4:  38   male 135 312 false 127  no   0 reversable defect
#> 5:  54 female 124 280 false 160  no   0            normal
#> 6:  40 female 120 279 false 157  no   0            normal
#>          chest_pain                  resting_ecg    ST_slope
#> 1: non-anginal pain left ventricular hypertrophy        flat
#> 2:  atypical angina                       normal        flat
#> 3: non-anginal pain                       normal        flat
#> 4:  atypical angina                       normal downsloping
#> 5:  atypical angina                       normal  upslopling
#> 6:  atypical angina                       normal        flat
#>         hd ST_d
#> 1:  absent  1.8
#> 2: present  0.7
#> 3: present  1.9
#> 4: present  1.6
#> 5:  absent  1.3
#> 6:  absent  1.6
\end{verbatim}

\section{Conclusions}
\label{sec:discussion}

In this chapter, we laid the methodological foundations for successfully applying machine learning in epidemiology. We covered the principles of supervised and unsupervised learning, discussed the most important learners, strategies for model evaluation and hyperparameter optimization and introduced interpretable machine learning. However, it is important to note that machine learning is not a panacea and there are several important considerations that researchers should keep in mind when using these methods. One such aspect is the quality and representativeness of the data. Machine learning algorithms can only be as good as the data they are trained on, and if, e.g., certain groups are under-represented in the training data, the model's output may be unreliable or misleading. That is not new for epidemiologists, but it is important to note that more data and machine learning rarely solve such problems. It is therefore as crucial as ever to ensure that the data used for analysis are of high quality and represent the population of interest. A related issue are the ethical implications of using machine learning methods in epidemiological research. Machine learning algorithms have the potential to perpetuate and amplify existing biases in the data, which, e.g., could have negative consequences for marginalized populations. Recently, such fairness considerations grew into a dedicated subfield about \textit{fair machine learning}. 

Selecting an appropriate, or even the best, learner for a given task is not straightforward. While it has been tried in many benchmarks, no single best learner for all tasks has emerged. Nevertheless, some general patterns can be observed. First, deep learning has revolutionized the analysis of image, speech and text data, and is clearly the first choice for tasks involving such data. However, deep learning is a broad field and a researcher still has to find a suitable neural network architecture (see Sec.~\ref{sec:neuralnets}) or even use a pre-trained model (e.g., \cite{he_2016}). Second, when working with tabular data, tree-based methods perform very well and often outperform deep learning \citep{grinsztajn2022why}, while being computationally faster and easier to use. In many cases, a random forest is a good starting point because it is fast, easy to tune and performs well \citep{couronne2018random}. For a final model, well-tuned gradient-boosted trees often perform slightly better, but are more prone to overfitting and thus require careful tuning. While, in this chapter, we focused on tree-based learners and neural networks, other learners such as \emph{$k$-nearest neighbors}\index{k-nearest neighbors} or \emph{support vector machines}\index{support vector machines}, are still useful and perform well in many settings; we refer to \cite{Gareth_intro_stat_learning_2021_short} and \cite{Hastie_elem_stat_learning_2017_long} for a detailed description.

Generally, machine learning methods are not a replacement for traditional statistical methods. Machine learning works extremely well for supervised learning, i.e., for prediction tasks, with large and complex data and also excels in unsupervised learning, e.g., for generation of synthetic data. However, parameter estimation is still difficult with machine learning. While interpretable machine learning helps with understanding the inner workings of a model, it typically does not provide estimates for, e.g., treatment or interaction effects. To this end, a promising direction are methods of causal machine learning such as targeted learning \citep{van2011targeted} or double machine learning \citep{chernozhukov2018double}. 

%
%

\newpage
\bibliographystyle{bibstyle}
\bibliography{refs}

@article{grinsztajn2022why,
title={Why do tree-based models still outperform deep learning on typical tabular data?},
author={Leo Grinsztajn and Edouard Oyallon and Gael Varoquaux},
 journal = {Advances in Neural Information Processing Systems},
 pages = {507--520},
 volume = {35},

year={2022}
}

@book{Gareth_intro_stat_learning_2021_short,
  title={An Introduction to Statistical Learning},
  author={Gareth, James and Witten, Daniela and Hastie, Trevor and Tibshirani, Robert},
  year={2021},
  publisher={Springer}, 
  address = {New York}
}

@book{Hastie_elem_stat_learning_2017_long,
  title={The Elements of Statistical Learning},
  author={Hastie, Trevor and Tibshirani, Robert and Friedmann, Jerome},
  year={2017},
  publisher={Springer},
 address = {New York}
}

@article{freund1997decision,
  title={A decision-theoretic generalization of on-line learning and an application to boosting},
  author={Freund, Yoav and Schapire, Robert E},
  journal={Journal of Computer and System Sciences},
  volume={55},
  number={1},
  pages={119--139},
  year={1997},
  publisher={Elsevier}
}

@article{breiman2001random,
  title={Random forests},
  author={Breiman, Leo},
  journal={Machine Learning},
  volume={45},
  pages={5--32},
  year={2001},
  publisher={Springer}
}

@inproceedings{xgboost_chen2016,
  title={{XGBoost}: A Scalable Tree Boosting System},
  author={Chen, Tianqi and Guestrin, Carlos},
  booktitle={Proceedings of the 22nd {ACM} {SIGKDD} International Conference on Knowledge Discovery and Data Mining},
  year={2016},
  pages = {785--794}
}

@book{goodfellow_2016,
    title={Deep Learning},
    author={Ian Goodfellow and Yoshua Bengio and Aaron Courville},
    publisher={MIT Press},
    address = {Cambridge},
    year={2016}
}

@article{lecun2015deep,
  title={Deep learning},
  author={LeCun, Yann and Bengio, Yoshua and Hinton, Geoffrey},
  journal={Nature},
  volume={521},
  number={7553},
  pages={436--444},
  year={2015},
  publisher={Nature Publishing Group UK London}
}

@inproceedings{he_2016,
  title={Deep residual learning for image recognition},
  author={He, Kaiming and Zhang, Xiangyu and Ren, Shaoqing and Sun, Jian},
  booktitle={Proceedings of the IEEE Conference on Computer Vision and Pattern Recognition},
  pages={770--778},
  year={2016}
}

@article{kingma2014,
  title={Adam: A method for stochastic optimization},
  author={Kingma, Diederik P and Ba, Jimmy},
  journal={Preprint},
  note = {{arXiv:1412.6980}},
  year={2014}
}

@article{krizhevsky_2012,
    author = {Krizhevsky, Alex and Sutskever, Ilya and Hinton, Geoffrey E},
    journal = {Advances in Neural Information Processing Systems},
    title = {{ImageNet} Classification with Deep Convolutional Neural Networks},
    volume = {25},
    year = {2012}
}

@article{bengio_2003,
    author = {Bengio, Yoshua and Ducharme, R\'{e}jean and Vincent, Pascal and Janvin, Christian},
    title = {A Neural Probabilistic Language Model},
    year = {2003},
    volume = {3},
    journal = {Journal of Machine Learning Research},
    pages = {1137–1155}
}

@article{bottou2018,
  title={Optimization methods for large-scale machine learning},
  author={Bottou, L{\'e}on and Curtis, Frank E and Nocedal, Jorge},
  journal={SIAM review},
  volume={60},
  number={2},
  pages={223--311},
  year={2018},
  publisher={SIAM}
}

@Manual{keras_R,
    title = {keras: R Interface to 'Keras'},
    author = {JJ Allaire and François Chollet},
    year = {2023},
    note = {{R} package version 2.11.1},
    url = {https://CRAN.R-project.org/package=keras},
}

@book{japkowicz2011evaluating,
  title = {Evaluating Learning Algorithms: A Classification Perspective},
  author = {Japkowicz, N. and Shah, M.},
  year = {2011},
  publisher = {Cambridge University Press},
  address = {Cambridge},
  isbn = {978-1-139-49414-4}
}

@book{gerds2021medicalrisk,
  title = {Medical Risk Prediction Models: With Ties to Machine Learning},
  author = {Gerds, Thomas A. and Kattan, Michael W.},
  year = {2021},
  publisher = {{CRC Press}},
  address = {Boca Raton},
  isbn = {978-1-138-38447-7}
}

@article{bischl2012resamplingmethodsa,
  title        = {Resampling Methods for Meta-Model Validation with Recommendations for Evolutionary Computation},
  author       = {Bischl, B. and Mersmann, O. and Trautmann, H. and Weihs, C.},
  year         = {2012},
  journal = {Evolutionary Computation},
  volume       = {20},
  number       = {2},
  pages        = {249--275}
}

@article{bischl2023hyperparampaper,
  title={Hyperparameter optimization: Foundations, algorithms, best practices, and open challenges},
  author={Bischl, Bernd and Binder, Martin and Lang, Michel and Pielok, Tobias and Richter, Jakob and Coors, Stefan and Thomas, Janek and Ullmann, Theresa and Becker, Marc and Boulesteix, Anne-Laure and others},
  journal={Wiley Interdisciplinary Reviews: Data Mining and Knowledge Discovery},
  volume={13},
  number={2},
  pages={e1484},
  year={2023},
  publisher={Wiley Online Library}
}

@article{bergstra2011algorithms,
  title={Algorithms for hyper-parameter optimization},
  author={Bergstra, James and Bardenet, R{\'e}mi and Bengio, Yoshua and K{\'e}gl, Bal{\'a}zs},
  journal={Advances in Neural Information Processing Systems},
  volume={24},
  year={2011}
}

@book{bartz2023hyperparameter,
  title={Hyperparameter Tuning for Machine and Deep Learning with R: A Practical Guide},
  author={Bartz, Eva and Bartz-Beielstein, Thomas and Zaefferer, Martin and Mersmann, Olaf},
  year={2023},
  publisher={Springer Nature},
  address = {Singapore}
}

@article{bergstra2012random,
  title={Random search for hyper-parameter optimization},
  author={Bergstra, James and Bengio, Yoshua},
  journal={Journal of Machine Learning Research},
  volume={13},
  pages = {281--305},
  year={2012}
}

@inproceedings{ahmad2018,
  title={Interpretable machine learning in healthcare},
  author={Ahmad, Muhammad Aurangzeb and Eckert, Carly and Teredesai, Ankur},
  booktitle={Proceedings of the 2018 ACM International Conference on Bioinformatics, Computational Biology, and Health Informatics},
  pages={559--560},
  year={2018}
}

@article{goldstein2015,
  title={Peeking inside the black box: Visualizing statistical learning with plots of individual conditional expectation},
  author={Goldstein, Alex and Kapelner, Adam and Bleich, Justin and Pitkin, Emil},
  journal={Journal of Computational and Graphical Statistics},
  volume={24},
  number={1},
  pages={44--65},
  year={2015},
  publisher={Taylor \& Francis}
}

@inproceedings{ribeiro2016,
  title={Why should {I} trust you? {E}xplaining the predictions of any classifier},
  author={Ribeiro, Marco Tulio and Singh, Sameer and Guestrin, Carlos},
  booktitle={Proceedings of the 22nd ACM SIGKDD International Conference on Knowledge Discovery and Data Mining},
  pages={1135--1144},
  year={2016}
}

@article{alvarez2018,
  title={On the robustness of interpretability methods},
  author={Alvarez-Melis, David and Jaakkola, Tommi S},
  journal={Preprint},
  note = {{arXiv:1806.08049}},
  year={2018}
}

@article{strumbelj2014,
  title={Explaining prediction models and individual predictions with feature contributions},
  author={{\v{S}}trumbelj, Erik and Kononenko, Igor},
  journal={Knowledge and Information Systems},
  volume={41},
  pages={647--665},
  year={2014},
  publisher={Springer}
}

@incollection{shapley1953,
  title = {A Value for n-Person Games},
  author = {Shapley, Lloyd S},
  booktitle = {Contributions to the Theory of Games},
  editor = {Kuhn, Harold W. and Tucker, Albert W.},
  pages = {307--317},
  year = {1953},
  publisher = {Princeton University Press},
  address = {Princeton}
}

@article{lundberg2017,
  title={A unified approach to interpreting model predictions},
  author={Lundberg, Scott M and Lee, Su-In},
  journal={Advances in Neural Information Processing Systems},
  volume={30},
  year={2017}
}

@article{Chen2023,
  year = {2023},
  publisher = {Springer Science and Business Media {LLC}},
  volume = {5},
  number = {6},
  pages = {590--601},
  author = {Hugh Chen and Ian C. Covert and Scott M. Lundberg and Su-In Lee},
  title = {Algorithms to estimate {S}hapley value feature attributions},
  journal = {Nature Machine Intelligence}
}

@article{Hooker2021,
  year = {2021},
  month = oct,
  publisher = {Springer Science and Business Media {LLC}},
  volume = {31},
  number = {6},
  author = {Giles Hooker and Lucas Mentch and Siyu Zhou},
  title = {Unrestricted permutation forces extrapolation: variable importance requires at least one more model,  or there is no free variable importance},
  journal = {Statistics and Computing}
}

@book{molnar2022,
  title      = {Interpretable Machine Learning},
  author     = {Christoph Molnar},
  year       = {2022},
  publisher  = {}, 
  subtitle   = {A Guide for Making Black Box Models Explainable},
  edition    = {2},
  url        = {https://christophm.github.io/interpretable-ml-book}
}

@article{fisher2019all,
  title={All Models are Wrong, but Many are Useful: Learning a Variable's Importance by Studying an Entire Class of Prediction Models Simultaneously.},
  author={Fisher, Aaron and Rudin, Cynthia and Dominici, Francesca},
  journal={Journal of Machine Learning Research},
  volume={20},
  number={177},
  pages={1--81},
  year={2019}
}

@article{friedman2001greedy,
  title={Greedy function approximation: a gradient boosting machine},
  author={Friedman, Jerome H},
  journal={Annals of Statistics},
  pages={1189--1232},
  year={2001},
  volume={29},
  number = {5}
}

@article{apley2020visualizing,
  title={Visualizing the effects of predictor variables in black box supervised learning models},
  author={Apley, Daniel W and Zhu, Jingyu},
  journal={Journal of the Royal Statistical Society Series B: Statistical Methodology},
  volume={82},
  number={4},
  pages={1059--1086},
  year={2020},
  publisher={Oxford University Press}
}

@article{laabs2023identification,
  title={Identification of representative trees in random forests based on a new tree-based distance measure},
  author={Laabs, Bjoern-Hergen and Westenberger, Ana and K{\"o}nig, Inke R},
  journal={Advances in Data Analysis and Classification},
  doi={10.1007/s11634-023-00537-7},
  year={2023}
}

@article{lundberg2020local,
  title={From local explanations to global understanding with explainable {AI} for trees},
  author={Lundberg, Scott M and Erion, Gabriel and Chen, Hugh and DeGrave, Alex and Prutkin, Jordan M and Nair, Bala and Katz, Ronit and Himmelfarb, Jonathan and Bansal, Nisha and Lee, Su-In},
  journal={Nature Machine Intelligence},
  volume={2},
  number={1},
  pages={56--67},
  year={2020},
  publisher={Nature Publishing Group UK London}
}

@article{simonyan2013,
  title={Deep inside convolutional networks: Visualising image classification models and saliency maps},
  author={Simonyan, Karen and Vedaldi, Andrea and Zisserman, Andrew},
  journal={Preprint},
  note = {{arXiv:1312.6034}},
  year={2013}
}

@article{shrikumar2016,
  title={Not just a black box: Learning important features through propagating activation differences},
  author={Shrikumar, Avanti and Greenside, Peyton and Shcherbina, Anna and Kundaje, Anshul},
  journal={Preprint},
  note = {{arXiv:1605.01713}},
  year={2016}
}

@article{smilkov2017,
  title={Smoothgrad: removing noise by adding noise},
  author={Smilkov, Daniel and Thorat, Nikhil and Kim, Been and Vi{\'e}gas, Fernanda and Wattenberg, Martin},
  journal={Preprint},
  note = {{arXiv:1706.03825}},
  year={2017}
}

@article{bach2015,
  title={On pixel-wise explanations for non-linear classifier decisions by layer-wise relevance propagation},
  author={Bach, Sebastian and Binder, Alexander and Montavon, Gr{\'e}goire and Klauschen, Frederick and M{\"u}ller, Klaus-Robert and Samek, Wojciech},
  journal={PloS One},
  volume={10},
  number={7},
  pages={e0130140},
  year={2015},
  publisher={Public Library of Science San Francisco, CA USA}
}

@article{Montavon2019,
  title={Layer-wise relevance propagation: an overview},
  author={Montavon, Gr{\'e}goire and Binder, Alexander and Lapuschkin, Sebastian and Samek, Wojciech and M{\"u}ller, Klaus-Robert},
  journal={Explainable AI: Interpreting, Explaining and Visualizing Deep Learning},
  pages={193--209},
  year={2019},
  publisher={Springer}
}

@inproceedings{shrikumar2017,
  title={Learning important features through propagating activation differences},
  author={Shrikumar, Avanti and Greenside, Peyton and Kundaje, Anshul},
  booktitle={International Conference on Machine Learning},
  pages={3145--3153},
  year={2017},
  organization={PMLR}
}

@article{samek2021explaining,
  author={Samek, Wojciech and Montavon, Grégoire and Lapuschkin, Sebastian and Anders, Christopher J. and Müller, Klaus-Robert},
  journal={Proceedings of the IEEE}, 
  title={Explaining Deep Neural Networks and Beyond: A Review of Methods and Applications}, 
  year={2021},
  volume={109},
  number={3},
  pages={247-278}}

@article{scholkopf19991SVM,
  title={Support vector method for novelty detection},
  author={Sch{\"o}lkopf, Bernhard and Williamson, Robert C and Smola, Alex and Shawe-Taylor, John and Platt, John},
  journal={Advances in Neural Information Processing Systems},
  volume={12},
  year={1999}
}

@inproceedings{liu2008isolationForest,
  title={Isolation forest},
  author={Liu, Fei Tony and Ting, Kai Ming and Zhou, Zhi-Hua},
  booktitle={2008 eighth IEEE International Conference on Data Mining},
  pages={413--422},
  year={2008}
}

@article{kingma2021variationalDiff,
  title={Variational diffusion models},
  author={Kingma, Diederik P and Salimans, Tim and Poole, Ben and Ho, Jonathan},
  journal={Advances in Neural Information Processing Systems},
  volume={34},
  pages={21696--21707},
  year={2021}
}

@inproceedings{kingma2014VAE,
  title={Auto-Encoding Variational {B}ayes},
  author={Kingma, Diederik P and Welling, Max},
  booktitle={International Conference on Learning Representations},
  year={2014}
}

@article{goodfellow2014GAN,
  title={Generative adversarial nets},
  author={Goodfellow, Ian and Pouget-Abadie, Jean and Mirza, Mehdi and Xu, Bing and Warde-Farley, David and Ozair, Sherjil and Courville, Aaron and Bengio, Yoshua},
  journal={Advances in Neural Information Processing Systems},
  volume={27},
  pages={2672--2680},
  year={2014}
}

@inproceedings{rezende2015NF,
  title={Variational inference with normalizing flows},
  author={Rezende, Danilo and Mohamed, Shakir},
  booktitle={International Conference on Machine Learning},
  pages={1530--1538},
  year={2015},
  organization={PMLR}
}

@article{ho2020DPM,
  title={Denoising diffusion probabilistic models},
  author={Ho, Jonathan and Jain, Ajay and Abbeel, Pieter},
  journal={Advances in Neural Information Processing Systems},
  volume={33},
  pages={6840--6851},
  year={2020}
}

@inproceedings{nichol2021DPM,
  title={Improved denoising diffusion probabilistic models},
  author={Nichol, Alexander Quinn and Dhariwal, Prafulla},
  booktitle={International Conference on Machine Learning},
  pages={8162--8171},
  year={2021},
  organization={PMLR}
}

@inproceedings{vaswani2017Transformer,
  title={Attention is all you need},
  author={Vaswani, Ashish and Shazeer, Noam and Parmar, Niki and Uszkoreit, Jakob and Jones, Llion and Gomez, Aidan N and Kaiser, {\L}ukasz and Polosukhin, Illia},
  booktitle={Advances in Neural Information Processing Systems},
  volume={30},
  pages={5998--6008},
  year={2017}
}

@inproceedings{watson2023ARF,
  title={Adversarial random forests for density estimation and generative modeling},
  author={Watson, David S and Blesch, Kristin and Kapar, Jan and Wright, Marvin N},
  booktitle={International Conference on Artificial Intelligence and Statistics},
  pages={5357--5375},
  year={2023},
  organization={PMLR}
}

@incollection{rumelhart1985RNN,
  title={Learning internal representations by error propagation},
  booktitle={Parallel Distributed Processing: Explorations in the Microstructure of Cognition: Foundations}, 
  author={Rumelhart, David E and Hinton, Geoffrey E and Williams, Ronald J},
  year={1985},
  pages={318-362},
  publisher = {MIT Press},
  address = {Cambridge}
}

@article{hochreiter1997LSTM,
  title={Long short-term memory},
  author={Hochreiter, Sepp and Schmidhuber, J{\"u}rgen},
  journal={Neural Computation},
  volume={9},
  number={8},
  pages={1735--1780},
  year={1997},
  publisher={MIT press}
}

@article{shi2006URF,
  title={Unsupervised learning with random forest predictors},
  author={Shi, Tao and Horvath, Steve},
  journal={Journal of Computational and Graphical Statistics},
  volume={15},
  number={1},
  pages={118--138},
  year={2006},
  publisher={Taylor \& Francis}
}

@inproceedings{theis2016evaLL,
  author = "L. Theis and A. van den Oord and M. Bethge",
  title = "A note on the evaluation of generative models",
  year = 2016,
  booktitle = "International Conference on Learning Representations",
  month = "Apr"
}

@article{sajjadi2018evaPR,
  title={Assessing generative models via precision and recall},
  author={Sajjadi, Mehdi SM and Bachem, Olivier and Lucic, Mario and Bousquet, Olivier and Gelly, Sylvain},
  journal={Advances in Neural Information Processing Systems},
  volume={31},
  year={2018}
}

@inproceedings{alaa2022evaalphabetaPR,
  title={How faithful is your synthetic data? sample-level metrics for evaluating and auditing generative models},
  author={Alaa, Ahmed and Van Breugel, Boris and Saveliev, Evgeny S and van der Schaar, Mihaela},
  booktitle={International Conference on Machine Learning},
  pages={290--306},
  year={2022},
  organization={PMLR}
}

@article{heusel2017evaFID,
  title={{GANs} trained by a two time-scale update rule converge to a local nash equilibrium},
  author={Heusel, Martin and Ramsauer, Hubert and Unterthiner, Thomas and Nessler, Bernhard and Hochreiter, Sepp},
  journal={Advances in Neural Information Processing Systems},
  volume={30},
  year={2017}
}

@inproceedings{szegedy2016evaInception,
  title={Rethinking the inception architecture for computer vision},
  author={Szegedy, Christian and Vanhoucke, Vincent and Ioffe, Sergey and Shlens, Jon and Wojna, Zbigniew},
  booktitle={Proceedings of the IEEE Conference on Computer Vision and Pattern Recognition},
  pages={2818--2826},
  year={2016}
}

@inproceedings{lopez2017C2ST,
title={Revisiting Classifier Two-Sample Tests},
author={David Lopez-Paz and Maxime Oquab},
booktitle={International Conference on Learning Representations},
year={2017},
}

@inproceedings{zein2022evaXGBOOST,
  title={Tabular Data Generation: Can We Fool {XGB}oost?},
  author={Zein, EL Hacen and Urvoy, Tanguy},
  booktitle={NeurIPS 2022 First Table Representation Workshop},
  year={2022}
}

@inproceedings{choi2017evaMLeffic,
  title={Generating multi-label discrete patient records using generative adversarial networks},
  author={Choi, Edward and Biswal, Siddharth and Malin, Bradley and Duke, Jon and Stewart, Walter F and Sun, Jimeng},
  booktitle={Machine Learning for Healthcare Conference},
  pages={286--305},
  year={2017},
  organization={PMLR}
}

@article{borisov2022DLTabular,
  title={Deep neural networks and tabular data: A survey},
  author={Borisov, Vadim and Leemann, Tobias and Se{\ss}ler, Kathrin and Haug, Johannes and Pawelczyk, Martin and Kasneci, Gjergji},
  journal={IEEE Transactions on Neural Networks and Learning Systems},
  year={2022},
  publisher={IEEE}, 
  doi={10.1109/TNNLS.2022.3229161}}

@article{dwork2014DP,
  title={The algorithmic foundations of differential privacy},
  author={Dwork, Cynthia and Roth, Aaron and others},
  journal={Foundations and Trends in Theoretical Computer Science},
  volume={9},
  number={3--4},
  pages={211--407},
  year={2014},
  publisher={Now Publishers, Inc.}
}

@inproceedings{chen2020MIAGAN,
  title={{GAN-Leaks}: A taxonomy of membership inference attacks against generative models},
  author={Chen, Dingfan and Yu, Ning and Zhang, Yang and Fritz, Mario},
  booktitle={Proceedings of the 2020 ACM SIGSAC Conference on Computer and Communications Security},
  pages={343--362},
  year={2020}
}

@article{park2018DCR,
    author = {Park, Noseong and Mohammadi, Mahmoud and Gorde, Kshitij and Jajodia, Sushil and Park, Hongkyu and Kim, Youngmin},
    title = {Data Synthesis Based on Generative Adversarial Networks},
    year = {2018},
    publisher = {VLDB Endowment},
    volume = {11},
    number = {10},
    journal = {Proceedings of the VLDB Endowment},
    pages = {1071–1083}
}

@article{lowe2004NNDR,
  title={Distinctive image features from scale-invariant keypoints},
  author={Lowe, David G},
  journal={International Journal of Computer Vision},
  volume={60},
  pages={91--110},
  year={2004}
}

@article{hussein2021epiCluster,
  title={Cluster Analysis on {COVID}-19 outbreak sentiments from {T}witter data using {K}-means algorithm},
  author={Hussein, Adnan and Ahmad, Farzana Kabir and Kamaruddin, Siti Sakira},
  journal={Journal of System and Management Sciences},
  volume={11},
  number={4},
  pages={167--189},
  year={2021}
}

@article{taguchi2013epiPCA,
  title={Principal component analysis based feature extraction approach to identify circulating {microRNA} biomarkers},
  author={Taguchi, YH and Murakami, Yoshiki},
  journal={PloS One},
  volume={8},
  number={6},
  pages={e66714},
  year={2013},
  publisher={Public Library of Science San Francisco, USA}
}

@article{nagata2021epiAD,
  title={Detection of overdose and underdose prescriptions—An unsupervised machine learning approach},
  author={Nagata, Kenichiro and Tsuji, Toshikazu and Suetsugu, Kimitaka and Muraoka, Kayoko and Watanabe, Hiroyuki and Kanaya, Akiko and Egashira, Nobuaki and Ieiri, Ichiro},
  journal={PloS One},
  volume={16},
  number={11},
  pages={e0260315},
  year={2021},
  publisher={Public Library of Science San Francisco, CA USA}
}

@article{bithell1990epiKDE,
  title={An application of density estimation to geographical epidemiology},
  author={Bithell, John F},
  journal={Statistics in Medicine},
  volume={9},
  number={6},
  pages={691--701},
  year={1990},
  publisher={Wiley Online Library}
}

@article{Ramesh2022DALLE,
  title={Hierarchical Text-Conditional Image Generation with CLIP Latents},
  author={Aditya Ramesh and Prafulla Dhariwal and Alex Nichol and Casey Chu and Mark Chen},
  journal={Preprint},
  note = {{arXiv:2204.06125}},
  year={2022},
}

@inproceedings{ausset2021epiNF,
  title={Individual survival curves with conditional normalizing flows},
  author={Ausset, Guillaume and Ciffreo, Tom and Portier, Francois and Cl{\'e}men{\c{c}}on, Stephan and Papin, Timoth{\'e}e},
  booktitle={2021 IEEE 8th International Conference on Data Science and Advanced Analytics (DSAA)},
  pages={1--10},
  year={2021},
  organization={IEEE}
}

@article{correia2020GF,
  title={Joints in random forests},
  author={Correia, Alvaro and Peharz, Robert and de Campos, Cassio P},
  journal={Advances in Neural Information Processing Systems},
  volume={33},
  pages={11404--11415},
  year={2020}
}

@article{ng2001supgen,
  title={On discriminative vs. generative classifiers: A comparison of logistic regression and naive {B}ayes},
  author={Ng, Andrew and Jordan, Michael},
  journal={Advances in Neural Information Processing Systems},
  volume={14},
  year={2001}
}

@article{dosovitskiy2014unsupdiscr,
  title={Discriminative unsupervised feature learning with convolutional neural networks},
  author={Dosovitskiy, Alexey and Springenberg, Jost Tobias and Riedmiller, Martin and Brox, Thomas},
  journal={Advances in Neural Information Processing Systems},
  volume={27},
  year={2014}
}

@book{dangeti2017slusl_label,
  title={Statistics for Machine Learning},
  author={Dangeti, Pratap},
  year={2017},
  publisher={Packt Publishing Ltd},
  address = {Birmingham}
}

@book{babcock2021discgen_label,
  title={Generative AI with Python and TensorFlow 2},
  author={Babcock, Joseph and Bali, Raghav},
  year={2021},
  publisher={Packt Publishing Ltd},
  address = {Birmingham}
}

@book{bishop2006slusllabel_discgendistr,
  title={Pattern Recognition and Machine Learning},
  author={Bishop, Christopher M and Nasrabadi, Nasser M},
  volume={4},
  year={2006},
  publisher={Springer}, 
  address = {New York}
}

@book{foster2023overviewGEN1,
  title={Generative deep learning},
  author={Foster, D.},
  isbn={9781098134181},
  year={2023},
  publisher={O'Reilly Media}, 
  address = {Sebastopol}
}

@article{OpenAI2023GPT4,
  title={{GPT-4} {T}echnical Report},
  author={OpenAI},
  journal={Preprint},
  note = {{arXiv:2303.08774}},
  year={2023}
}

@article{brier1950verification,
  title={Verification of forecasts expressed in terms of probability},
  author={Brier, Glenn W},
  journal={Monthly Weather Review},
  volume={78},
  number={1},
  pages={1--3},
  year={1950}
}

@article{openML2013,
  author = {Joaquin Vanschoren and Jan N. van Rijn and Bernd Bischl and Luis Torgo},
  title = {{OpenML}: networked science in machine learning},
  journal = {SIGKDD Explorations},
  volume = {15},
  number = {2},
  year = {2013},
  pages = {49-60},
  publisher = {ACM}
}

@misc{heartdiseaseuci,
  author       = {Janosi,Andras and Steinbrunn,William and Pfisterer,Matthias and Detrano,Robert},
  title        = {{Heart Disease}},
  year         = {1988},
  howpublished = {UCI Machine Learning Repository},
  doi = {10.24432/C52P4X}
}

@book{van2011targeted,
  title={Targeted Learning: Causal Inference for Observational and Experimental Data},
  author={Van der Laan, Mark J and Rose, Sherri},
  volume={4},
  year={2011},
  publisher={Springer}, 
  address = {New York}
}

@article{chernozhukov2018double,
  title={Double/debiased machine learning for treatment and structural parameters},
  author={Chernozhukov, Victor and Chetverikov, Denis and Demirer, Mert and Duflo, Esther and Hansen, Christian and Newey, Whitney and Robins, James},
  journal = {The Econometrics Journal},
  volume = {21},
  number = {1},
  pages = {C1-C68},
  year = {2018}
}

@article{couronne2018random,
  title={Random forest versus logistic regression: a large-scale benchmark experiment},
  author={Couronn{\'e}, Raphael and Probst, Philipp and Boulesteix, Anne-Laure},
  journal={BMC Bioinformatics},
  volume={19},
  pages={1--14},
  year={2018},
  publisher={Springer}
}

@article{hothorn2004bagging,
  title={Bagging survival trees},
  author={Hothorn, Torsten and Lausen, Berthold and Benner, Axel and Radespiel-Tr{\"o}ger, Martin},
  journal={Statistics in Medicine},
  volume={23},
  number={1},
  pages={77--91},
  year={2004},
  publisher={Wiley Online Library}
}

@Article{mlr3,
    title = {{mlr3}: A modern object-oriented machine learning framework in {R}},
    author = {Michel Lang and Martin Binder and Jakob Richter and Patrick Schratz and Florian Pfisterer and Stefan Coors and Quay Au and Giuseppe Casalicchio and Lars Kotthoff and Bernd Bischl},
    journal = {Journal of Open Source Software},
    year = {2019},
      volume={4},
    number={44},
    pages={1903},
  }

@book{Bischl2024,
    title = {Applied Machine Learning Using mlr3 in R},
    editor = {Bernd Bischl and Raphael Sonabend and Lars Kotthoff and Michel Lang},
    url = {https://mlr3book.mlr-org.com},
    year = {2024},
    isbn = {9781032507545},
    publisher = {CRC Press}
}

\printindex

\newpage
\section*{Appendix}  \label{sec: appendix_dataset}
\addcontentsline{toc}{section}{Appendix}

Throughout the chapter, we use the heart disease dataset for illustration (see Sec.~\ref{sec:intro}). Table~\ref{table:Dataset_heart} provides a description and the corresponding values for each feature in the heart disease dataset. The features are divided into categorical (top) and numerical (bottom). For better understanding, we renamed the features compared to the original data source \citep{heartdiseaseuci}. For details, we refer to our GitHub page \url{https://github.com/bips-hb/epi-handbook-ml}.

\begin{table}[h]
\centering
\begin{tabular}{p{0.30\textwidth}p{0.3\textwidth}@{\hspace{ 0.03\textwidth}}p{0.3\textwidth} }
\toprule
Feature (short name) & Description & Values \\
\midrule
\texttt{heart\_disease} (\texttt{hd}) & Presence of heart disease & absent, present  \\
\texttt{chest\_pain} & Type of chest pain experienced & typical angina, atypical angina, non-anginal pain, asymptomatic \\
\texttt{exercise\_induced\_angina} (\texttt{eia}) & Exercise induced angina (chest pain) & no, yes \\
\texttt{fasting\_blood\_sugar} (\texttt{fbs}) & Fasting blood sugar \unit{mg/dl} & false, true \\
\texttt{sex} & Sex of the patient & female, male \\
\texttt{resting\_ecg} & Resting electrocardiographic results & normal, ST-T wave ab\-nor\-mal\-i\-ty\tablefootnote{T wave inversions and/or ST elevation or depression of $> 0.05$ mV}, showing probable or definite left ventricular hypertrophy by Estes' criteria\\
\texttt{ST\_slope} & Slope of the peak exercise ST\tablefootnote{Position on  electrocardiography (ECG) plot.} segment & upsloping, flat, downsloping \\
\texttt{thal} & Thallium stress test results\tablefootnote{HealthLine Contributors (2019). Thallium stress test. HealthLine. Available at:  \url{https://www.healthline.com/health/thallium-stress-test}} & normal, fixed defect, reversible  defect   \\
\midrule
\texttt{age} & Age of the patient in years & 29 - 77 \\
\texttt{max\_heart\_rate} (\texttt{mhr}) & Maximum heart rate achieved & 71 - 202 \\
\texttt{num\_major\_vessels} (\texttt{nmv}) & Number of major vessels colored by fluoroscopy & 0 - 3 \\
\texttt{ST\_depression} (\texttt{ST\_d}) & ST depression induced by exercise relative to rest & 0.0 - 6.2 \\
\texttt{resting\_blood\_pressure} (\texttt{rbp}) & Resting blood pressure (in \unit{mm.Hg}  on admission to the hospital) & 94 - 200 \\
\texttt{serum\_cholesterol} (\texttt{sc}) & Serum cholesterol level (in \unit{mg/dl}) & 126 - 564 \\
\bottomrule
\end{tabular}
\caption{Description and values of the features in the heart disease dataset. Divided into categorical (above) and numerical (below) features.} 
\label{table:Dataset_heart}
\end{table}

\end{document}